\documentclass{article}
\usepackage{Mis/arxiv}

\usepackage[utf8]{inputenc} 
\usepackage[T1]{fontenc}    
\usepackage{hyperref}       
\usepackage{url}            
\usepackage{booktabs}       
\usepackage{amsfonts}       
\usepackage{nicefrac}       
\usepackage{microtype}      
\usepackage{lipsum}		
\usepackage{graphicx}
\usepackage{natbib}
\usepackage{doi}

\usepackage{listings}
\usepackage{authblk}

\usepackage{amsfonts,amsmath,amssymb,amsthm}
\usepackage{booktabs}
\usepackage{longtable}
\usepackage{comment}
\usepackage{float,url}
\usepackage{lipsum}
\usepackage[ruled,vlined]{algorithm2e}
\usepackage{subfig}
\usepackage{caption}
\usepackage[acronym]{glossaries}
\makeglossaries
\usepackage{hyperref}
\newacronym{CDM}{CDM}{Causal Decision Making}
\newacronym{CSL}{CSL}{Causal Structure Learning}
\newacronym{CEL}{CEL}{Causal Effect Learning}
\newacronym{CPL}{CPL}{Causal Policy Learning}
\newacronym{RL}{RL}{Reinforcement Learning}
\newacronym{MBRL}{MBRL}{Model-based reinforcement learning}
\newacronym{ATE}{ATE}{Average Treatment Effect}
\newacronym{HTE}{HTE}{Heterogeneous Treatment Effect}
\newacronym{MDP}{MDP}{Markov Decision Process}
\newacronym{POMDP}{POMDP}{Partially Observable Markov Decision Process}
\newacronym{DTR}{DTR}{Dynamic Treatment Regimes}
\newacronym{ICU}{ICU}{intensive care unit}
\newacronym{EHR}{EHR}{electronic health record}
\newacronym{MIMIC-III}{MIMIC-III}{Medical Information Mart for Intensive Care III}
\newacronym{IV}{IV}{Instrumental Variables}
\newacronym{SOFA}{SOFA}{Sepsis-related Organ Failure Assessment}
\newacronym{SEM}{SEM}{Structural Equation Models}
\newacronym{LSEM}{LSEM}{linear structural equation model}
\newacronym{PC}{PC}{Peter-Clark}
\newacronym{CI}{CI}{conditional independence}
\newacronym{RCM}{RCM}{Rubin Causal Model} 
\newacronym{OPE}{OPE}{Off-Policy Evaluation}
\newacronym{OPO}{OPO}{Off-Policy Optimizaiton}
\newacronym{SCM}{SCM}{Structural Causal Model}
\newacronym{DAG}{DAG}{directed acyclic graph}
\newacronym{SUTVA}{SUTVA}{Stable Unit Treatment Value Assumption}
\newacronym{NUC}{NUC}{No Unmeasured Confounders}
\newacronym{SNPs}{SNPs}{single nucleotide polymorphisms}
\newacronym{MEC}{MEC}{Markov equivalence class}
\newacronym{CPDAG}{CPDAG}{completed partially directed acyclic graph}
\newacronym{OLS}{OLS}{ordinary least squares}
\newacronym{TE}{TE}{total effect}
\newacronym{DE}{DE}{direct effect}
\newacronym{IE}{IE}{indirect effect}
\newacronym{ANOCE}{ANOCE}{analysis of causal effects}
\newacronym{TCDF}{TCDF}{Temporal Causal Discovery Framework}
\newacronym{DM}{DM}{direct (outcome regression) estimator}
\newacronym{IPW}{IPW}{inverse probability weighting}
\newacronym{DR}{DR}{doubly robust}
\newacronym{AIPW}{AIPW}{augmented inverse probability weighting}
\newacronym{TMLE}{TMLE}{Targeted Maximum Likelihood Estimation}
\newacronym{DS}{DS}{double score}
\newacronym{IS}{IS}{importance sampling}
\newacronym{ATT}{ATT}{Average Treatment Effect on the Treated}
\newacronym{DiD}{DiD}{Difference-in-Difference}
\newacronym{SC}{SC}{Synthetic Control}
\newacronym{FQE}{FQE}{Fitted-Q Evaluation}
\newacronym{SVM}{SVM}{support vector machine}
\newacronym{CART}{CART}{classification and regression trees}
\newacronym{MAB}{MAB}{Multi-Armed Bandit}
\newacronym{EG}{EG}{Epsilon Greedy}
\newacronym{UCB}{UCB}{Upper Confidence Bound}
\newacronym{TS}{TS}{Thompson Sampling}
\newacronym{TD}{TD}{Temporal Difference}
\newacronym{SE}{SE}{spillover effect}
\newacronym{MARL}{MARL}{Multi-Agent RL}
\newacronym{PCE}{PCE}{Path-specific Counterfactual Effect}
\newacronym{2SLS}{2SLS}{Two-Stage Least Squares}
\usepackage[backgroundcolor=white]{todonotes}

\usepackage{graphicx}
\usepackage{tabularx}
\usepackage{makecell}
\usepackage{amssymb}
\usepackage{pifont}
\usepackage{natbib}

\usepackage{filecontents}
\usepackage{multirow}

\theoremstyle{definition}
\newtheorem{definition}{Definition}[section]
\theoremstyle{assumption}
\newtheorem{assumption}{Assumption}[section]

\theoremstyle{remark}

\usepackage{amsmath}
\DeclareMathOperator*{\argmax}{arg\,max}

\newcommand\blfootnote[1]{%
  \begingroup
  \renewcommand\thefootnote{}\footnotetext{#1}%
  \addtocounter{footnote}{-1}%
  \endgroup
}

\begin{document}
\date{}

\title{A Review of Causal Decision Making}

\author[1]{\normalsize \textbf{Lin Ge$^*$$^\ddag$}}
\author[2]{\normalsize \textbf{Hengrui Cai$^*$}}
\author[1]{\normalsize \textbf{Runzhe Wan$^*$$^\ddag$}}
\author[3]{\normalsize \textbf{Yang Xu$^*$}}
\author[1]{\normalsize \textbf{Rui Song$^\dag$$^\ddag$}}
\affil[1]{\normalsize Amazon, Seattle, USA}
\affil[2]{\normalsize University of California, Irvine, USA}
\affil[3]{\normalsize North Carolina State University, Raleigh, USA}


\maketitle

\begin{abstract}
To make effective decisions, it is important to have a thorough understanding of the causal relationships among actions, environments, and outcomes. This review aims to surface three crucial aspects of decision-making through a causal lens: 1) the discovery of causal relationships through causal structure learning, 2) understanding the impacts of these relationships through causal effect learning, and 3) applying the knowledge gained from the first two aspects to support decision making via causal policy learning. 
    Moreover, we identify challenges that hinder the broader utilization of causal decision-making and discuss recent advances in overcoming these challenges.
    Finally, we provide future research directions to address these challenges and to further enhance the implementation of causal decision-making in practice, with real-world applications illustrated based on the proposed causal decision-making. We aim to offer a comprehensive methodology and practical implementation framework by consolidating various methods in this area into a Python-based collection. URL: \url{https://causaldm.github.io/Causal-Decision-Making}.

\blfootnote{$^*$Equal contribution.}
\blfootnote{$^\dag$ Corresponding author.}
\blfootnote{$^\ddag$ This work does not relate to the positions at Amazon.}
\end{abstract}


\section{Introduction}%

Decision-making is at the heart of artificial intelligence systems, enabling agents to navigate complex environments, achieve goals, and adapt to changing conditions. Traditional decision-making frameworks often rely on associations or statistical correlations between variables, which can lead to suboptimal outcomes when the underlying causal relationships are ignored \citep{pearl2009causal}. 
The rise of causal inference as a field has provided powerful frameworks and tools to address these challenges, such as structural causal models and potential outcomes frameworks \citep{rubin1978bayesian,pearl2000causality}. 
Unlike traditional methods, \textit{causal decision-making} focuses on identifying and leveraging cause-effect relationships, allowing agents to reason about the consequences of their actions, predict counterfactual scenarios, and optimize decisions in a principled way \citep{spirtes2000causation}. In recent years, numerous decision-making methods based on causal reasoning have been developed, finding applications in diverse fields such as recommender systems \citep{zhou2017large}, clinical trials \citep{durand2018contextual}, finance \citep{bai2024review}, and ride-sharing platforms \citep{wan2021pattern}. Despite these advancements, a fundamental question persists: 

\begin{center}
    \textit{When and why do we need causal modeling in decision-making?}
\end{center} 




This question is closely tied to the concept of counterfactual thinking—reasoning about what might have happened under alternative decisions or actions. Counterfactual analysis is crucial in domains where the outcomes of unchosen decisions are challenging, if not impossible, to observe. For instance, a business leader selecting one marketing strategy over another may never fully know the outcome of the unselected option \citep{rubin1974estimating, pearl2009causal}. Similarly, in econometrics, epidemiology, psychology, and social sciences, \textit{the inability to observe counterfactuals directly often necessitates causal approaches} \citep{morgan2015counterfactuals, imbens2015causal}. 
Conversely, non-causal analysis may suffice in scenarios where alternative outcomes are readily determinable. For example, a personal investor's actions may have negligible impact on stock market dynamics, enabling potential outcomes of alternate investment decisions to be inferred from existing stock price time series \citep{angrist2008mostly}. However, even in cases where counterfactual outcomes are theoretically calculable—such as in environments with known models like AlphaGo—exhaustively computing all possible outcomes is computationally infeasible \citep{silver2017mastering, silver2018general}. 
In such scenarios, causal modeling remains advantageous by offering \textit{structured ways to infer outcomes efficiently and make robust decisions}.

\begin{figure}[!t]
    \centering
    \includegraphics[width = .75\linewidth]{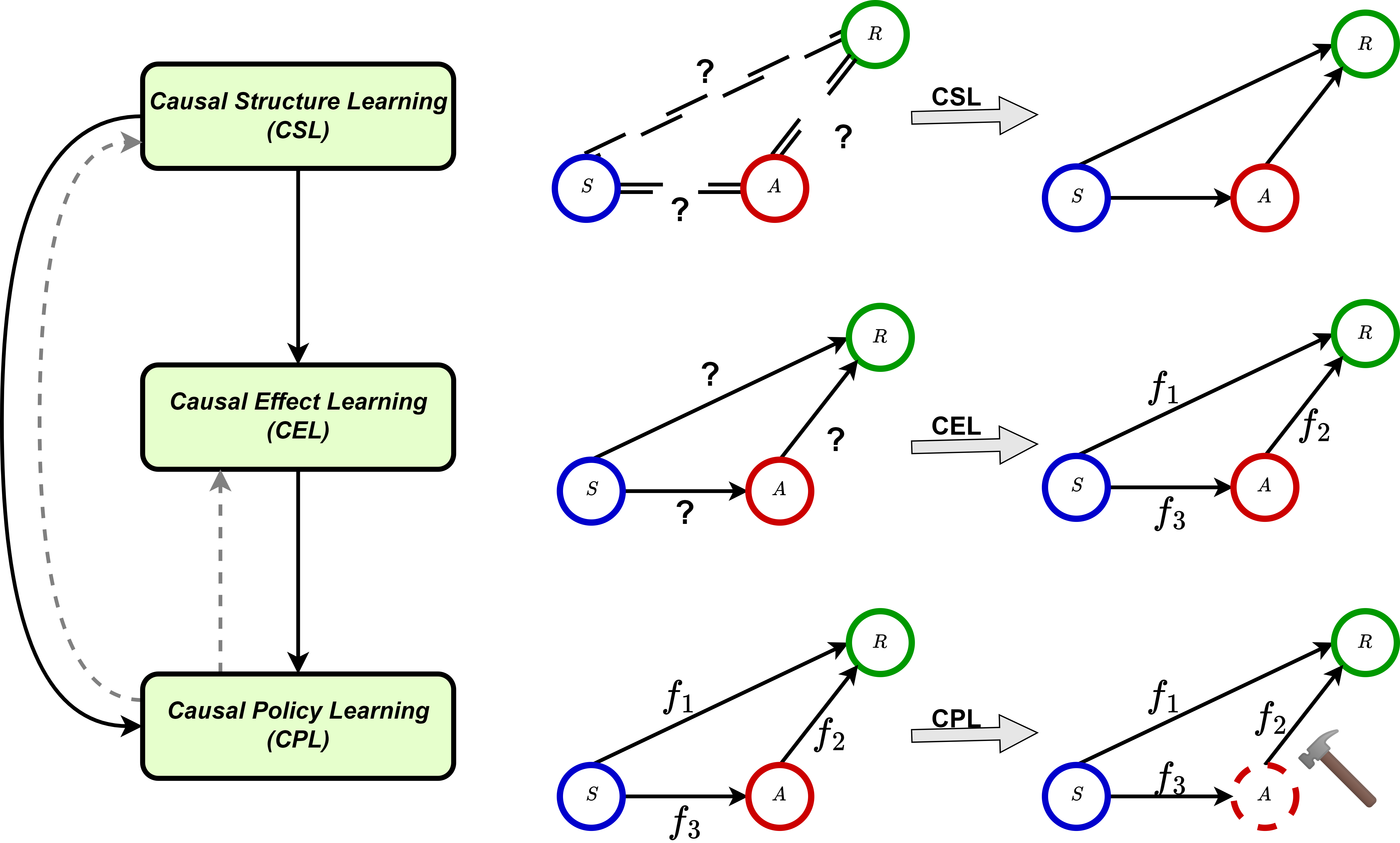}
    \caption{Workflow of the \acrlong{CDM}. $f_1$, $f_2$, and $f_3$ represent the impact sizes of the directed edges. Variables enclosed in solid circles are observed, while those in dashed circles are actionable.}\label{fig:cdm}
\end{figure}

Most existing works primarily assume either sophisticated prior knowledge or strong causal models to conduct follow-up decision-making. To make effective and trustworthy decisions, it is critical to have a thorough understanding of the causal relationships among actions, environments, and outcomes. This review synthesizes the current state of research in \acrfull{CDM}, providing an overview of foundational concepts, recent advancements, and practical applications. Specifically, this work discusses the connections of \textbf{three primary components of decision-making} through a causal lens: 1) discovering causal relationships through \textit{\acrfull{CSL}}, 2) understanding the impacts of these relationships through \textit{\acrfull{CEL}}, and 3) applying the knowledge gained from the first two aspects to decision making via \textit{\acrfull{CPL}}. 

Let $\boldsymbol{S}$ denote the state of the environment, which includes all relevant feature information about the environment the decision-makers interact with, $A$ the action taken, $\pi$ the action policy that determines which action to take, and $R$ the reward observed after taking action $A$. As illustrated in Figure \ref{fig:cdm}, \acrshort{CDM} typically begins with \acrshort{CSL}, which aims to uncover the unknown causal relationships among various variables of interest. Once the causal structure is established, \acrshort{CEL} is used to assess the impact of a specific action on the outcome rewards. To further explore more complex action policies and refine decision-making strategies, \acrshort{CPL} is employed to evaluate a given policy or identify an optimal policy. In practice, it is also common to move directly from \acrshort{CSL} to \acrshort{CPL} without conducting \acrshort{CEL}. Furthermore, \acrshort{CPL} has the potential to improve both \acrshort{CEL} and \acrshort{CSL} by facilitating the development of more effective experimental designs \citep{zhu2019causal,simchi2023multi} or adaptively refining causal structures \citep{sauter2024core}. 

\begin{figure}[!t]
    \centering
    \includegraphics[width = .9\linewidth]{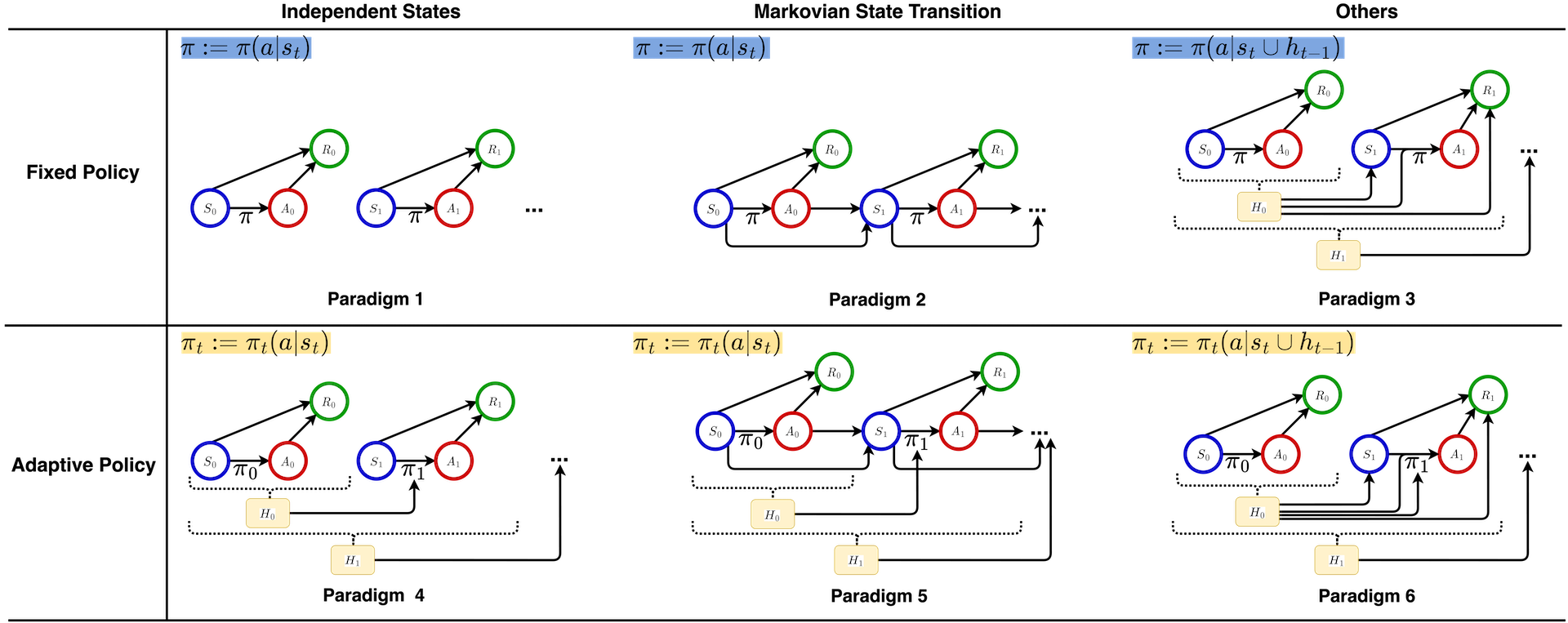}
    \caption{Common data dependence structures (paradigms) in \acrshort{CDM}. Detailed notations and explanations can be found in Section \ref{sec:paradigms}.}
    \label{Fig:paradigms}
\end{figure}
Building on this framework, decision-making problems discussed in the literature can be further categorized into \textbf{six paradigms}, as summarized in Figure \ref{Fig:paradigms}. These paradigms summarize the common assumptions about data dependencies frequently employed in practice. Paradigms 1-3 describe the data structures in offline learning settings, where data is collected according to an unknown and fixed behavior policy. In contrast, paradigms 4-6 capture the online learning settings, where policies dynamically adapt to newly collected data, enabling continuous policy improvement. These paradigms also reflect different assumptions about state dependencies. The simplest cases, paradigms 1 and 4, assume that all observations are independent, implying no long-term effects of actions on future observations. To account for sequental dependencies, the \acrfull{MDP} framework, summarized in paradigms 2 and 5, assumes Markovian state transition. Specifically, it assumes that given the current state-action pair $(S_t, A_t)$, the next state $S_{t+1}$ and reward $R_t$ are independent of all prior states $\{S_j\}_{j < t}$ and actions $\{A_j\}_{j < t}$. When such independence assumptions do not hold, paradigms 3 and 6 account for scenarios where all historical observations may impact state transitions and rewards. This includes but not limited to researches on \acrfull{POMDP} \citep{hausknecht2015deep, littman2009tutorial}, panel data analysis \citep{hsiao2007panel,hsiao2022analysis}, \acrfull{DTR} with finite stages \citep{chakraborty2014dynamic, chakraborty2013statistical}. 

Each \acrshort{CDM} task has been studied under different paradigms, with \acrshort{CSL} extensively explored within paradigm 1. \acrshort{CEL} and offline \acrshort{CPL} encompass paradigms 1-3, while online \acrshort{CPL} spans paradigms 4-6. By organizing the discussion around these three tasks and six paradigms, this review aims to provide a cohesive framework for understanding the field of \acrlong{CDM} across diverse tasks and data structures.



\textbf{Contribution.} In this paper, we conduct a comprehensive survey of \acrshort{CDM}. 
Our contributions are as follows. 
\begin{itemize}
    \item We for the first time organize the related causal decision-making areas into three tasks and six paradigms, connecting previously disconnected areas (including economics, statistics, machine learning, and reinforcement learning) using a consistent language. For each paradigm and task, we provide a few taxonomies to establish a unified view of the recent literature.
    \item We provide a comprehensive overview of \acrshort{CDM}, covering all three major tasks and six classic problem structures, addressing gaps in existing reviews that either focus narrowly on specific tasks or paradigms or overlook the connection between decision-making and causality (detailed in Section \ref{sec::related_work}).
    \item We provide real-world examples to illustrate the critical role of causality in decision-making and to reveal how \acrshort{CSL}, \acrshort{CEL} and \acrshort{CPL} are inherently interconnected in daily applications, often without explicit recognition.
    \item We are actively maintaining and expanding a GitHub repository and online book, providing detailed explanations of key methods reviewed in this paper, along with a code package and demos to support their implementation, with URL: \url{https://causaldm.github.io/Causal-Decision-Making}.
\end{itemize}
The remainder of this paper is organized as follows: Section \ref{sec::related_work} provides an overview of related survey papers. Section \ref{sec:preliminary} introduces the foundational concepts, assumptions, and notations that form the foundation for the subsequent discussions. In Section \ref{sec:3task6paradigm}, we offer a detailed introduction to the three key tasks and six learning paradigms in \acrshort{CDM}. Sections \ref{Sec:CSL} through \ref{sec:Online CPL} form the core of the paper, with each section dedicated to a specific topic within \acrshort{CDM}: \acrshort{CSL}, \acrshort{CEL}, Offline \acrshort{CPL}, and Online \acrshort{CPL}, respectively. Section \ref{sec:assump_violated} then explores extensions needed when standard causal assumptions are violated. To illustrate the practical application of the \acrshort{CDM} framework, Section \ref{sec:real_data} presents two real-world case studies. Finally, Section \ref{sec:conclusion} concludes the paper with a summary of our contributions and a discussion of additional research directions that are actively being explored.

\section{Related Work} \label{sec::related_work}
Many reviews have examined causality or decision-making, but to the best of our knowledge, they typically focus on an individual paradigm or an individual task, with some emphasizing methodologies without clearly delineating the fundamental connections between causality and decision-making. In contrast, this review offers a unified framework that integrates all key steps, explicitly illustrating the role of causality and the relationships across different stages of the decision-making process. A detailed discussion of related surveys follows.

\noindent \textbf{Causal Inference and Causal Structural Learning.} In recent years, several review papers have emerged in the field of causal inference, typically categorized into two main frameworks: the \acrfull{SEM} framework, introduced by Pearl \citep{pearl1995causal}, and the potential outcomes framework, pioneered by Rubin \citep{rubin1974estimating, rubin1978bayesian}.

The \acrshort{SEM} framework models causal relationships using graphical structures, where nodes represent variables and directed edges depict cause-effect relationships. Pearl’s work \citep{pearl2003statistics} was instrumental in developing the do-calculus to formalize the effect of interventions on causal diagrams. Subsequent review papers \citep{pearl2009causal, pearl2010causal, pearl2010foundations} also provide comprehensive overviews of \acrshort{CSL}, with detailed discussions on related topics such as confounding issues and and mediation analysis. To the best of our knowledge, there is no recent review that comprehensively summarizes the latest advances in causal discovery under \acrshort{SEM}, which is one of our focus in Section \ref{Sec:CSL}.

The potential outcome framework, also known as the \acrfull{RCM} \citep{rubin1974estimating}, defines causal effects by comparing potential outcomes under different treatment conditions. A key resource in this area is \citet{imbens2015causal}, which systematically summarizes the origins and development of causal inference with the \acrshort{RCM}, covering topics from estimation and inference to sensitivity analysis. Recent reviews have also addressed specific aspects of causal inference, including observational studies \citep{yao2021survey}, matching methods \citep{stuart2010matching}, handling missing data \citep{ding2018causal}, and addressing confounding in text analysis \citep{keith2020text}. This part of the work closely aligns with \acrshort{CEL}, where previous studies may fall short in comprehensively summarizing effect learning across different data structures and paradigms. Our work addresses this gap by incorporating scenarios where assumptions are both satisfied and violated, positioning it as a critical intermediate task in the decision-making process.

\noindent 
\textbf{Policy Learning.} In the offline policy learning area, the related review papers can be classified as focusing on \acrfull{OPE} and \acrfull{OPO}.
For \acrshort{OPE}, 
\citet{voloshin2019empirical} systemetically studys the empirical performance of a list of common \acrshort{OPE} methods for the offline \acrfull{RL} setting (i.e. paradigm 2). 
\citet{uehara2022review} is the latest review of the key methods and theories in \acrshort{OPE}, covering paradigms 1-3. For \acrshort{OPO}, \citet{prudencio2023survey} and  \citet{Sergey2020offlineRL} both review the key concepts, methods and open problems in offline \acrshort{RL} (paradigm 2). Besides, from the statistics perspective, \citet{kosorok2019precision} comprehensively reviewed the progress of applying \acrshort{DTR} to precision medicine, covering paradigms 1 and 3. 

In contrast, most reviews on online policy learning focus on policy optimization, with online policy evaluation being a newer and less explored area. For policy optimization in paradigm 4, \citet{lattimore2020bandit} and \citet{slivkins2024introductionmultiarmedbandits} offer the most recent comprehensive texts on bandit algorithms. These works cover a broad range of topics, with a particular emphasis on algorithm design and regret analysis, including stochastic bandits, adversarial bandits, contextual bandits, etc.  
In the broader context of online policy learning, problems modeled as \acrshort{MDP}s (paradigm 5) are typically studied through \acrshort{RL}. \citet{shakya2023reinforcement} offers a thorough overview of \acrshort{RL} fundamentals, while \citet{wang2022deep} and \citet{arulkumaran2017brief} focus on RL's integration with deep learning. \citet{gu2022review} reviews \acrshort{RL} methods designed to address safety concerns in real-world applications, and \citet{canese2021multi} and \citet{gronauer2022multi} review multi-agent RL. For more complex settings like \acrshort{POMDP}s (paradigm 6), \citet{xiang2021recent} provides a detailed review of recent advances. However, these reviews generally overlook the relationship between causality and policy learning.

Recently, the integration of causal knowledge into policy learning has garnered growing attention, leading to the emergence of the field of causal \acrshort{RL}. For example, \citet{scholkopf2021toward} briefly discusses the role and importance of causality in \acrshort{RL}, with a main focus on causal representation learning. \citet{kaddour2022causal} examined causal machine learning, including a brief chapter summarizing how \acrshort{RL} can benefit from exploiting causal paradigms. Additionally, \citet{grimbly2021causal} reviewed causal multi-agent \acrshort{RL}, and \citet{bannon2020causality} focused on causality in batch \acrshort{RL}. The reviews by \citet{zeng2023survey} and \citet{deng2023causal} are the most comprehensive, outlining how causal knowledge from causal discovery and causal inference can address challenges faced by non-causal \acrshort{RL} and systematically reviewing existing causal \acrshort{RL} methods.





\section{Preliminary} \label{sec:preliminary}
In this section, we provide a brief overview of the key concepts and assumptions commonly used throughout this paper. We begin with an introduction to the general principles of causal inference and reinforcement learning. Next, we delve into the specific concepts related to \acrshort{CSL}, \acrshort{CEL}, and \acrshort{CPL} in Section \ref{sec:prelim_CSL}-\ref{sec:p2_MDP}, and conclude with the assumptions outlined in Section \ref{sec:prelim_assump}.

\begin{definition} (Potential Outcome): For each individual, denote $A=a$ as the action or treatment assignment. We define $R(A=a)$ as the outcome/reward if the individual receives action $A=a$.
The potential outcomes framework, also known as the Neyman-Rubin Causal Model, is a foundational concept in causal inference.
\end{definition} 

\begin{definition}(Do-Operator): Given any two variables $X$ and $Y$ in a causal system, the do-operator denotes an intervention on $X$, which is often defined as $do(X=x)$. The conditional probability of $Y$ given $do(X=x)$ is defined as $\mathbb{P}(Y|do(X=x))$.
\end{definition}
Without further assumptions about the causal structure involving $(A,R)$, the probability $\mathbb{P}(R|do(A=a))$ generally differs from $\mathbb{P}(R|A=a)$. The discrepancy arises because $\mathbb{P}(R|do(A=a))$ represents the probability of $R$ under an intervention where $A$ is forcibly set to $a$, while all other potential causes of $R$, whether observed or not, are held fixed. Mathematically, if we denote $Z$ as the set of all other variables that are causally upstream of $R$ (excluding $X$), the intervention probability can be expressed as
$$
\mathbb{P}(R|do(A=a)) = \sum_z \mathbb{P}(R|A=a,Z=z)\mathbb{P}(Z=z),
$$
which captures how intervening on $A$ with do-operator disrupts the natural causal mechanisms.
\begin{definition}(Confounder): In causal structures, a variable $C$ is considered a confounder between $A$ and $R$ if $C$ is a common cause of $(A,R)$, i.e. $C\rightarrow A$ and $C\rightarrow R$.
    
\end{definition}

\begin{definition}(Mediator): A variable $M$ is considered a mediator between $A$ and $R$ if $M$ is causally downstream of $A$ but upstream of $R$, i.e. $A\rightarrow M\rightarrow R$.
    
\end{definition}

\begin{definition}(Decision Process): 
A decision process is a framework used to describe the evolution of states, actions and rewards over time. In this general setting, with the dataset being $\{s_0, a_0, r_0, s_1, a_1, r_1, \dots, s_t, a_t, r_t, \dots\}$, the probability of reaching a future state and receiving a reward can depend on the entire history of states and actions up to that point as 
$
P(s_{t+1}, r_{t+1} \mid s_0, a_0, s_1, a_1, \ldots, s_t, a_t).
$
\end{definition}

\begin{definition}\label{def:MDP}(\acrfull{MDP}): 
\acrshort{MDP} is a special type of decision process where the probability of transitioning to the next state and receiving a reward depends only on the current state and action, and not on any previous states or actions. This ``memoryless'' property simplifies decision-making because it allows the process to be fully described by the current state and action alone. Formally, this is represented as
$
P(s_{t+1}, r_{t+1} \mid s_t, a_t, s_{t-1}, a_{t-1}, \ldots, s_0, a_0) = P(s_{t+1}, r_{t+1} \mid s_t, a_t).
$
\end{definition}


\subsection{Causal Graphical Model under a Potential Outcome Framework} \label{sec:prelim_CSL}

Consider a graph $\mathcal{G} =({X},{D}_{{X}})$ with a node set ${X}$ and an edge set ${D}_{{X}}$. There is at most one edge between any pair of nodes. If there is an edge between $X_i$ and $X_j$, then $X_i$ and $X_j$ are adjacent. A node $X_i$ is said to be a parent of $X_j$ if there is a directed edge from $X_i$ to $X_j$, i.e., $X_i$ is a direct cause of $X_j$. A node $X_k$ is said to be an ancestor of $X_j$ if there is a directed path from $X_k$ to $X_j$ regulated by at least one additional node $X_i$ for $i\not =k$ and $i \not =j$, i.e., $X_k$ is an indirect cause of $X_j$. Let the set of all parents/ancestors of node $X_j$ in $\mathcal{G}$ as $\textsc{PA}_{X_j} (\mathcal{G})$. A path from $X_i$ to $X_j$ in $\mathcal{G}$ is a sequence of distinct vertices, $\pi := \{a_0, a_1,\cdots,a_L\}\subset V$ such that $a_0 =X_i$, and $a_L=X_j$. A directed path from $X_i$ to $X_j$ is a path between $X_i$ and $X_j$ where all edges are directed towards $X_j$.  A directed graph $\mathcal{G}$ that does not contain directed cycles is called a \acrfull{DAG}. A directed graph is acyclic if and only if it has a topological ordering.

The \acrfull{SCM} characterizes the causal relationship among $|{X}|=d$ nodes via a DAG $\mathcal{G}$ and noises ${e}_{{X}} = [e_{X_1},\cdots,e_{X_d}]^\top$ such that
$X_i := h_i\{\textsc{PA}_{X_i} (\mathcal{G}), e_{X_i}\}$ for some unknown $h_i$ and $i=1,\cdots,d$. Here, we allow the collections of nodes to take different causal roles in the causal graph. For instance, let $A \in \mathbb{R}$ be an exposure/treatment, $M := (M_1,M_2,\cdots,M_p)^{\top} \in \mathbb{R}^p$ be mediators with dimension $p$ in its support ${M} = \mathcal{M}_1 \times \cdots \times \mathcal{M}_p \subseteq \mathbb{R}^p$, and $R \in \mathbb{R}$ be the outcome of interest. Additionally, we also consider that there are $t - 1$ confounders ${S}: = (S_1, \ldots, S_{t - 1})^{\top} \in \mathbb{R}^{t - 1}$ in its support $\mathcal{S} \subseteq \mathbb{R}^{t - 1}$. We would just let $t = 1$ here to represent the absence of confounders, that is $\mathcal{S} = \varnothing$. Suppose that there exists a \acrshort{DAG} $\mathcal{G} =({X},{D}_{{X}})$ that characterizes the causal relationship among ${X}=({S}^{\top}, A, {M}^\top, R)^\top$, where the dimension of ${X}$ is $d = t + p + 1$. 

In such a scenario, we can also define the potential outcome framework \citep{rubin1978bayesian} through the `do-operator' \citep{pearl2000causality}. Specifically, using the reward and action as an example, let $R(a)\equiv R(A=a)$ be the potential reward $R$ that would be observed after setting the action variable $A$ as $a$, following notation in \citet{rubin1978bayesian}'s framework. This term is stated to be equivalent to the value of $R$ by imposing a `do-operator' of $do(A=a)$ as in \citet{pearl2009causal}:
\begin{eqnarray*}
    R(a)\equiv R(A=a)\equiv R\{do(A=a)\},
\end{eqnarray*}
where $do(A=a)$ is a mathematical operator to simulate physical interventions that hold $A$ constant as $a$ while keeping the rest of the model unchanged, which corresponds to removing edges into $A$ and replacing $A$ by the constant $a$ in $\mathcal{G}$. 
Similarly, one can define the potential outcome,  $R(X_i=x_i)$, by setting an individual variable $X_i$ as $x_i$, while keeping the rest of the model unchanged.  
Suppose we observe data  ${X} = ({S}^{\top}, A, {M}^\top, R)^\top$ for $n$ subjects. The goal is to learn decision-oriented causal graphs $\mathcal{G}$ presenting the causal relationship among ${X}$ based on observed data.

\subsection{Treatment Effect Estimation under a Potential Outcome Framework}

The fundamental challenge in causal inference is counterfactual estimation. Specifically, once a decision has been made and action $A=a$ has been taken, we can only observe the outcome for that action. As a result, estimating the missing potential outcome $R(a')$ for the alternative action $A=a'$ becomes crucial. As a primary task in causal inference, treatment effect estimation can involve different concepts depending on the specific problem setting. Common causal estimands include the \acrfull{ATE}, \acrfull{HTE}, and Mediation Effect. 
\begin{definition} (\acrshort{ATE}): Under either potential outcome's framework or do-operator system, the Average treatment effect is defined as
        \begin{equation*}
    \text{ATE} = \mathbb{E}[R(1) - R(0)] = \mathbb{E}[ R|do(A=1)] -  \mathbb{E}[ R|do(A=0)].
    \end{equation*}
\end{definition}

For instance, when investigating the volume of fluids administered to patients with diabetes and its impact on their health status within 48 hours, the first question to address is, ``Is this IV fluid generally effective in reducing the mortality rate?'' This question pertains to estimating the treatment effect on the overall patient population, which is quantified by ATE as defined above.

\begin{definition} (\acrshort{HTE}): To account for the heterogeneous effects across different individuals or contextual groups, the HTE is defined as
    \begin{equation*}
    \tau(s) = \mathbb{E}[R(1) - R(0)|S=s] = \mathbb{E}[ R|do(A=1),S=s] -  \mathbb{E}[ R|do(A=0),S=s].
    \end{equation*}
\end{definition}
Unlike ATE which focuses on the overall effect across the population, HTE further explores the variation in treatment effects across different subgroups or individuals. In the diabetes example, HTE aims to understand whether IV fluid administration leads to different levels of causal effects for patients with varying characteristics, such as age, gender, or prescription history, as captured by the state variable $S$.

\begin{definition} (Mediation Effect): When mediators involved, the \acrfull{TE} can be decomposed into the natural \acrfull{DE}, and the natural \acrfull{IE}, where
    \begin{equation*}
    \begin{aligned}
    \text{TE}&= \mathbb{E}[R|do(A=a_1)]-\mathbb{E}[R|do(A=a_0)]\\
    \text{DE}&= \mathbb{E}[R|do(A=a_1,M=m^{(a_0)})]-\mathbb{E}[R|do(A=a_0), M=m^{(a_0)}]\\
    \text{IE}&= \mathbb{E}[R|do(A=a_1,M=m^{(a_1)})]-\mathbb{E}[R|do(A=a_0,M=m^{(a_1)})]\\
    \end{aligned}
    \end{equation*}
with $\text{TE} = \text{DE}+ \text{IE}$. 
\end{definition}
In the context of diabetes, the \acrfull{SOFA} score can be considered a mediator influenced by the administration of IV input. This score, in turn, affects the mortality rate within the next 48 hours. The mediation effect in this case allows us to decompose the total effect of IV input on mortality into two components: the direct effect, which directly measures how IV input impacts mortality, and the indirect effect, which operates through the SOFA score before ultimately influencing mortality.

\subsection{Causal Policy Learning under a Potential Outcome Framework} \label{sec:p2_MDP}

\begin{definition}\label{def:policy} (Policy): 
The policy $\pi$ is the agent's strategy, denoted as a function from the relevant information (context/state in Paradigm 1-2 or 4-5; all historical information in non-markovian decision process)  to the action (with a deterministic policy) or the action's probability space (with a random policy). 
\end{definition}

We commonly use the value functions to evaluate 
the goodness of a policy. 
We use a discounted infinite-horizon setting to illustrate. 

\begin{definition} (V-function): 
A policy $\pi$'s state value function (V-function) is  
\begin{eqnarray*}
	V^{\pi}(s)=\sum_{t\ge 0} \gamma^t \mathbb{E} \{Y_t^*(\pi)|S_0=s\},
\end{eqnarray*}
where $0<\gamma<1$ is a discount factor that reflects the trade-off between immediate and future outcomes. The value function measures the discounted cumulative outcome that the agent would receive had they followed $\pi$. 
\end{definition}

\begin{definition} (Q-function): 
A policy $\pi$'s state-action value function (Q-function) is  
\begin{eqnarray*}
	Q^{\pi}(a,s)=\sum_{t\ge 0} \gamma^t \mathbb{E} \{Y_t^*(\pi)|S_0=s, A_0 = a\}. 
\end{eqnarray*}
\end{definition}

\begin{definition}(Optimal Policy):
The optimal policy, $\pi^*$, is defined as $$\pi^* = \arg \max_{\pi} V^\pi(s), \forall s \in \mathcal{S}.$$
\end{definition}

\begin{definition} (Bellman optimality equations): 
The Q-learning-type policy learning is commonly based on the Bellman optimality equation, which characterizes the optimal policy $\pi^*$ and is commonly used in policy optimization. 
Specifically, $Q^*$ is the unique solution of 
\begin{equation}
    Q(a, s) = \mathbb{E} \Big(R_t + \gamma \arg \max_{a'} Q(a, S_{t+1})  | A_t = a, S_t = s \Big).\label{eqn:bellman_Q}
\end{equation}
The concepts above can be extended to the non-markovian case as well, with the state variable replaced by the full history. 
\end{definition}




\subsection{Three Key Causal Identifiability Assumptions}\label{sec:prelim_assump}
To address the problem of counterfactural estimation, causal inference typically relies on three key assumptions. While recent research has focused on relaxing these assumptions, we will first detail them here and discuss scenarios where these assumptions may be violated in Section \ref{sec:assump_violated}.

\begin{assumption}\label{assump:SUTVA} \acrfull{SUTVA} states that
    \begin{equation}
    R_i = \sum_{A=a}R_i(a) {1}\{A_i=a\} , i \in\{ 1, \cdots, n\},
    \end{equation}
    which can be broken down into two key sub-assumptions: (i) No interference between units, meaning the potential outcomes for one unit are unaffected by the actions assigned to other units, and (ii) Consistency of treatment, meaning there are no different versions of the same action that could lead to different potential outcomes. 
\end{assumption}

\begin{assumption}\label{assump:NUC} \acrfull{NUC} assumption states that 
$$R(a) \perp\!\!\!\perp A|S,\quad \forall a\in\mathcal{A},$$  
which quantifies the conditional independence of potential outcomes on the action being taken. 
\end{assumption}
For example, when investigating whether regular exercise reduces the risk of heart disease, genetic factors might influence both a person’s likelihood to exercise and their risk of heart disease. In this case, genetic predisposition acts as an unmeasured confounder, violating the \acrshort{NUC} assumption by affecting both the treatment (exercise) and the outcome (heart disease risk).

\begin{assumption}\label{assump:Positivity} Positivity, or Overlap assumption states that 
$$0 < c_0< P(A=a|S) <c_1< 1, \quad \forall a\in\mathcal{A},$$
which assumes that every unit in the study population has a non-zero probability of receiving each possible treatment or intervention.
\end{assumption}

\section{Three Tasks and Six Paradigms for \acrshort{CDM} }\label{sec:3task6paradigm}

\subsection{Three Tasks} \label{sec:3tasks}

\textbf{\acrlong{CSL}.} In recent years, \textit{causal discovery}, also known as \acrshort{CSL} \citep[e.g.,][]{pearl2000causality,peters2017elements}, has gained great attention for disentangling complex causal relationships in many areas \citep[e.g.,][]{nandy2017estimating,chakrabortty2018inference,cai2020anoce}. Building upon causal graphical models, \citep[see Section \ref{sec:prelim_CSL} and][for a comprehensive review]{pearl2009causal}, several CSL methods have been proposed \citep[e.g.,][]{spirtes2000constructing,shimizu2006linear,kalisch2007estimating,buhlmann2014cam,zheng2018dags,yu2019dag,zhu2019causal} to estimate the causal graphs from observed data, which is a crucial step in understanding the underlying mechanisms that govern changes in a system. It involves identifying the causal relationships between variables (see an illustration in the first panel of Figure \ref{fig:cdm}), which is fundamental for any subsequent analysis aiming to understand causal effects and make causal decisions. Most existing methodologies for average/heterogeneous treatment effects and personalized decision-making rely on a known causal structure, which provides the convenience of identifying the right variables to control (e.g., confounders), to intervene (e.g., treatments), and to optimize (e.g., rewards). However, such convenience is absent in many emerging real applications where the causal structure is unknown. Causal discovery thus has attracted more and more attention recently to infer causal structure from data and disentangle the complex relationship among variables.  In Section \ref{Sec:CSL}, we present state-of-the-art techniques for learning the skeleton of unknown causal relationships among input variables with embedded treatments.

\textbf{\acrlong{CEL}.} \acrshort{CEL} is a process of determining cause-and-effect relationships between variables \citep{pearl2003statistics,pearl2009causal}. Given a causal structure, either pre-assumed based on domain knowledge or derived using \acrshort{CSL} methods, the goal of \acrshort{CEL} is to identify, estimate, and infer the causal effect of interest. \acrshort{CEL} primarily focuses on estimating several key effects, including the \acrfull{ATE} \citep{hirano2003efficient,imbens2004nonparametric}, which measures the overall impact of a treatment across the entire population; the \acrfull{HTE} \citep{wager2018estimation,curth2021nonparametric}, which captures how the treatment's effect varies across different subgroups or individuals; and the mediation effect, which decomposes the causal effect by considering intermediate variables or mediators \citep{vanderweele2016mediation} that influence the relationship between the treatment and the outcome. In Section \ref{sec:CEL}, we provide a comprehensive review of the literature on \acrshort{CEL}, covering various paradigms and data structures.

\begin{figure}[t!]
        \subfloat[Offline Policy Optimization]{%
            \includegraphics[width=.48\linewidth]{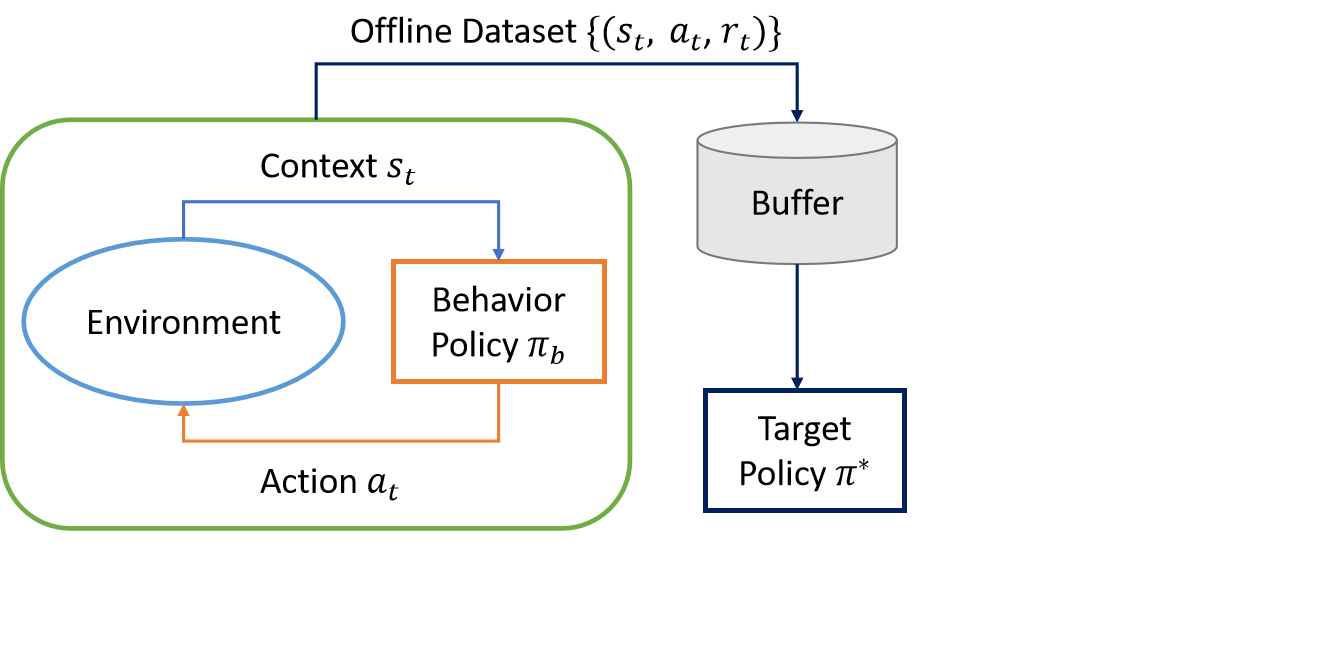}%
            \label{Fig:Off_PO}
        }\hfill
        \subfloat[Online Policy Optimization]{%
            \includegraphics[width=.48\linewidth]{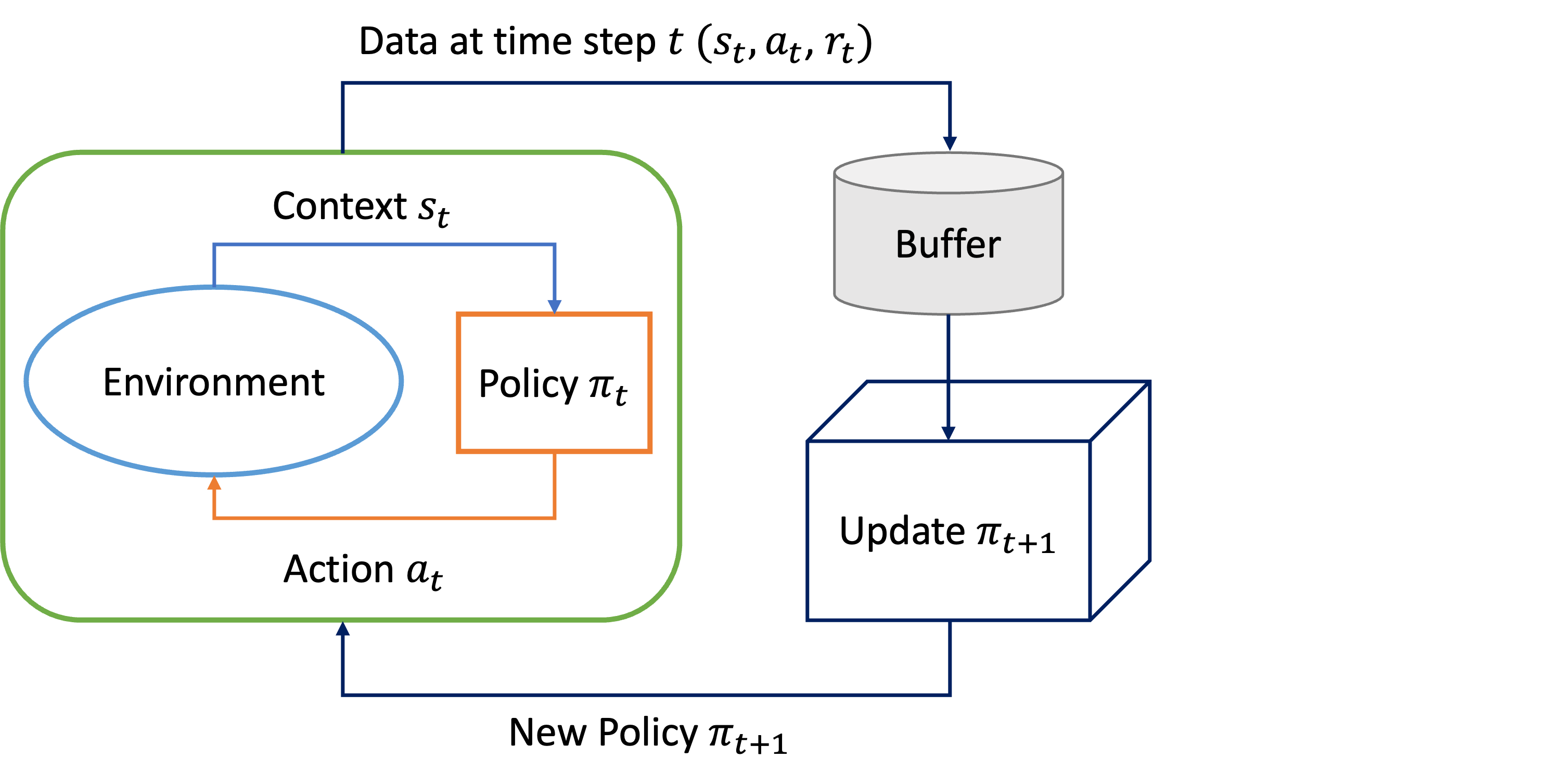}%
            \label{Fig:On_PO}
        }\\
        \subfloat[Offline Policy Evaluation]{%
            \includegraphics[width=.48\linewidth]{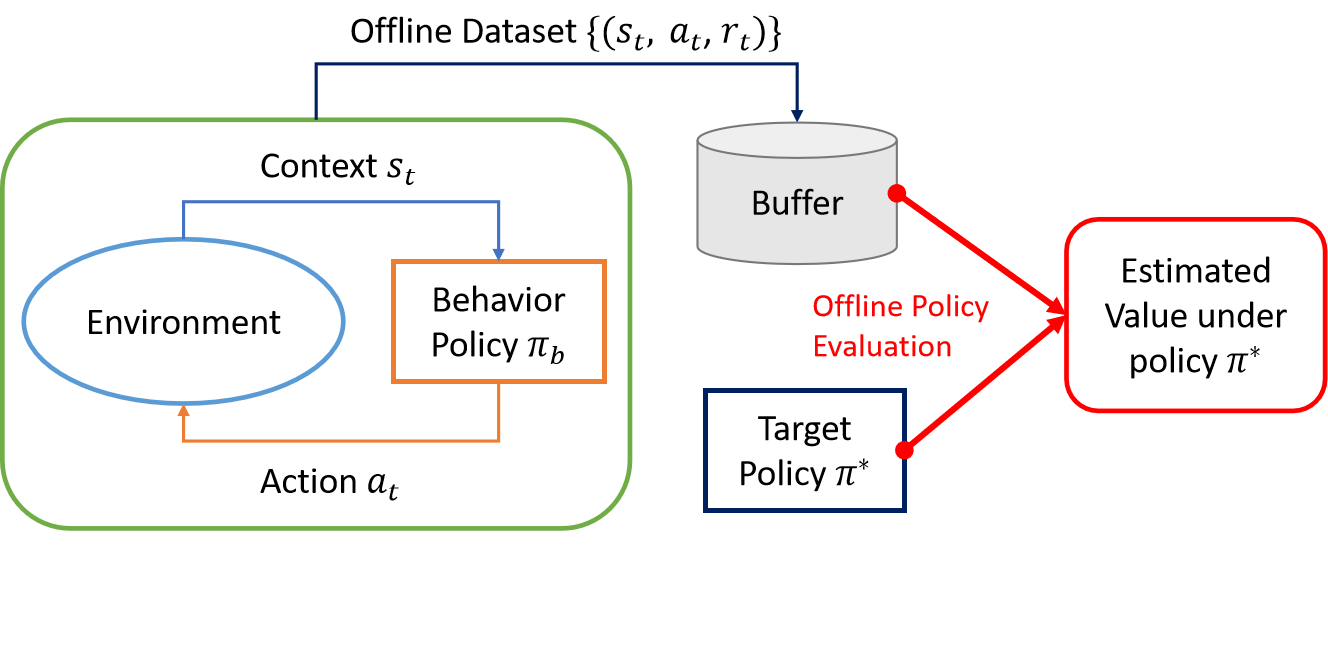}%
            \label{Fig:Off_PE}
        }\hfill
        \subfloat[Online Policy Evaluation]{%
            \includegraphics[width=.48\linewidth]{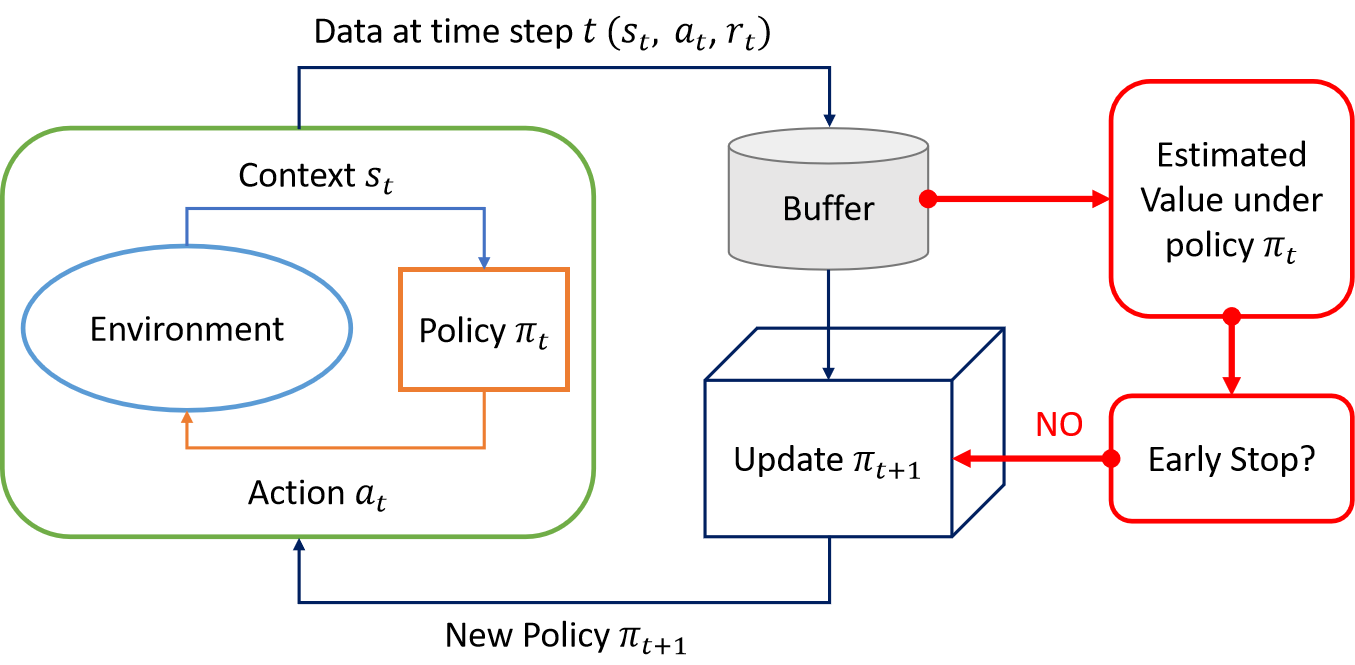}%
            \label{Fig:On_PE}
        }
        
        \caption{Causal Policy Learning}
        \label{Fig:CPL}
\end{figure}

\textbf{\acrlong{CPL}.} 
With the causal structure and effects between variables in mind, the ultimate goal is typically to evaluate and optimize our decision makings. 
When the decision-making is purely based on a fixed historical dataset (i.e., does not involve continuous data collection), we call such a setting \textit{offline} or \textit{off-policy} (see Figures \ref{Fig:Off_PO} 
and \ref{Fig:Off_PE}); 
while when the decision-making process involves continuous data collection and real-time policy updates based on incoming outcomes, we call such a setting \textit{online} (see Figures \ref{Fig:On_PO} 
and \ref{Fig:On_PE}). 
Regardless of the data collection method, two fundamental tasks in \acrshort{CPL} are 
\textit{Policy Evaluation} and \textit{Policy Optimization}. Policy Evaluation \citep{voloshin2019empirical, uehara2022review, ye2023doubly} involves estimating the value of a given \textit{target policy} with respect to a specific state distribution (see Figure \ref{Fig:Off_PE}) or assessing the value of the estimated optimal policy in online learning (see Figure \ref{Fig:On_PE}). Policy Optimization 
\citep{prudencio2023survey, Sergey2020offlineRL, liu2021map,bouneffouf2020survey,silva2022multi,zhou2015survey, shakya2023reinforcement,ladosz2022exploration,wang2022deep,moerland2023model}
focuses on determining the optimal policy that maximizes its value under certain problem-specific requirements (see Figure \ref{Fig:Off_PO}) or identifying the optimal actions that maximize the cumulative rewards during online interactions (see Figure \ref{Fig:On_PO}). In Section \ref{sec:offline_DM}, we review the literature on offline \acrshort{CPL}, while online \acrshort{CPL} is discussed in Section \ref{sec:Online CPL}.



\subsection{Six Paradigms}\label{sec:paradigms}
Regardless of the specific \acrshort{CDM} task, decision-making problems in the literature can be categorized into six paradigms, each capturing common data dependencies typically assumed in practice, as illustrated in Figure \ref{Fig:paradigms} and detailed below.

\textbf{Paradigm 1: Fixed Policy with Independent States.}
As illustrated in Figure \ref{Fig:paradigms}, observations in Paradigm 1 are i.i.d. samples. Each observation consists of three components: $S_i$ is the contextual information (if available), $A_i$ is the action taken, and $R_i$ is the reward received. When contextual information is present, it would influence the choice of the action, while both the contextual information and the action jointly determine the final reward. A classical class of problems that are widely studied in this context is the single-stage \acrshort{DTR} \citep{tsiatis2019dynamic}. Literature on \acrshort{CSL} within this paradigm is discussed in Section \ref{sec:CSL_P1}, while studies on \acrshort{ATE}, \acrshort{HTE}, and mediation effect analysis are reviewed in Section \ref{sec:CEL_p1}. Additionally, this paradigm serves as the main setting in Section \ref{sec:offline_DM} to illustrate offline policy learning methods for evaluating or learning (personalized) policies that aim to maximize the immediate rewards. 

\textbf{Paradigm 2: Fixed Policy with Markovian State Transition.}
The Paradigm 2 is widely recognized as \acrfull{MDP}, characterized by the Markovian state transitions. In particular, while $A_t$ is only affected by $S_t$, both $R_t$ and $S_{t+1}$ would be affected by the pair $(S_t,A_t)$. Given $S_{t}$ and $A_t$, a standard assumption in \acrshort{MDP} problems is that $R_t$ and $S_{t+1}$ are conditionally independent of all previous observations. In Section
\ref{sec:offline_DM}, we also extend the policy evaluation or optimization techniques developed in Paradigm 1 to this setting, where the policy aims to maximize the long-term reward. 

\textbf{Paradigm 3: Fixed Policy with Non-Markovian State Transition.}
Paradigm 3, commonly assumed in multi-stage \acrshort{DTR} problems \citep{tsiatis2019dynamic} and offline non-Markovian \acrshort{RL} problems, considers all possible causal relationships under a history-independent policy. \acrshort{CPL}-related studies within this paradigm are briefly reviewed in Section \ref{sec:offline_DM}. In the context of \acrshort{CEL} within paradigm 3, we primarily focus on a specific panel data setting and discuss effect estimation with respect to evolving time, as detailed in Section \ref{sec:CEL_p3}.

\textbf{Paradigm 4: Adaptive Policy with Independent States.}
Paradigm 4 is extensively studied in the online decision-making literature as bandit problem, where the treatment policy is time-adaptive. In this paradigm, the history $H_{t-1}$, which includes all prior observations up to time $t-1$, is used to update the action policy at time $t$, thereby influencing the decision making of action $A_t$. While $S_t$ is sampled i.i.d. from the corresponding distribution, the reward $R_t$ is influenced by both $A_t$ and $S_t$. The new observation $(S_t,A_t,R_t)$, combined with all previous observations, forms the updated history $H_{t+1}$, which then affects the next action $A_{t+1}$. A common variation of this structure is when the contextual information $S_t$ is absent. Relevant literature on policy optimization under Paradigm 4 is reviewed in Section \ref{sec:bandit_optimization}, while related policy evaluation approaches are discussed in Section \ref{sec:bandit_evaluation}.

\textbf{Paradigm 5: Adaptive Policy with Markovian State Transition.}
Building on Paradigm 4, Paradigm 5 introduces the \acrshort{MDP} framework, with adaptive policy and Markovian state transitions governing the data generation process. Specifically, $S_t$ follows a Markovian state transition, depending only on the most recent state and action, and $A_t$ is determined by the entire observation history $H_{t-1}$ through the dynamically updated action policy. This setup corresponds to the typical online \acrshort{RL} setup, which is reviewed in Section \ref{sec:RL_optimization}.

\textbf{Paradigm 6: Adaptive Policy with Non-Markovian State Transition.}
Paradigm 6 extends Paradigm 5 by relaxing the Markovian assumption, allowing for non-Markovian state transitions. This paradigm encompasses problems such as the \acrfull{POMDP}, with relevant approaches briefly reviewed in Section \ref{sec:RL_optimization}.


\section{Causal Structure Learning}\label{Sec:CSL}

Most existing methodologies for average/heterogeneous treatment effects and personalized decision-making rely on a known causal structure. This enables us to locate the right variables to control (e.g., confounders), to intervene (e.g., treatments), and to optimize (e.g., rewards). However, such convenience is violated in many emerging real applications with unknown causal structures. Causal discovery thus has attracted more and more attention recently, as it allows inferring causal structure from data and disentangling the complex relationship among variables.  In this section, we present state-of-the-art techniques for learning the skeleton of unknown causal relationships among input variables with embedded treatments. We start to detail why \acrshort{CSL} is needed for causal decision-making, then introduce commonly proposed causal graphical models in Paradigm 1 and present existing representative classes of causal discovery methods, followed by discussions and extensions to Paradigms 2 and 3.

\subsection{Why \acrshort{CSL} is Needed for Causal Decision Making}

 
\acrshort{CSL} is a crucial step in understanding the underlying mechanisms that govern changes in a system. It involves identifying the causal relationships between variables, which is fundamental for any subsequent analysis aiming to understand causal effects and make causal decisions. The important reasons for \acrshort{CSL} ahead of \acrshort{CEL} and \acrshort{CPL} can be summarized as follows.




First, \acrshort{CSL} is the \textit{first} step of causal decision-making. More specifically, \acrshort{CSL} is essential for \textit{designing effective interventions and policies} by identifying the exposure or treatment in the causal graph \citep[see e.g.,][]{spirtes2000constructing,chickering2002optimal,shimizu2006linear,kalisch2007estimating,harris2013pc,buhlmann2014cam,ramsey2017million,zhang2018non}. In fields such as epidemiology \citep{hernan2004definition}, medicine \citep{hernan2000marginal},  and economics \citep{panizza2014public}, the underlying causal mechanism among variables of interest is typically unknown. \acrshort{CSL} allows intervention evaluators or policymakers to understand the potential ramifications of their actions by revealing how different factors interact causally.


\begin{figure}[!thp] 
\centering 
 \includegraphics[width=0.32\linewidth]{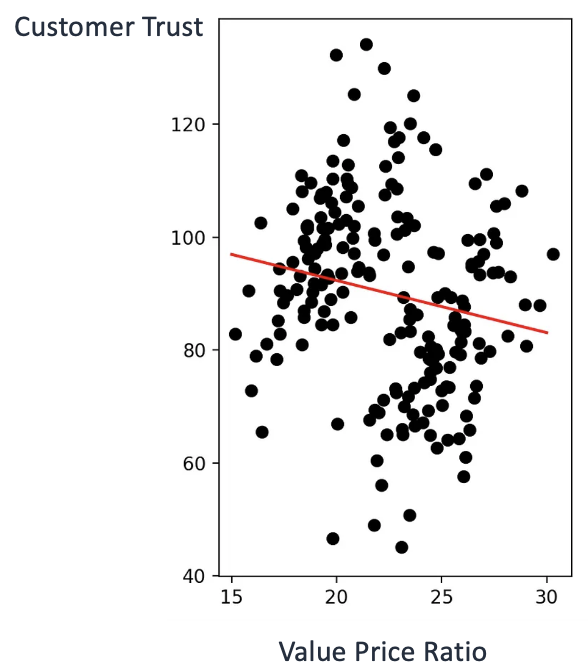}~~~~
 \includegraphics[width=0.3\linewidth]{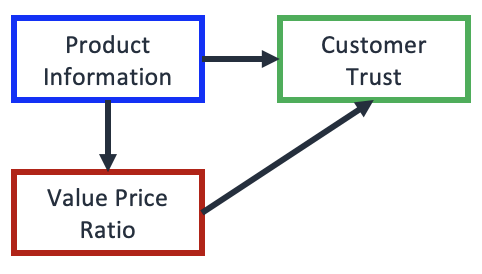} 
 \includegraphics[width=0.32\linewidth]{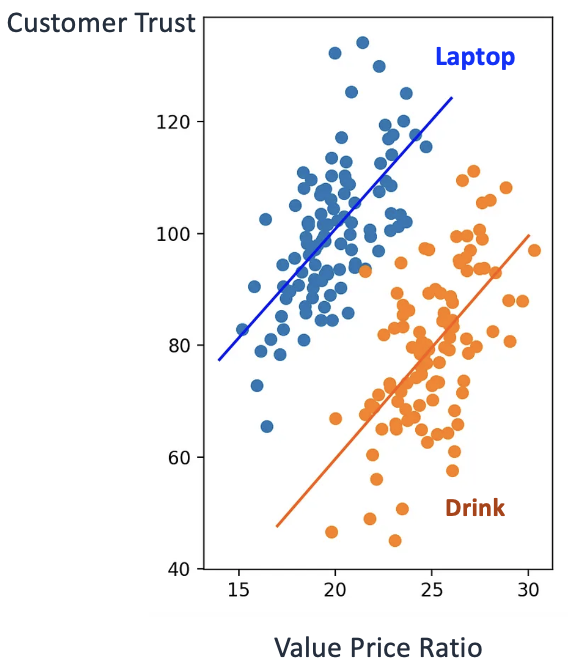} 
 
 \caption{An illustration of Simpson's Paradox.}\label{fig:t-para}
 \end{figure}
 
Second, understanding the causal pathway is essential when estimating the impact of changes in one variable on another to inform decision-making. Interventions based on \textit{incomplete causal knowledge risks yielding biased outcomes}, as depicted in Figure \ref{fig:t-para}. This figure illustrates the complexities of discerning the relationship between the value-price ratio and customer trust. A simplistic regression that ignores confounders suggests a counterintuitive negative correlation: higher value-price ratios correspond to lower customer trust, as shown in the figure's left panel. However, this analysis is flawed due to omitted variable bias. The middle panel of Figure \ref{fig:t-para} introduces a complete causal graph that accounts for potential confounders, offering a more accurate representation of the relationship. When product information is incorporated, the apparent contradiction resolves—higher value-price ratios actually correlate with increased trust in both laptop and drink product categories. This scenario exemplifies Simpson's Paradox \citep{blyth1972simpson}, where aggregated data can mask or reverse trends present within stratified groups.

\begin{figure}[!t] 
\centering 
  \includegraphics[width=0.45\linewidth]{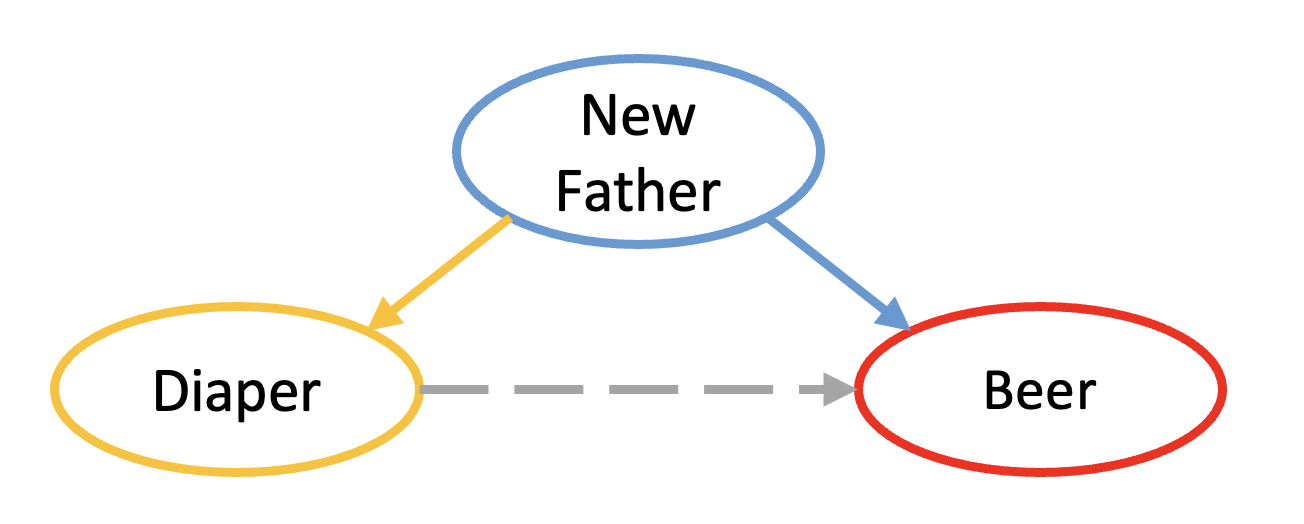} 
  \includegraphics[width=0.45\linewidth]{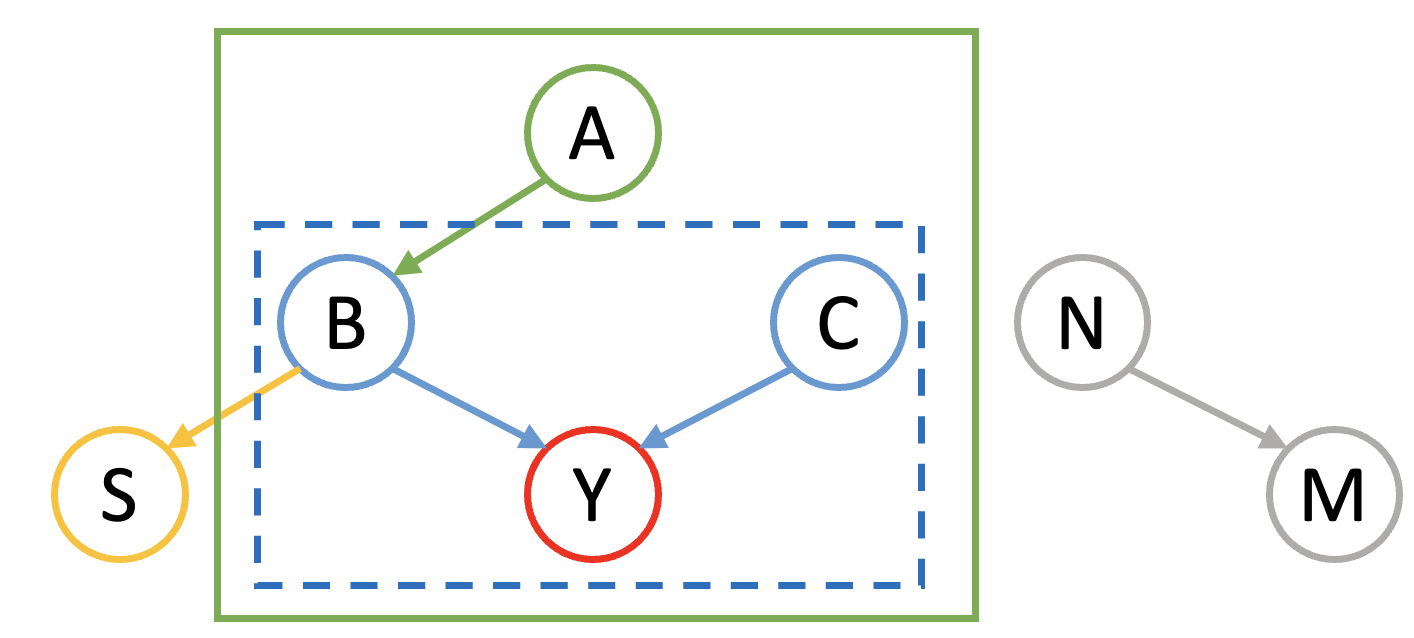} 
\caption{\textbf{Left}: Illustration of the causal relationship between the customer being a new father or not, beer purchasing, and diaper purchasing, where solid lines represent the true model, and the dashed line corresponds to the spurious correlation between beer purchasing and diaper purchasing. \textbf{Right}: Relationship between various causal structures. Nodes $A$, $B$, and $C$ belong to the necessary and sufficient causal graph for the target outcome $Y$ and are depicted inside the green solid square. Among them, nodes $B$ and $C$ are members of the Markov blanket of $Y$, enclosed by the blue dotted square. Node $S$ is the spurious variable to $Y$, while nodes $N$ and $M$ are unrelated to the target.} \label{fig:0}
\end{figure}

Third, with \acrshort{CSL}, we can \textit{avoid spurious relationships}. 
Building upon the causal graphical model \citep[see e.g., ]
[]{pearl2009causal}, many \acrshort{CSL} algorithms have been developed \citep[see e.g.,][]{spirtes2000constructing,chickering2002optimal,shimizu2006linear,kalisch2007estimating,buhlmann2014cam,ramsey2017million,zheng2018dags,yu2019dag,zhu2019causal,cai2020anoce} but rely on the assumption of causal sufficiency (the absence of unmeasured confounders). In real-world applications, to satisfy such an assumption, we strive to learn large-scale causal graphs \citep[see e.g.,][]{nandy2017estimating,chakrabortty2018inference,tang2020long,niu2021counterfactual}, in the hopes of {sufficiently} describing how an outcome of interest depends on its relevant variables.  
In addition to sufficiency, it is also crucial to account for the concept of {necessity} by excluding redundant variables in explaining the outcome of interest. Failure to do so can result in the inclusion of spurious variables in the learned causal graphs, which are highly correlated with but have no causal impact on the outcome. These variables can impede causal estimation with limited data and lead to falsely discovered spurious relationships, leading to poor generalization performance for downstream prediction \citep{scholkopf2021toward}. For instance, it might be observed that men aged 30 to 40 who buy diapers are also likely to purchase beer. However, beer purchase is a spurious feature for diaper purchases: their correlation is not necessarily causal, as both purchases might be confounded by a shared cause, such as new fathers buying diapers for childcare while also buying beer to alleviate stress. Therefore, merely increasing the availability of diapers or beer will not causally enhance the demand for the other (see also Figure \ref{fig:0}(left)).

Fourth, with \acrshort{CSL}, we aim to \textit{simplify complex models} by identifying the most relevant causal relationships for decision making. This simplification can make models more understandable, efficient, and less prone to overfitting. The number of variables causally relevant to the outcome of interest is often considerably smaller than the number of variables included in estimating a causal graph (see  Figure \ref{fig:0}(right)). For example, while an individual's genome may encompass 4 to 5 million \acrfull{SNPs}, only a limited number of non-spurious genes or proteins are found to systematically regulate the expression of the phenotype of interest \citep[e.g.,][]{chakrabortty2018inference}. Similarly, in natural language processing tasks, excluding spurious embeddings such as writing style and dialect can enhance model accuracy and downstream prediction performance \citep[e.g.,][]{feder2021causal}. 

\subsection{Decision-Oriented \acrshort{CSL} under Paradigm 1 }\label{sec:CSL_P1}

This section presents state-of-the-art techniques for learning the skeleton of unknown causal relationships among input variables with the presence of treatments or decision variables under Paradigm 1. 

\subsubsection{Overview of Decision-Oriented Causal Discovery}
Under a general treatment-embedded causal graph, the treatment or exposure may have a direct effect on the outcome and also an indirect effect regulated by a set of mediators (or intermediate variables), confounded by some baseline covariates. 
In the era of the causal revolution, identifying the causal effect of exposure on the outcome of interest is an important problem in many areas \citep[see e.g., ][]{chakrabortty2018inference,cai2020anoce,watson2023heterogeneous}. 
An analysis of causal effects that interprets the causal mechanism contributed through mediators is hence challenging but on demand, and naturally bridged the gap between \acrshort{CSL} and \acrshort{CEL}, and learned results further served as the middle step for \acrshort{CPL}. 

Existing statistical and machine learning tools for learning the causal graphs with multiple mediators \citep[see e.g.,][]{chakrabortty2018inference,cai2020anoce,shi2021testing} 
comprise the following three principal steps. Initially, \acrshort{CSL} methodologies \citep[see e.g.,][]{spirtes2000constructing,chickering2002optimal,nandy2018high,li2019likelihood,yuan2019constrained,li2023inference} are applied to estimate the causal graph, often presented by a \acrshort{DAG}, using observational data. 
With preliminaries of the causal graphical model \citep{pearl2009causal,peters2014identifiability} introduced in Section \ref{sec:assum_csl}, we present three representative classes of causal discovery methods, including testing-based learners \citep{spirtes2000constructing,kalisch2007estimating}, functional-based learners \citep{shimizu2006linear,buhlmann2014cam}, and score-based learners \citep{zheng2018dags,yu2019dag}, in Section \ref{sec:learn_dag}. In the absence of additional assumptions \citep{shimizu2006linear, neal2020introduction}, the graph is only identified up to a \acrfull{MEC}, and a \acrfull{CPDAG} in such a class is often used to represent the graph structure (see details in Section \ref{sec:iden_dag}). 
Following advances in causal mediation analysis \citep{cai2020anoce} with complex graph structure (see details in Section \ref{sec:med_dag}), 
the subsequent step is the estimation of the causal effects of mediators based on the \acrshort{DAG} or \acrshort{CPDAG} obtained from the initial phase. For this task, a variety of estimation techniques have been proposed, including the application of \acrfull{OLS} estimators \citep{vanderweele2014causal,lin2017interventional,chakrabortty2018inference}, parametric models \citep{vanderweele2014mediation,vanderweele2016causal,chen2023discovery}, and nonparametric methods \citep{an2022opening,brand2023recent}. 

\subsubsection{Preliminaries in Decision-Oriented Causal Discovery under Paradigm 1}\label{sec:assum_csl}

As commonly imposed in the works of \acrshort{CSL} \citep[e.g.,][]{spirtes2000causation,peters2014causal}, we assume the Causal Markov and faithfulness 
assumptions. To detail these assumptions, we first introduce the concept of the D-separation.
\begin{definition}[D-separation]\label{def-d-sep}
Nodes, $X$ and $Y$, are d-separated by a set of nodes, $Z$, if and only if for every path, $\pi$, there exists a node, $m\in Z$, that extends $\pi$ ($i\to m\to j$) or forks $\pi$ ($i\xleftarrow{} m\to j$) and for any node, $c$, along $\pi$ that is a so-called collider ($i\to m\xleftarrow{} j$), $c$ and all descendents of $c$ are not in $Z$ \citep{pearl2009causal}.
\end{definition}
Given that $Z$ d-separates $X$ and $Y$ and $X$ preceeds $Y$ causally, the implication of d-seperation is that $X\perp Y|Z$.
\begin{assumption}[Causal Markov assumption]
For a given causal graph, $\mathcal{G} =(\mathbf{Z},\mathbf{E})$, the set of independences among the nodes, $\mathbf{Z}$, contains the set of independences implied by d-separation.
\end{assumption}
\begin{assumption}[Faithfulness assumption]
For a given causal graph, $\mathcal{G} =(\mathbf{Z},\mathbf{E})$, the set of independences among the nodes, $\mathbf{Z}$, is \textbf{exactly} described by the set of independences implied by applying d-separation to $\mathcal{G}$.
\end{assumption}

Note that the assumptions made in this review paper are commonly imposed in the literature of causal inference. Please refer to \cite{pearl2000causality, pearl2009causal, athey2015machine, nandy2017estimating, wager2018estimation, kunzel2019metalearners, nie2021quasi} for discussions of these assumptions and their impact. There are a few future extensions to relax or diagnose these assumptions. For instance, a full sensitivity analysis of the assumptions would be useful to the field when it is hard to include all variables causally related to any variable in the data in practice. In addition, utilizing the instrumental variables in the context of causal graphs with multiple mediators may be beneficial in addressing no unmeasured confounders, as specified below.



\begin{assumption}[Causal Sufficiency assumption]
    The causal graph $\mathcal{G}$ satisfies \textit{Causal Sufficiency} \citep{hasan2023a}. The random vector $\mathbf{X}$ satisfies the structure assumption: (i) No potential mediator is a direct cause of confounders $\mathbf{S}$; (ii) The outcome $R$ has no descendant; (iii) The only parents of treatment $A$ are confounders.
\end{assumption}

In many instances, the accessible data offers an incomplete view of the inherent causal structure.  To address this gap, \textit{Causal Markov Condition}, \textit{Causal Faithfulness Condition}, and \textit{Causal Sufficiency} in the above assumption provide a sufficient condition for causal discovery in i.i.d. data contexts  \citep{lee2020towards,assaad2022survey,hasan2023a}. The rigorous definitions for them and related details can be found in Section 2.4 in \cite{hasan2023a}. Furthermore, the structural assumptions for decision-oriented \acrshort{CSL} aim at ensuring the identifiability of the causal model, which are similar to Consistency Assumption and Sequential Ignorability Assumption in \cite{tchetgen2012semiparametric}, and the structure assumptions in Section 2.4 of \cite{chakrabortty2018inference}.  We introduce some commonly considered causal graphical models as follows under Paradigm 1.

\smallskip

\noindent\textbf{Linear Structural Equation Model.} 
Let $\boldsymbol{B}=\{b_{i,j}\}_{1\leq i\leq d,1\leq j\leq d}$ be a $d\times d$ matrix, where $b_{i,j}$ is the weight of the edge $X_i\rightarrow X_j \in \mathbf{E}$, and $b_{i,j}=0$ otherwise. Then, we say that $\mathcal{G} =(\boldsymbol{X},\mathbf{E})$ is a weighted DAG with the node set $\boldsymbol{X}$ and the weighted adjacency matrix $\boldsymbol{B}$ (the edge set $\mathbf{E}$ is nested in $\boldsymbol{B}$). The \acrfull{LSEM} \citep{sobel1987direct} such that $\boldsymbol{X} = (\boldsymbol{S}^{\top}, A, \boldsymbol{M}^\top, R)^\top$ characterized by the pair ($\mathcal{G}$, $\epsilon$) is generated by 
\begin{equation}\label{lsem_x}
\boldsymbol{X}=\boldsymbol{B}^\top \boldsymbol{X} +\epsilon,
\end{equation}
where $\epsilon $ is a random vector of jointly independent error variables. 
We next explicitly characterize the weighted adjacency matrix  $\boldsymbol{B}$ that satisfies Model (\ref{lsem_x}) based on causal knowledge among $\boldsymbol{S}, A, \boldsymbol{M},$ and $R$, in the decision-oriented \acrshort{CSL}. Specifically, 
the following matrix  $\boldsymbol{B}^\top$ consists of unknown parameters whose  sparsity is due to prior causal information:
\[\boldsymbol{B}^\top = \begin{bmatrix}
\boldsymbol{0}_{p\times p}&\boldsymbol{0}_{p\times 1}&\boldsymbol{0}_{p\times s}&\boldsymbol{0}_{p\times 1}\\
\boldsymbol{\delta_S}&0&\boldsymbol{0}_{1\times s}&0\\
\boldsymbol{B_S}^\top&\boldsymbol{\beta}_A& \boldsymbol{B_M}^\top&\boldsymbol{0}_{s\times 1}\\
\boldsymbol{\gamma_S}&\gamma_A& \boldsymbol{\gamma_M}&0
\end{bmatrix},\]
where $\boldsymbol{0}_{a\times b}$ is a $a\times b$ zero matrix/vector, and the parameters $\boldsymbol{\delta_S},\boldsymbol{B_S}^\top,$ and $\boldsymbol{\gamma_S}$ represent the influence of $\boldsymbol{S}$, on the treatment $A$, the mediators $\boldsymbol{M}$, and the outcome $R$, respectively. Likewise, $\boldsymbol{\beta}_A$ and $\gamma_A$ represent the influence of $A$ on $\boldsymbol{M}$ and $R$, respectively, and $\boldsymbol{\gamma_M}$ represent the influence of $\boldsymbol{M}$ on $R$. $\boldsymbol{B_M}^\top$ represents the influence of the mediators on other mediators. If $\boldsymbol{B_M}^\top = \boldsymbol{0}_{s\times s}$ for the $s$-dimensional mediators, then we say that mediators are \textit{parallel}, otherwise they are \textit{sequentially ordered}. The extension to the \acrshort{LSEM} with the interaction between the possible moderators and the treatment can be found in \citet{watson2023heterogeneous}.

\smallskip

\noindent\textbf{Additive Noise Model.} 
Suppose there exists a weighted DAG $\mathcal{G}=(\boldsymbol{X},\mathbf{E})$ that characterizes the causal relationship among $|\boldsymbol{X}|=d$ nodes in $\boldsymbol{X} = (\boldsymbol{S}^{\top}, A, \boldsymbol{M}^\top, R)^\top$. Each variable $X_i$ is associated with a node $i$ in the \acrshort{DAG} $\mathcal{G}$, and the observed value of $X_i$ is obtained as a function of its parents in the graph plus an independent additive noise $n_i$, as the additive noise model \citep{buhlmann2014cam}, i.e., 
\begin{equation}\label{anm}
X_i := f_i\{PA_{X_i} (\mathcal{G})\} + n_i,i = 1,2,...,d, 
\end{equation}
where $PA_{X_i} (\mathcal{G})$ denotes the set of parent variables of $X_i$ so that there is an edge from $X_j\in PA_{X_i} (\mathcal{G})$ to $X_i$ in the graph, and the noises $n_i$ are assumed to be jointly independent. Here, Model \eqref{lsem_x} is a special case of Model \eqref{anm}. 

\smallskip

\noindent\textbf{Generalized \acrshort{LSEM}.} 
To handle complex relationships, a generalized version of LSEM has been studied \citep{yu2019dag} as
\begin{equation}\label{g_lsem}
f_2(\boldsymbol{X})=B^\top f_2(\boldsymbol{X}) +f_1(\epsilon),
\end{equation}
where the parameterized functions $f_1$ and $f_2$ effectively perform (possibly nonlinear) transforms on $\epsilon$ and $\boldsymbol{X}$, respectively. Here, Model \eqref{lsem_x} is also a special case of Model \eqref{g_lsem}.

\subsubsection{Decision-Oriented Causal Discovery Methods under Paradigm 1}\label{sec:learn_dag}

 In this section, we mainly focused on the decision-oriented \acrshort{CSL} methods under Paradigm 1. Plentiful \acrshort{CSL} methods have been proposed, with the large literature categorized into three types. 
The testing-based methods \citep[e.g.,][for the well-known PC algorithm]{spirtes2000constructing} rely on the conditional independence tests to find the causal skeleton and edge orientations under the linear \acrshort{SEM}. Based on additional model assumptions, the functional-based methods handle both linear \acrshort{SEM} \citep[e.g.,][]{shimizu2006linear} and non-linear \acrshort{SEM} \citep[e.g.,][]{buhlmann2014cam}. Recently, the score-based methods formulate the \acrshort{CSL} problem into optimization by certain score functions, for both linear \acrshort{SEM} \citep[e.g.,][]{ramsey2017million,zheng2018dags} and non-linear \acrshort{SEM} \citep[e.g.,][]{yu2019dag,zhu2019causal,zheng2020learning,rolland2022score}. However, all these methods treat nodes in the graph as generic variables without any causal meaning. In the following, we review a few recent works that learn causal graphs with decision variables oriented.

To start with, we briefly introduce the \acrfull{PC} algorithm \citep{spirtes2000causation}, named by the first two authors, Pater and Clark, 
as one of the oldest testing-based (or constraint-based) algorithms for causal discovery, and the existing decision-oriented \acrshort{CSL} methods based on the \acrfull{PC} algorithm. 
To learn the underlying causal structure, the PC algorithm depends largely on \acrfull{CI} tests. If two variables are statistically independent or conditional independent, there is no causal link between them.  
\citet{maathuis2009estimating} started to use an unknown \acrshort{DAG} without hidden variables to estimate the causal effects from the high-dimensional observational data based on the \acrshort{PC} algorithm. Later, \citet{nandy2017estimating} extended the work of \citet{maathuis2009estimating} with the linear structure equation model. More recently, following these works, \citet{chakrabortty2018inference} firstly introduced the treatment or decision variable into the linear structure equation model, and further 
defined the individual mediation effect. To identify such a causal graph,  \citet{chakrabortty2018inference} fixed the first variable as the treatment or decision and the last variable as the outcome of interest, and then applied the PC algorithm to the rest of the model, i.e., the multiple mediators which influence the outcome but controlled by the treatment. More specifically, their algorithm finds and orients the v-structures or colliders (i.e. $\mathrm{X} \rightarrow \mathrm{Y} \leftarrow \mathrm{Z}$) based on the d-separation set of node pairs (see Definition \ref{def-d-sep}). 
All of these models rely on the PC algorithm to search the Markov equivalence class of the partial \acrshort{DAG}, and usually require strong sparsity and normality assumptions due to the computational limit.

Next, we focus on another type of causal discovery approach, the score-based methods, including greedy equivalence search ~\citep{chickering2002optimal,ramsey2017million,huang2018generalized} and acyclicity optimization methods~\citep{zheng2018dags,yu2019dag,zhu2019causal,lachapelle2019gradient,cai2020anoce,zheng2020learning,vowels2021d}. In the following, we detail a score-based learner, NOTEARS \citep{zheng2018dags} as an example and extend to recent decision-oriented \acrshort{CSL} methods.  \citet{zheng2018dags}  constructed an optimization with an acyclicity constraint under the \acrshort{LSEM}, i.e. the NOTEARS. A follow-up work using a variational autoencoder parameterized by a graph neural network that generalizes \acrshort{LSEM} was proposed in \citet{yu2019dag} with a more computational-friendly constraint, namely DAG-GNN. Also, see \citet{zhu2019causal} and \citet{lachapelle2019gradient} for other cutting-edge score-based structural learning methods. Yet, these methods cannot be directly applied to decision-oriented causal graphs.
To address this challenge, \citet{cai2020anoce}
considered a new constrained structural learning, by incorporating the background knowledge (the temporal causal relationship among variables) into the score-based algorithms. They formulated such prior information as the identification constraint and added it as the penalty term in the objective function for the causal discovery. In the following, we typically detail the NOTEARS for an illustration, which can be easily extended to other score-based algorithms. Specifically, we can write the linear structural model in Equation \eqref{lsem_x} under the causal sufficiency assumption without states as an example as
\begin{eqnarray}\label{lsem}
\begin{bmatrix}

   			A\\
   			\boldsymbol{M}\\
			R\\
		\end{bmatrix}
		=\boldsymbol{B}^\top \begin{bmatrix}
   			A\\
   			\boldsymbol{M}\\
			R\\
		\end{bmatrix}+\epsilon
		=\begin{bmatrix}
   			0& \textbf{0}_{p\times1} &0\\
			\boldsymbol{\alpha}&B_M^\top&0\\
   			\gamma&\boldsymbol{\beta}^\top&0\\
		\end{bmatrix}
		\begin{bmatrix}
   			A\\
   			\boldsymbol{M}\\
			R\\
		\end{bmatrix}
		+\begin{bmatrix}
   			\epsilon_A\\
   			\epsilon_{M_p}\\
			\epsilon_R\\
		\end{bmatrix},
\end{eqnarray}
where $\gamma$ is a scalar, $\boldsymbol{\alpha}$, $\boldsymbol{\beta} $, and $\textbf{0}_{p\times1}$ are $p\times 1$ vectors, $B_M$ is a $p\times p$ matrix, and $\epsilon\equiv [\epsilon_A,\epsilon_{M}^\top, \epsilon_R]^\top $. Here, $\gamma$ presents the weight of the edge $A\rightarrow R$, the $i$-th element of $\boldsymbol{\alpha}$ corresponds to the weight of the edge $A\rightarrow M_i$, and the  $i$-th element of $\boldsymbol{\beta} $ is the weight of the edge $M_i \rightarrow R$. Note that by the causal Sufficiency assumption, we have the exposure $A$ has no parents and the outcome $R$ has no descendants, so equivalently, the first row and the last column of $\boldsymbol{B}^\top$ are all zeros (i.e., the first column and the last row of $\boldsymbol{B}$ are all zeros). 
To estimate the weighted adjacency matrix $B$, 
the score-based learners formulate the acyclicity constraint \citep{yu2019dag,zheng2018dags}  as 
$    h_1(\boldsymbol{B})\equiv \text{tr}\big[(I_{d+1}+t \boldsymbol{B} \circ \boldsymbol{B})^{d+1}\big]-(d+1)=0 $, 
where $I_{d+1}$ is a $d+1$-dimensional identity matrix, and $\text{tr}(\cdot)$ is the trace of a matrix and $t$ is a hyperparameter that depends on the estimated largest eigenvalue of $\boldsymbol{B}$. 
The task of learning DAG is transformed into a constrained optimization problem with the loss by the augmented Lagrangian as
$L(\boldsymbol{B},\theta,\lambda)=f(\boldsymbol{B},\theta)+\lambda h_1(\boldsymbol{B})$, 
where $f(\boldsymbol{B},\theta)$ is some loss such as the least square error in NOTEARS \citep{zheng2018dags} or the Kullback-Leibler divergence in DAG-GNN \citep{yu2019dag} with parameters $\theta$, and $\lambda$ is the Lagrange multiplier.  
Other causal structural leaning algorithms \citep[see e.g.,][]{spirtes2000constructing,chickering2002optimal,shimizu2006linear,kalisch2007estimating,buhlmann2014cam,ramsey2017million,zhu2019causal} can also be applied by formulating the corresponding score or loss function. 

In order for $\boldsymbol{B}$ to satisfy  structural constraints under decision-oriented \acrshort{CSL}, such as $g_1(\boldsymbol{B})$ to $ g_3(\boldsymbol{B})$ (see Section \ref{sec:assum_csl}), it must satisfy: $
h_2(\boldsymbol{B}) =\sum_{i=1}^3 g_i(\boldsymbol{B}) =0.$ As remarked earlier, more structural constraints can be added and any added would be included in $h_2(\boldsymbol{B})$. Combining the two constraints above ($h_1$ and $h_2$) yields the following objective loss by an augmented Lagrangian 
\citep{cai2020anoce}, 
\begin{align*}
    L(\boldsymbol{B},\theta) = f(\boldsymbol{B},\theta) + \lambda_1 h_1(\boldsymbol{B})  +  \lambda_2 h_2(\boldsymbol{B}) + c|h_1(\boldsymbol{B})|^2 + d|h_2(\boldsymbol{B})|^2,
\end{align*}
where model parameter $\theta$, $\lambda_1$ and $\lambda_2$ are Lagrange multipliers, and $c$ and $d$ are tuning parameters to ensure a hard constraint on $h_1$ and $h_2$.

\subsubsection{Model Identifiablities}\label{sec:iden_dag}

In the absence of further assumptions regarding the form of functions and/or noises, the model in \eqref{lsem_x} can only be identified up to \acrshort{MEC} following the Markov and faithful assumptions \citep{spirtes2000constructing,peters2014causal}. Below, we explore the conditions for the unique identifiability of the DAG and potential strategies for addressing scenarios involving the \acrshort{MEC}. More specifically, a general causal \acrshort{DAG}, $\mathcal{G}$, may not be identifiable from the distribution of $\boldsymbol{X}$. According to \cite{pearl2000causality}, a \acrshort{DAG} only encodes conditional independence relationships through the concept of $d$-separation. In general, several \acrshort{DAG}s can encode the same conditional independence relationships, and such \acrshort{DAG}s form a Markov equivalence class. Two \acrshort{DAG}s belong to the same Markov equivalence class if and only if they have the same skeleton and the same v-structures \citep{kalisch2007estimating}. A Markov equivalence class of DAGs can be uniquely represented by a \acrfull{CPDAG} \citep{spirtes2000constructing}, which is a graph that can contain both directed and undirected edges. A \acrshort{CPDAG} satisfies the following: $X_i \leftrightarrow X_j$ in the \acrshort{CPDAG} if the Markov equivalence class contains a \acrshort{DAG} including $X_i \rightarrow X_j$, as well as another \acrshort{DAG} including $X_j \rightarrow X_i$. The Markov equivalence class for a fixed CPDAG $\mathcal{C}$ is denoted by $\operatorname{MEC}(\mathcal{C})$, which is a set containing all \acrshort{DAG}s $\mathcal{G}$ that have the \acrshort{CPDAG} structure $\mathcal{C}$. 
If we can obtain the true DAG from the data, we can simply treat it as a special case of the ``\acrshort{MEC}'' containing only this \acrshort{DAG}, i.e., $\operatorname{MEC}(\mathcal{G})= \{ \mathcal{G} \}$. 

Initially, we consolidate cases where the DAG is uniquely identifiable. In the context of the \acrshort{LSEM}, when the noises $\boldsymbol{\epsilon}$ follow a Gaussian distribution, the resulting model corresponds to the standard linear-Gaussian model class, as investigated in \citet{spirtes2000constructing} and \citet{peters2017elements}. In instances where the noises $\boldsymbol{\epsilon}$ maintain equal variances, according to \citet{peters2014identifiability}, the \acrshort{DAG} $\mathcal{G}$ can be uniquely identified from observational data. Further, when the functions are linear but the noises are non-Gaussian, one can derive the LiNGAM as described in \citet{shimizu2006linear}, where the true \acrshort{DAG} can be uniquely identified under certain favorable conditions. In addition, as cited in \citet{zheng2020learning,rolland2022score}, the nonlinear additive model can be identified from observational data. Another scenario of note arises when the corresponding \acrshort{MEC} encompasses only one \acrshort{DAG}; here, the \acrshort{DAG} can be inherently identified from observational data. Recent score-based causal discovery algorithms \citep{zheng2018dags,yu2019dag,zhu2019causal,cai2020anoce} typically take into account synthetic datasets generated from fully identifiable models, which provides practical relevance in evaluating the estimated graph in relation to the true \acrshort{DAG}.

In instances where the true \acrshort{DAG} is not identifiable, a \acrshort{CPDAG} uniquely symbolizes a MEC of DAGs that yield the same joint distribution of variables. This \acrshort{CPDAG} can be inferred from observational data via a variety of causal discovery algorithms \citep[see e.g.,][]{spirtes2000constructing,chickering2002optimal,shimizu2006linear,kalisch2007estimating,harris2013pc,buhlmann2014cam,ramsey2017million,zhang2018non}. One feasible approach to dealing with \acrshort{MEC} involves enumerating all \acrshort{DAG}s in the \acrshort{MEC} derived from a given \acrshort{CPDAG} \citep{chakrabortty2018inference}. It is conventional to encapsulate a range of potential effects or probabilities by their average or the minimum absolute value \citep{chakrabortty2018inference,shi2021testing}. However, such an approach typically proves computationally prohibitive for large graphs, necessitating computational shortcuts to acquire the causal effects or probabilities of causation without enumerating all \acrshort{DAG}s in the \acrshort{MEC} of the estimated \acrshort{CPDAG}. With the additional identification constraints in the decision-oriented \acrshort{CSL}, the size of \acrshort{MEC} is smaller and thus easier to uniquely identify based on the observational data.

\subsubsection{Decision-Oriented Causal Mediation Analysis}\label{sec:med_dag} 


Causal mediation analysis holds significant importance in causal decision-making, particularly due to its ability to interpret causal mechanisms through mediators. This analysis is challenging yet highly sought after as it effectively bridges the gap between \acrshort{CSL} and \acrshort{CEL}. The integration of causal mediation analysis into decision-making processes enables a deeper understanding of how different variables and interventions interact and influence each other, leading to more informed and effective decisions. Another key motivation behind the use of causal mediation analysis is its role in \acrshort{CPL}. By understanding the pathways through which causal effects are transmitted, policymakers and researchers can develop more nuanced and effective strategies.
Identifying the causality among variables enables us to understand the key factors that influence the target variable, quantify the causal effect of an exposure on the outcome of interest, and use these effects to further guide downstream machine-learning tasks. 





To visualize causes and counterfactuals, \citet{pearl2009causal} proposed to use the causal graphical model and the `do-operator' to quantify the causal effects. A number of follow-up works \citep[e.g.,][]{maathuis2009estimating,nandy2017estimating,chakrabortty2018inference} have been developed recently to estimate direct and indirect causal effects that are regulated by mediators in the linear SEM. These studies relied on the PC algorithm \citep{spirtes2000constructing} which requires strong assumptions of graph sparsity and noise normality due to computational limits. To overcome these difficulties,  \citet{cai2020anoce} proposed to leverage score-based \acrshort{CSL} methods \citep[e.g.,][]{ramsey2017million,zheng2018dags,yu2019dag,zhu2019causal} with background causal knowledge to estimate mediation effects. 
 In the following, we detail the \acrfull{ANOCE} \citep{cai2020anoce}.   
Let $A$ be the exposure/treatment, $\mathbf{M}=[M_1,M_2,\cdots,M_p]^\top $ be mediators with dimension $p$, and $R$ be the outcome of interest. Suppose there exists a weighted \acrshort{DAG} $\mathcal{G}=(\mathbf{Z},B)$ that characterizes the causal relationship among $\mathbf{Z}=[A, \mathbf{M}^\top, R]^\top $, where the dimension of $\mathbf{Z}$ is $d=p+2$.  We next give the \acrlong{TE} ($TE$), the natural \acrlong{DE} that is not mediated by mediators ($DE$), and the natural \acrlong{IE} that is regulated by mediators ($IE$) defined in Pearl (2009).
\begin{equation*}
\begin{split}
TE &={\partial E\{R|do(A=a)\} / \partial a}= E\{R|do(A=a+1)\}-E\{R|do(A=a)\},\\
DE &= E\{R|do(A=a+1, \mathbf{M}=\mathbf{m}^{(a)})\}-E\{R|do(A=a)\},\\
IE &= E\{R|do(A=a, \mathbf{M}=\mathbf{m}^{(a+1)})\}-E\{R|do(A=a)\},
\end{split}
\end{equation*}
where $do(A=a)$ is a mathematical operator to simulate physical interventions that hold $A$ constant as $a$ while keeping the rest of the model unchanged, which corresponds to removing edges into $A$ and replacing $A$ by the constant $a$ in $\mathcal{G}$. Here,  $\mathbf{m}^{(a)}$ is the value of $\mathbf{M}$ if setting $do(A=a)$, and $\mathbf{m}^{(a+1)}$ is the value of $\mathbf{M}$ if setting $do(A=a+1)$. Refer to \citet{pearl2009causal} for more details of `do-operator'. 
First, we will define the natural direct effect of an individual mediator ($DM$). 
\begin{eqnarray*}
    &&DM_i=  \Big[E\{M_i|do(A=a+1)\}-E\{M_i|do(A=a)\}\Big] \\
&&~~~~~~~~\times \Big[E\{R|do(A=a, M_i=m^{(a)}_i+1, \Omega_i=o^{(a)}_i)\}- E\{R|do(A=a)\}\Big], 
\end{eqnarray*}  
where $m^{(a)}_i$ is the value of $ M_i$ when setting $do(A=a)$, $\Omega_i=\mathbf{M}\setminus M_i$ is the set of mediators except $M_i$, and $o^{(a)}_i$ is the value of $\Omega_i$ when setting $do(A=a)$. 
The natural indirect effect for an individual mediator ($IM$) can be defined similarly.
\begin{eqnarray*}
    &&\label{def_IM} 
IM_i= \Big[E\{M_i|do(A=a+1)\}-E\{M_i|do(A=a)\}\Big] \\
&&~~~~~~~~ \times \Big[E\{R|do(A=a, M_i=m^{(a)}_i+1)\}-E\{R|do(A=a, M_i=m^{(a)}_i+1, \Omega_i=o^{(a)}_i)\}\Big]. 
\end{eqnarray*}


Based on the result $TE = DE+ IE$ in Pearl (2009) and the above definitions, we summarize the defined causal effects and their relationship in Table \ref{anoce_table} for the \acrfull{ANOCE}. Firstly, the causal effect of $A$ on $Y$ has two sources, the direct effect from $A$ and the indirect effect via $p$ mediators $\mathbf{M}$ ($M_1,\cdots, M_p$). Next, the direct source has the degree of freedom ($d.f.$) as 1, while the indirect source has $d.f.$ as $p$ from $p$ mediators. Note the true $d.f.$ of the indirect effect may be smaller than $p$, since $A$ may not be regulated by all mediators. Then, the causal effect for the direct source is the $DE$ and for the indirect source is the $IE$, where the $IE$ can be further decomposed into $p$ $DM$s and each component corresponds to the natural direct effect for a specific mediator. The last row in the table shows that the $DE$ and the $IE$ compose the total effect $TE$ with $d.f.$ as $p+1$. 
The ANOCE-CVAE learner \citep{cai2020anoce} is a constrained \acrshort{CSL} method by incorporating a novel identification constraint that specifies the temporal causal relationship of variables. The above decision-oriented causal mediation analysis involves causal structure with possibly multiple mediators under the structural equation model. In Section \ref{sec:sel_med}, we will detail the semi-parametric efficient estimation of mediation effects with a single mediator in the framework of causal effect learning.

\begin{table}[!t]
\centering
\caption{Table of Analysis of Causal Effects (\acrshort{ANOCE} Table).}
\label{anoce_table}
\scalebox{1}{
\begin{tabular}{ccc}
\toprule
Source                  & Degree of freedom    & Causal effects   \\ 
\hline
Direct effect from $A$      & 1           & $DE$           \\
Indirect effect via $M$    & $p$       & $IE$          \\
$\quad \quad \quad   \quad \quad \quad \quad \quad \quad \quad \quad 
	\left\{\begin{array}{ll}
		M_1\\
		M_2\\
		\vdots\\
		M_p\\
	\end{array}
	\right.
$&
$\quad \quad \quad  
	\left\{\begin{array}{ll}
		1\\
		1\\
		\vdots\\
		1\\
	\end{array}
	\right.
$&
$\quad \quad \quad \quad \quad 
	\left\{\begin{array}{ll}
		DM_1\\
		DM_2\\
		\vdots\\
		DM_p\\
	\end{array}
	\right.
$\\
\hline
Total                     & $1+p$           & $TE$          \\
\bottomrule
\end{tabular}}
\end{table}

\subsection{Decision-Oriented Causal Discovery for Paradigm 1+}

Recent advances in causal discovery for time series data have significantly pushed the boundaries of this field to non-i.i.d. settings. Traditional methods like Granger causality \citep{Granger1969} have been supplemented by more sophisticated techniques that address nonlinearity and high dimensionality. The PCMCI algorithm, as developed by \cite{Runge2019}, represents a notable advance, effectively dealing with complex dependencies in time series data. Furthermore, machine learning approaches have shown promise, particularly Gaussian process-based methods for non-linear causal inference \citep{LopezPaz2017}. More recently, the development of deep learning frameworks such as the \acrfull{TCDF} has offered novel insights by identifying cause-effect relationships through attention mechanisms \citep{Nauta2019}. Additionally, the integration of transfer entropy with deep learning models has opened new avenues for understanding causal dynamics in multivariate series \citep{Tank2021}. These innovations have not only improved the accuracy of causal analyses but have also broadened their applicability in real-world scenarios. However, to the best of our knowledge, none of these works consider the treatment or decision in their setting, leaving the decision-oriented causal discovery for Paradigms 2 and 3 still a missing piece in the existing literature.

\section{Causal Effect Learning} \label{sec:CEL}
This section aims to provide a detailed introduction to causal effect learning. We categorize \acrshort{CEL} into three groups based on the underlying causal structure: \acrshort{CEL} with 1) i.i.d. data \citep{kunzel2019metalearners,athey2019generalized}, 2) Markov transition state \citep{liu2018breaking,jiang2016doubly,kallus2022efficiently}, and 3) panel data \citep{viviano2022synthetic,lechner2011estimation}. 
Our aim is to provide a systematic review of estimation techniques for the average treatment effect, the heterogeneous treatment effect, and the mediation effect under the above three scenarios.

In the context of \acrshort{CEL}, we aim to answer the following question:
\begin{center}
    \textit{What is the causal effect of some intervention/treatment/policy? \\If it's well-defined, then how can we quantify it stably and efficiently?}
\end{center}
This question mainly concludes the three main tasks in the realm of \acrshort{CEL}: identification, estimation, and inference. 

Let's take \acrshort{MIMIC-III} data as an example. Once the causal structure of the data (including all potential confounders and mediators) has been determined, the next step is to quantify the effect of intravenous input on the mortality status of patients. This problem involves identifying under reasonable assumptions (such as no latent confounders), estimating the average effect of IV input across all patients, estimating personalized treatment effects on mortality status, and ultimately finding the optimal policy to tailor individualized medical treatments to minimize the overall mortality rate.

Among these four tasks, we are interested in different causal estimands: \acrfull{ATE}, \acrfull{HTE}, and mediation effect. In the next subsections, we will mix the four main tasks and three main estimands together to conduct a comprehensive review according to the data structure of different paradigms.

\subsection{Why Causal Effect Learning is Needed for Causal Decision Making}

\acrshort{CEL} aims to accurately quantify the causal effect of some intervention/policy on a group of units. As an intermediate stage of causal decision-making, \acrshort{CEL} plays an important role in conducting primary analysis on a given causal diagram, as well as providing necessary information for post-stage policy learning and decision-making. The internal connections are multi-fold and detailed below.

First, \acrshort{CEL} provides \textit{primary insights} for decision making.
To answer the question of  ``which policy yields the highest reward or desired outcome within a given population'', a fundamental prerequisite is comprehending how a given policy impacts distinct units within that population in a heterogeneous manner. This challenge, often encapsulated in the estimation of \acrshort{HTE}, aligns with the domain of \acrshort{CEL} perfectly.  
In experimental design, A/B testing is a widely used method in industry to measure the effectiveness of changes or interventions. In observational studies, \acrshort{HTE} estimation (such as $\tau(s) = \mathbb{E}[R(1)- R(0)\mid S = s]$ in a binary action space) can be directly applied to decision-making by selecting the action $\boldsymbol{1}\{\tau(s) > 0\}$. In general, \acrshort{CEL} with observational data provides valuable insights into the effectiveness of specific treatments in a more cost-efficient manner.

Second, \acrshort{CEL} offers \textit{a systematic identification framework} that supports the \textit{validity} of decision-making.
Firstly, while not always stated explicitly, most decision-making methods in \acrshort{RL} rely on certain causal assumptions, such as \acrshort{SUTVA} and \acrshort{NUC}, as outlined in Section \ref{sec:prelim_assump}. These assumptions, though sometimes restrictive, ensure the identifiability of specific value functions, which is essential for conducting valid policy learning. Secondly, in more complex scenarios with interference issues \citep{savje2021average} or unmeasured confounders \citep{wang2018bounded}, \acrshort{CEL} incorporates techniques like instrumental variables or specifying interference structures (see Section \ref{sec:assump_violated}), supporting reliable decision-making based on effect estimates. For example, when assessing whether smoking increases the risk of lung cancer, genetic predisposition serves as a confounder, as it may causally influence both the likelihood of smoking and the risk of developing lung cancer. Properly accounting for this unseen factor is essential to avoid misleading conclusions.
In summary, \acrshort{CEL} acts as a ``safeguard'', formalizing the identification framework to ensure that effect learning remains estimable and that subsequent decision-making is valid.

Third, \acrshort{CEL} \textit{filters out ineffective treatment options with confidence} for better decision making.
Beyond providing value estimates, \acrshort{CEL} also serves as a platform for statistical inference based on the causal effects of interest \citep{mealli2011statistical,benkeser2017doubly,athey2018approximate,xu2022quantile}. For example, consider a decision-making problem in recommending personalized treatments for patients in a clinical trial. While point estimates of the value function for different treatment options reflect the expected outcomes, accurate statistical inference that often bounded within \acrshort{CEL} framework goes further by quantifying uncertainty, which further helps determine the necessity of quasi-control treatment options and simplifies the decision-making process with greater confidence.

\subsection{\acrshort{CEL} in Single Stage (Paradigm 1)}\label{sec:CEL_p1}

Over the past decades, there has been extensive study on conducting causal inference in the classical single-stage setup. Next, we will detail some representative approaches according to the specific tasks they are dealing with: i) \acrshort{ATE}, ii) \acrshort{HTE}, and iii) mediation effect.

\subsubsection{\acrshort{ATE}}

Based on the ``big three assumptions’’ of causal inference (see Assumption \ref{assump:SUTVA}-\ref{assump:Positivity} in Section \ref{sec:prelim_assump}), there are representative \acrshort{ATE} estimation methods in literature, commonly referred to as the \acrfull{DM}, \acrfull{IPW} estimator, and \acrfull{DR} estimator. As we move to later chapters, we will see that the concept of DR estimation is widely applied in various effect estimations, including \acrshort{HTE} and mediation analysis, across both single-stage and infinite-horizon frameworks.

The intricacy of causal inference manifests prominently in the challenge of counterfactual estimation. It is inherently impossible to directly observe the outcomes users would obtain had they chosen differently at the time of treatment assignment. However, failing to observe counterfactuals doesn’t mean that we cannot estimate/infer interested quantities under some reasonable assumptions, which are detailed in Section \ref{sec:prelim_assump}. When \acrshort{SUTVA}, \acrshort{NUC}, and Positivity assumptions hold, the potential outcomes can be rewritten as $\mathbb{E}[R(a)|\boldsymbol{S}=\boldsymbol{s}] = \mathbb{E}[R|\boldsymbol{S}=\boldsymbol{s}, A=a]$, where the right-hand side is entirely estimable from observational data. Therefore, the most intuitive way is to estimate the counterfactual part via a regression model. This yields the first estimator, known as the direct method, as outlined below:
\begin{equation} \label{eq:CEL_p1_DM}
\begin{aligned} \widehat{\text{ATE}}_{\text{DM}} = \frac{1}{n}\sum_{i=1}^n \{\hat{\mu}(\boldsymbol{S_i}, 1) - \hat{\mu}(\boldsymbol{S_i}, 0)\} ,
\end{aligned}
\end{equation}
where $\hat{\mu}(\boldsymbol{s},a)$ is the estimated outcome regression model for $\mathbb{E}[R|\boldsymbol{S}=\boldsymbol{s}, A=a]$.

The second type of estimator is called the \acrfull{IPW} estimator, or \acrfull{IS} estimator in \acrshort{RL} literature. Define propensity score $\mathbb{P}(A=1|\boldsymbol{S})$ as the probability of receivting treatmente $A=1$. \acrshort{IPW} estimator uses propensity scores to reweight observations, balancing the distribution of covariates between the treated and control groups by mimicing a randomized experiment.
\begin{equation} \label{eq:CEL_p1_IS}
\begin{aligned} 
\widehat{\text{ATE}}_{\text{IS}} =\frac{1}{n}\sum_{i=1}^n \left\{\frac{A_iR_i}{\hat\pi(\boldsymbol{S_i})} - \frac{(1-A_i)R_i}{1-\hat\pi(\boldsymbol{S_i})} \right\}. 
\end{aligned} 
\end{equation}

By combining both estimators, the DR estimator (or augmented \acrshort{IPW}, i.e. \acrshort{AIPW}) is consistent as long as either the outcome regression model or the propensity score model is correctly specified. 
\begin{equation} \label{eq:CEL_p1_DR}
\begin{aligned} 
\widehat{\text{ATE}}_{\text{DR}} = \frac{1}{n}\sum_{i=1}^n \left\{\hat{\mu}(\boldsymbol{S_i},1)- \hat{\mu}(\boldsymbol{S_i},0)+\frac{A_i(R_i - \hat{\mu}(\boldsymbol{S_i},1))}{\hat{\pi}(\boldsymbol{S_i})} - \frac{(1-A_i)(R_i-\hat{\mu}(\boldsymbol{S_i},0))}{1-\hat{\pi}(\boldsymbol{S_i})} \right\}. 
\end{aligned} 
\end{equation}

Under certain mild entropy conditions or through sample splitting, the DR estimator is also a semi-parametrically efficient estimator when the convergence rate of both $\hat{\mu}$ and $\hat{\pi}$ are at least $o(n^{-1/4})$. The innovative approach of incorporating sample splitting into treatment effect estimation was conceptually formalized in Double Machine Learning (DML) by \citet{chernozhukov2018double}.

Although the \acrshort{AIPW} estimator guarantees double robustness, it may still result in a poor estimator when both the propensity score and outcome regression models are not correctly specified. In addressing this challenge, an alternative line of research has emerged, focusing on reducing estimation bias via the optimization of model parameters \citep{vermeulen2016data, yang2020doubly}. Additionally, to address estimates that fall outside the admissible parameter range (e.g., a mortality rate outside $[0,1]$), \acrfull{TMLE} \citep{gruber2010targeted, gruber2012tmle} was developed to incrementally adjust the estimator while maintaining the double robustness of \acrshort{AIPW} method.

Under the ideal scenario where the treatment and control groups share similar covariates distributions, the aforementioned estimators are expected to perform well. However, when there is strong selection bias, matching techniques \citep{heckman1998matching, abadie2006large, abadie2008failure, abadie2011bias, caliendo2008some} offer a valuable avenue to improve the performance of the estimation. As the gold standard of causal inference, randomized experiments impose fewer assumptions for identification and estimation. Therefore, the fundamental idea of matching is a way to find the closest ``randomized experiment'' hidden inside the observational study, so as to adjust for covariate imbalances between groups. Under a certain distance metric, one of the most intuitive ways is to select the top $k$ nearest neighbors for each unit, and conduct an average to estimate the corresponding counterfactuals \citep{abadie2008failure}. 

However, in scenarios with a relatively large number of covariates, traditional distance metrics for neighbor selection may encounter challenges due to the curse of dimensionality. To address this, propensity score \citep{rosenbaum1983central, austin2008critical, abadie2016matching} and prognostic score \citep{hansen2008prognostic} are commonly used as two representative balancing scores to conduct dimensionality reduction when adjusting for covariate imbalance. Later on, \acrfull{DS} matching \citep{leacy2014joint, antonelli2018doubly, yang2023multiply} was proposed to jointly combine the above two scores, which further improves the matching performance. This method is double robust in the sense that the DS matching estimator remains consistent for \acrshort{ATE} if either the propensity score or the prognostic score is correctly specified. 
For further practical insights into matching methodologies, refer to \cite{zhang2022practical}.

Overall, the methods of \acrshort{ATE} estimation have been thoroughly studied in literature over the past decades. For other related review papers, we refer to \citet{yao2021survey} under potential outcome's framework and \citet{pearl2009causal} under \acrshort{SCM}. In most decision-making contexts involving a specific population, \acrshort{ATE} estimation plays a crucial role in quantifying the overall impact of different decision rules. This is applied across various domains, including, but not limited to, assessing the effectiveness of advertising campaigns \citep{farahat2012effective}, labor market interventions in public policy \citep{dehejia1999causal}, and epidemiology \citep{hernan2006estimating}. In summary, \acrshort{ATE} estimation serves as a fundamental tool for determining the population or sub-population level effects of different treatments.

\subsubsection{\acrshort{HTE}}\label{sec:CEL_HTE_p1}

In real applications, our focus usually extends beyond the average treatment effect from the population level; rather, the estimation of personalized treatment effects for individuals or within specific subgroups often intrigues our interests. In \acrshort{MIMIC-III} data, our ultimate goal is to figure out the optimal IV input strategy for patients, which often starts with understanding the personalized treatment effect from an individual level.

Existing work in single-stage \acrshort{HTE} estimation starts from meta-learners \citep{kunzel2019metalearners} and subsequently extends to more comprehensive approaches, which either demonstrate improved theoretical properties in statistical inference \citep{kennedy2020optimal} or exhibit enhanced performance in specific settings \citep{nie2021quasi, shi2019adapting}. We will detail some representative methods below.

The first type of learners are meta-learners, which consist of S-learner, T-learner, and X-learner. Under the \acrshort{SUTVA}, \acrshort{NUC}, and positivity assumptions, we have 
\begin{equation*}
    \tau(\boldsymbol{s}) = \mathbb{E}[R(1) - R(0)|\boldsymbol{S}=\boldsymbol{s}] = \mathbb{E}[R|\boldsymbol{S} = \boldsymbol{s}, A=1] - \mathbb{E}[R|\boldsymbol{S} = \boldsymbol{s}, A=0].
\end{equation*}
If we estimate $\mathbb{E}[R|\boldsymbol{S}=\boldsymbol{s}, A=1]$ and $\mathbb{E}[R|\boldsymbol{S}=\boldsymbol{s}, A=0]$ together by fitting $R\sim (\boldsymbol{S},A)$, we obtain what is known as the S-learner. Conversely, if we divide the data into treated and control groups, and fit $\mathbb{E}[R|\boldsymbol{S} = \boldsymbol{s}, A=1]$ and $\mathbb{E}[R|\boldsymbol{S} = \boldsymbol{s}, A=0]$ separately with two independent models, this gives rise to the T-learner. While both learners are straightforward to implement, S-learner tends to exhibit slightly better performance when the treated group and control group share a similar reward structure. Conversely, T-learner may be preferable due to its ability to differentiate action $A$ from all other covariates $X$ in reward modeling. This distinction prevents the risk of neglecting the ``action" among other covariates, which could occur with S-learner. Based on the two base learners, X-learner was proposed by \citet{kunzel2019metalearners}, which shows more favorable performance especially when dealing with sample size imbalance between treatment and control group, or when the separate parts of the X-learner can exploit the structural properties (such as smoothness or sparsity) of the reward and treatment effect functions.

Later on, several additional learners were proposed, including R-learner \citep{nie2021quasi}, DR-learner, and Lp-R-learner \citep{kennedy2020towards}, all following a two-step approach and demonstrating promising theoretical results.  The concept of R-learner originated from \citet{robinson1988root} in 1988 and was formalized by \citet{nie2021quasi} in 2021. R-learner, which stands for ``residual'' learner, is a two-step methodology that involves regressing reward residuals on propensity score residuals, which is able to adapt to various modeling needs and ensure a quasi-oracle property with penalized kernel regression. DR-learner, introduced by \citet{kennedy2020towards}, integrates insights from the \acrshort{DR} estimator to construct an \acrshort{HTE} estimator at the first stage, followed by regression on pseudo outcomes to obtain the final learner. In the same paper, the Lp-R-learner combines residual regression with local polynomial adaptation, employing cross-fitting to relax conditions for achieving the oracle convergence rate. Despite providing promising theoretical results, this algorithm may incur computational intensity when applying local polynomial regressions to a large degree.

Recently, a new stream of work incorporates neural networks in \acrshort{HTE} estimation to provide potentially more flexible modeling choices. The majority of existing work shares a similar two-step pattern in \acrshort{HTE} estimation: In Step 1, the nuisance functions (including propensity score and outcome models) are fitted via some NN-based methods; In Step 2, fitted nuisance functions are combined to estimate the  \acrshort{HTE} via some downstream estimating equation (e.g., plug-in estimator, IPW estimator, or \acrshort{DR} estimator). Notably,  \citet{shi2019adapting} proposed a novel neural network architecture based on the sufficiency of the propensity score for causal estimation in Step 1, and a regularization procedure in Step 2 to optimize nonparametric performance. Similar neural network-based approaches are explored in related works such as \citet{johansson2016learning, shalit2017estimating, hassanpour2019learning}. For a comprehensive review and comparison of these different approaches, we refer to \citet{curth2021nonparametric} for a more detailed review.

In many application scenarios, effect learning is often a crucial pre-step before making decisions. The relationship between \acrshort{HTE} and decision-making can be as simple as $\boldsymbol{1}\{\tau(\boldsymbol{s}) > 0\}$, or it can be adapted to more realistic concerns such as resource or budget constraints. For example, decisions can be made by selecting treatments for patients whose predicted treatment effect exceeds a decision threshold \citep{dorresteijn2011estimating} in clinical trials. Alternatively, decision-making may involve a more complex function of \acrshort{HTE}, incorporating factors such as costs and pricing associated with actions \citep{miller2020personalized}, or serve as an intermediate step feeding into downstream optimization tasks under resource constraints \citep{qiu2022individualized}. The flexibility of the \acrshort{HTE} methods introduced above allows decision-makers to select the most appropriate approach based on their specific needs.

\subsubsection{Mediation Effect}\label{sec:sel_med}

Mediators are variables that are causally affected by action $A$ and, in turn, influence the reward modeling of $R$. They create an additional causal pathway from action to reward, which is often considered when analyzing complex causal relationships. In Section \ref{sec:med_dag}, we discussed several key methods in \acrshort{CSL} for identifying mediators in causal graphs, with particular attention to recent advances in score-based approaches \citep{cai2020anoce} that address mediator identification and effect learning simultaneously. While it is great to kill two birds with one stone, this type of approach may suffer from potential limitations in modeling such as linearity. 

In this section, we shift our focus to \acrshort{CEL}, specifically in the context of single mediator when the causal structure is already known. Mediation effect estimation approaches can be roughly divided into three categories: (1) classical approaches based on parametric modeling, (2) non-parametric and semi-parametric causal mediation analysis, and (3) later extensions. Compared to the methods discussed in Section \ref{sec:med_dag}, the more recent techniques emphasize flexible modeling, which can be particularly advantageous when focusing solely on effect learning with a known mediation structure.

Causal mediation analysis is well-developed in single-stage settings. Recently, there has been a growing body of work focused on extending these approaches to \acrshort{MDP}s and other data structures, such as \acrshort{DTR}. Here, we will briefly summarize the main approaches in paradigm 1 and refer to some review papers for further reading.

\begin{figure}[tbh]
    \centering
    \includegraphics[width = 0.75\linewidth]{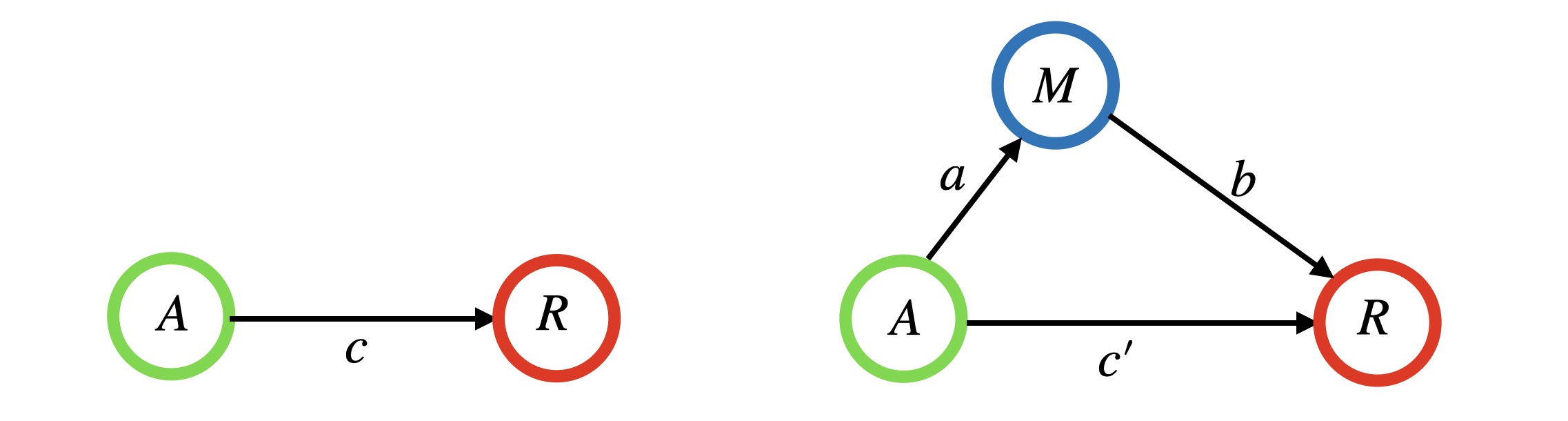}
    \caption{Mediation effect}
    \label{fig:mediation_1}
\end{figure}

We will start by introducing the classical approaches \citep{mackinnon2007mediation}. In the presence of a mediator $M$, the causal relationship between action $A$ and reward $R$ is illustrated in Figure \ref{fig:mediation_1}. Classical approaches focus on decomposing the strength of the causal paths using three parametric models:
\begin{equation}\label{eq:mediation_2}
    \begin{aligned}
        R &= \beta_1 + cA + \epsilon_1\\
        R & = \beta_2 + c'A + bM +\epsilon_2\\
        M & = \beta_3 + aA +\epsilon_3,
    \end{aligned}
\end{equation}
where the coefficients $(a,b,c,c')$ correspond to the strength of the causal relationships depicted in Figure \ref{fig:mediation_1}. There are three main approaches based on these equations, which are (i) causal steps \citep{baron1986moderator}, (ii) difference in coefficients \citep{mackinnon1993estimating}, and (iii) product of coefficients \citep{alwin1975decomposition}. The causal step approach is a four-step regression-based procedure to decompose the significance and strength of different paths in Figure \ref{fig:mediation_1}. The difference in coefficients approach approximates the mediated effect by calculating $\hat{c}-\hat{c}'$, while the product of the coefficients approach estimates the mediated effect by $\hat{a}\hat{b}$. The last two estimators are equivalent when modeling with linear regression. All three approaches are widely used in practical applications due to their simplicity and interpretability. Later on, there are some follow-up reviews under the \acrfull{LSEM} \citep{hayes2017introduction,bollen1987total,imai2010identification,mackinnon2002comparison,pearl2022direct} that allows to measure the causal relationships between multiple variables in a more flexible way. However, these approaches also may suffer from the drawbacks of parametric assumptions such as linearity.

The second type of estimator is based on more recent studies on causal mediation analysis under the potential outcomes framework \citep{imai2010general}. In recent years, extensive work has focused on non-parametric identification and non(semi)-parametric estimation and inference of mediated direct and indirect effects \citep{tchetgen2012semiparametric}. Similar to \acrshort{ATE} estimation, researchers have proposed corresponding versions of \acrshort{DM} \cite{imai2010general}, \acrshort{IPW} \citep{hong2010ratio}, and \acrshort{DR} \citep{tchetgen2012semiparametric} estimators for estimating direct and indirect effects in mediation analysis. Notably, the \acrshort{DR} mediation effect estimands proposed by \citet{tchetgen2012semiparametric} achieve semi-parametric efficiency.



The third stream of work focuses on extending certain NUC assumptions or modeling requirements. For instance, some studies handle binary mediators using principal stratification \citep{rubin2004direct, vanderweele2008simple, gallop2009mediation}. Others relax the linear assumptions in LSEM by employing alternative regression models \cite{mackinnon2007intermediate} or by incorporating exposure-by-covariate and mediator-by-covariate interactions \citep{hayes2017introduction}. Additionally, some work allows for the presence of specific types of confounders \citep{vanderweele2009conceptual, vanderweele2015explanation}. Recently, there has been some work to handle mediation effect estimation in reinforcement learning (paradigm 2) \citep{ge2023reinforcement} and dynamic treatment regimes (paradigm 3) \citep{selig2009mediation, zheng2017longitudinal, roth2013mediation}. These approaches are gaining increasing attention for their flexibility in handling various multi-stage decision-making scenarios. For more detailed discussions on specific modeling and assumptions, please refer to these review papers \citep{hayes2017introduction, ten2012review, rijnhart2021mediation, preacher2015advances}.

\subsection{\acrshort{CEL} under \acrshort{MDP} (Paradigm 2)}
In some real cases, researchers may encounter longitudinal data with long horizons or even infinite horizons. For example, in clinical trials, the doctor will periodically check the health status of patients to provide them with personalized treatment each time they visit. Under this scenario, we aim to estimate the long-term causal effect of taking a specific treatment across all stages. 
Under the Markovian assumption, this problem is referred to as causal effect estimation within an \acrshort{MDP} framework, as detailed in Definition \ref{def:MDP}. 

Unlike the single-stage setting where much work focuses on estimating the difference in potential outcomes under $A = 1$ and $A=0$, the definition of causal effect becomes more general in multi-stage settings after introducing the concept of \textit{policy}. As defined in Definition \ref{def:policy}, a policy $\pi: \mathcal{S}\rightarrow \mathcal{A}$ is a mapping from state to action space that quantifies the treatment assignment strategy for different actions in $\mathcal{A}$. 
The causal effect estimation problem is thus generalized to estimating the state-value function $V^{\pi}(s) = \sum_{t'=t}^T \mathbb{E}^{\pi}[\gamma^{t-t'}R_{t'}|S_t = s]$, a discounted cumulative reward aggregated under policy $\pi$. Following similar logic in single stage, we can still define  \acrshort{HTE} and \acrshort{ATE} under \acrshort{MDP} or any other multi-stage settings as the difference in the value function under two policies ($\pi$ and $\pi_0$), i.e.,
\begin{equation*}
    \text{HTE}(s) = V^{\pi}(s) -  V^{\pi_0}(s), \;\;\; \text{and} \;\;\;
    \text{ATE} =\mathbb{E}_{s\sim \mathcal{S}}\Big[ V^{\pi}(s) -  V^{\pi_0}(s) \Big].
\end{equation*}

While we can naively estimate the quantities using techniques in single-stage \acrshort{CEL} by regarding $\pi$ and $\pi_0$ as two treatments, this approach overlooks the unique Markovian structure in state transitions, resulting in less efficient estimates.
Instead, by leveraging the sequential decision-making structure and the Markovian assumptions, we can derive estimators that converge much faster. 

By involving the definition of \textit{policy}, the problem of \acrshort{HTE} and \acrshort{ATE} estimation can be regarded as a direct byproduct of conducting policy evaluation on the value function $V^{\pi}(s)$ \citep{tang2022reinforcement,shi2023dynamic}.
In the RL literature, this is widely known as \acrfull{OPE}. As this part highly overlaps with \acrshort{CPL} in paradigm 2, we will leave the main discussion to Section \ref{sec:OPE}.

\subsection{\acrshort{CEL} in Panel Data (Paradigm 1)}\label{sec:CEL_p3}
Panel data analysis examines data collected over time from the same individuals, companies, or entities (known as \textit{panels}), often under varying treatment conditions. This approach is commonly used by governments and organizations to assess the long-term effects of policies on outcomes such as income, health, and education. By leveraging longitudinal data, panel analysis supports informed decision-making and provides valuable insights into the lasting impacts of policy interventions. The traditional literature in panel data analysis primarily focuses on estimating the \acrfull{ATT}, defined as the expected difference in outcomes between treated and control units:
\begin{equation}
\mathbb{E}[R_{i,t}(1)-R_{i,t}(0)|G_i =1],
\end{equation}
where $G_i \in \{0,1\}$ indicates whether unit $i$ is in the treatment group ($G_i=1$) or the control group ($G_i=0$). Different from single-stage \acrshort{ATE}/\acrshort{HTE} estimation, panel data analysis aims to quantify the change of causal effect over time. Based on SUTVA, a classical assumption in causal inference, \acrshort{ATT} can be identified from observed data. Since $R_{i,t}$ can be observed to unbiasedly estimate $\mathbb{E}[R_{t}(1)|G_i=1]$, the main challenge of panel data analysis is to impute the missing values for $R_{i,t}(0)$ for treated units. That is, we would like to answer the question of 

\textit{``What would happen to the treated units if they were exposed to control back to the treatment time?”}

Since the true answer is unobservable to us, we need to rely on additional assumptions to leverage existing information, particularly control units, for counterfactual estimation. 
To address this problem, there are two main streams of work in literature: \acrfull{DiD} \citep{lechner2011estimation} and \acrfull{SC} \citep{abadie2003economic, li2020statistical}. \acrshort{DiD} approach relies on the parallel trend assumption, where we learn the change of causal effect over time for control units and apply them to treated units for counterfactual estimation. Conversely, \acrshort{SC} approximates each treated unit with a weighted combination of controls, so as to borrow this weighted information for estimation. To be clearer, \acrshort{DiD} borrows the information of control units over time and inherits it to treated units; while \acrshort{SC} borrows the information of pre-treated stage over units and inherits it to post-treatment time. Due to the difference in estimation strategy, \acrshort{DiD} is often used when we are willing to assume the parallel trend assumption, while \acrshort{SC} is often applied to cases where only a few units are exposed to treatment.

Later, \citet{athey2021matrix} proposed a unified approach to integrate \acrshort{DiD} and \acrshort{SC} using matrix completion. Unlike \acrshort{DiD} and \acrshort{SC}, which rely on specific parallel or orthogonal assumptions, \citet{athey2021matrix} reframed causal effect estimation as a missing data imputation problem, assuming a low-rank structure to estimate counterfactuals in $\boldsymbol{R}(0)$. Recent advancements include but not limited to (1) R-DiD \citep{nie2019nonparametric}, which extends classical DiD by relaxing linear functional assumptions to accommodate more flexible estimands, (2) Synthetic DiD (SDiD) \citep{arkhangelsky2021synthetic}, combining the benefits of DiD and SC by re-weighting and matching pre-exposure trends to mitigate parallel assumptions while remaining invariant to additive unit-level shifts, (3) Synthetic Learner \citep{viviano2023synthetic}, an ensemble method enhancing precision through model-free inference, and (4) H1SL and H2SL \citet{shen2022heterogeneous}, which enable  \acrshort{HTE} estimation in panel data with one-sided and two-sided synthetic learners.

This field is rapidly evolving, offering greater flexibility in modeling choices and relaxing assumptions. 
Classical literature in panel data analysis primarily emphasizes effect estimation under relatively simple decision choices, often by examining patterns of change before and after treatment assignment, with less focus on directly modeling policy learning. Recent work, such as \citet{harris2024strategyproof}, introduces a strategy-proof framework for policy learning that maps pre-treatment outcomes to various intervention choices. This advancement has helped define policy learning explicitly within the context of panel data. For an in-depth overview, see the recent review by \citet{hsiao2022analysis} and \citet{arkhangelsky2023causal}.

\section{Offline \acrlong{CPL}}
\label{sec:offline_DM}

This section presents policy evaluation and optimization methods for offline/off-policy settings (i.e., Paradigms 1-3). In contrast to online policy learning, additional data collection is infeasible in the offline setting, resulting in distribution shifts across multiple dimensions—particularly in actions and states—a critical challenge. These shifts 1) introduce selection bias that necessitates causal adjustments as employed in \acrshort{CEL}, and 2) increase uncertainty in policy evaluation and hence optimization, requiring a pessimistic or penalty-based approach to avoid over-optimization. 

We begin with formal definitions of the tasks in \acrshort{CPL}. 
We use Paradigm 1 to illustrate, where the observed data consists of $n$ data points $\{(S_{i},A_{i},R_{i})\}_{1\le i\le n}$. The dataset is collected by following a stationary policy $\pi_b$, known as the \textit{behavior policy}. 
We study two tasks in offline \acrshort{CPL}:
\begin{itemize}
    \item \textbf{\acrfull{OPE}}: The goal of \acrshort{OPE} is to estimate the goodness of a given \textit{target policy} $\pi$, which is typically evaluated by the integrated value 
$\eta^{\pi} =  \mathbb{E}_{s \sim \mathbb{G}} V^{\pi}(s)$ with respect to some state distribution $\mathbb{G}$. 
    \item \textbf{\acrfull{OPO}}: The goal of \acrshort{OPO} is to solve the optimal policy $\pi^*$, or in other words, to learn a policy $\hat{\pi}$ so as to minimize the regret $\eta^{\pi^*} - \eta^{\hat{\pi}}$. 
\end{itemize}

\subsection{Offline Policy Evaluation}\label{sec:OPE}
OPE focuses on estimating the expected reward of an evaluation policy using historical data generated by a different behavior policy. This is particularly valuable in offline RL settings, where experimenting with policies is not possible due to ethical, financial, or safety concerns. OPE methods have gained importance across fields, including healthcare, education, and recommendation systems, where reliable evaluation of new policies without online testing is critical.

\textbf{Model-based estimators.} Model-based \acrshort{OPE} \citep{paduraru2013off, yin2020asymptotically} approaches estimate state transition and reward functions directly from data, which can then be used to simulate trajectories and estimate policy value. These methods achieve asymptotic efficiency in discrete \acrshort{MDP}s. 
Such estimators often leverage probabilistic neural networks to model transitions, improving performance in complex continuous control tasks. Although model-based methods allow for easier parameter tuning, particularly through supervised learning techniques, they can struggle in high-dimensional settings where modeling state transitions becomes more complex than direct estimation of value functions.

\textbf{Model-free estimators.}
To adjust for selection bias caused by the distribution shift in offline dataset, the most popular methods include \acrshort{DM}, \acrshort{IPW}, and \acrshort{DR} estimators. 
The classic forms of these methods can be derived from the following relationship: 
\begin{align}
    \eta(\pi) 
    = \mathbb{E}_{a \sim \pi(\cdot|S), s \sim p(S) } R(a) 
    &= \mathbb{E}_{a \sim \pi(S), s \sim p(S)} \Big[\mathbb{E} \big\{R(a)|S = s \big\}\Big]\label{eqn:ope_1}\\
    &= \mathbb{E}_{a \sim b(S), s \sim p(S)} \frac{\pi(a|s)}{b(a|s)} R(a)\label{eqn:ope_2}\\
    &= \mathbb{E}_{a \sim b(S), s \sim p(S)} \frac{\pi(a|s)}{b(a|s)} \Big[R(a) - V(s)\Big] + \mathbb{E}_{s \sim p(S)} V(s)\label{eqn:ope_3}. 
\end{align}
Specifically, by replacing $\mathbb{E} \big\{R(a)|S = s \big\} = Q(s, a)$ in \eqref{eqn:ope_1} with its estimator, we obtain the direct method estimator; 
by replacing the expectation over $R(a)$ in \eqref{eqn:ope_2} with a sample average of the observed rewards under action $a$, we obtain the \acrshort{IPW} estimator; 
and finally we can combine these two approaches in \eqref{eqn:ope_3} to derive the \acrshort{DR} estimator. 
Notably, it is easy to see that these three methods are direct extensions of their counterparts in \acrshort{CEL} (Equations \eqref{eq:CEL_p1_DM}, \eqref{eq:CEL_p1_IS} and 
 \eqref{eq:CEL_p1_DR}), by taking additional expectation over the state and action distributions. 
Similar to the argument in \acrshort{CEL}, these methods effectively employ different ways to adjust for the confounding effect from $s$, by either removing its imbalance across actions or its impact on the rewards. 
%


\textbf{Extensions to paradigms 2 and 3.}
These methods can all be extended to more complicated settings in Paradigms 2 and 3, by additionally accounting for the dependency over decision points. 
To simplify the problem in Paradigm 2, we can utilize the \textit{recursive} or \textit{iterative} structure. 
Take the direct method as an example, where as long as we can obtain an estimate of the Q-function, we can directly take its expectation to calculate the value as in \eqref{eqn:ope_1}. 
To estimate the $Q$-function, we introduce two prominent approaches. 
The most straightforward method, \acrfull{FQE} \citep{le2019batch}, leverages the Bellman Optimality Equation \eqref{eqn:bellman_Q} which characterizes the sequential dependency structure. 
\acrshort{FQE} solves the loss function corresponding to \eqref{eqn:bellman_Q} until we converge to a final value function estimator. 
Another method is Minimax Q-Learning \citep{uehara2020minimax}, which enhances Q-function evaluation by framing it as a competition between two components: the Q-function itself and a discriminator function. This method leverages the Bellman equation, where the discriminator is introduced to assess differences between the predicted and actual rewards, guiding the learning process to focus on areas of high prediction error. By balancing estimation errors with a tuning parameter and carefully choosing model classes, the approach becomes robust against specific data patterns and high variance.


The \acrshort{IPW} and \acrshort{DR} methods can similarly be extended to Paradigms 2 and 3. For example, for \acrshort{IPW}, we can replace the density ratio by that along the entire trajectory $\prod_{t'=0}^{t} [\pi(A_{i,t'}|S_{i,t'}) /b(A_{i,t'}|S_{i,t'})].$ 
However, these traditional \acrshort{IS} methods (and related \acrshort{DR} methods) have exponential variance with the number of steps and hence will soon become unstable when the trajectory is long.  
To avoid this issue, various structural assumptions have been utilized. 
One notable advance is by considering the marginal importance ratio under stationary assumptions \citep{liu2018breaking, dai2020coindice, shi2021deeply, zhu2023robust}, where essentially we consider the average density of visiting a state instead of considering the different densities at different time points, which allows us to greatly reduce the problem dimension.  
Similar techniques have been extended to the \acrshort{DR} estimator as well \citep{jiang2016doubly}, notably the Double Reinforcement Learning method \citep{kallus2022efficiently}

\subsection{Offline Policy Optimization}\label{sec:OPO}

Another central task is to learn a good policy from the offline data. 
Formally, we want to find
$\pi^* = \arg \max_{\pi} \eta\big(\pi\big). $
Similar to OPE, one key challenge is to adjust the selection bias caused by the distribution shift in offline data, therefore we will see similar tools in \acrshort{OPE} (and also \acrshort{CEL}) are extended here. 
We will additionally consider the unique goal of policy learning itself to design the so-called \textit{pessimism}-based algorithms.


\subsubsection{Model-Free Value-based Approches}
The first class of algorithms is \textit{value}-based, focusing on utilizing the value function. \textit{Q-learning} is, arguably, the most popular algorithm due to its simplicity and good performance. Noting that 
$\pi^*(\boldsymbol{s}) =  \text{arg max}_{a}Q(\boldsymbol{s}, a),$ 
the core of Q-learning is a regression modeling problem based on positing regression models for outcome. 
Overall, Q-learning is practical and easy to understand, as it allows straightforward implementation of diverse established regression methods. 
Different Q-function model classes (such as linear models, sparse linear models, neural networks, etc.) and their statistical properties have been studied extensively in the literature \citep{song2015penalized, zhu2019proper}.

In some cases, Q-learning could be overkill for policy optimization: for decision making what we need to know is which action is the best, which could be similar to knowing the expected potential outcome of all actions. In such cases, Advantage-learning (A-learning) \citep{murphy2003optimal, robins2004optimal, schulte2014q} offers a more efficient alternative by modeling only the contrasts between treatments and a control action, as 
$
Q(\boldsymbol{s}, a) = Q(\boldsymbol{s},0) + A(\boldsymbol{s}, a). 
$
With the A-function $A(\boldsymbol{s}, a)$, we have that
$
\pi^*(\boldsymbol{s}) =  \text{arg max}_{a \neq 0}A(\boldsymbol{s}, a) \mathbb{I}(\text{max}_{a \neq 0}A(\boldsymbol{s}, a) > 0).
$
Similar to Q-learning, various regression functions can be used to specify the advantage function. Typically, the underlying relationship in the advantage functions is simpler than that in $Q(\boldsymbol{s},0)$, which is a \textit{nuisance} function in decision making. 
The extension of A-learning to high-dimensional \citep{shi2018high} and non-parametric models \citep{liang2018deep} have also been studied.

\textbf{Extensions to paradigms 2 and 3.}
These approaches can be similarly extended to MDP and non-Markovian decision processes. 
For example, the fitted-Q iteration \citep{ernst2005tree} extends the single-period Q-learning, by noting that 
the optimal value function $Q^*$ is the unique solution to the Bellman optimality equation \eqref{eqn:bellman_Q}. 
Additionally, the right-hand side of \eqref{eqn:bellman_Q} is a contraction mapping, allowing us to consider a fixed-point method similar to fitted-Q evaluation. 
A-learning is also recently extended to the \acrshort{MDP} setting \citep{shi2024statistically}. 
Extending to non-Markovian problems, such as \acrshort{DTR} in precision medicine and decision science,  
we can estimate the optimal dynamic treatment regimes (policy) via G-estimation \citep{stephens2015chapter, robins2004proceedings}, a type of A-learning \citep{schulte2014q, shi2018high}. 
Q-learning can also be extended to non-markovian problems via recursive regression \citep{song2015penalized}. 
The main challenge comes from the increasing dimensionality with the expanding horizon, as the loss of the Markovian property requires us to use the full history in the feature space instead of only the latest state variable.

\subsubsection{Model-Free Policy-based Approaches}
For interpretability, domain constraints, or statistical efficiency, it may be preferable to directly learn a policy $\pi$ within a pre-specified (parametric or non-parametric) policy class $\Pi$ as 
$\pi^* = \argmax_{\pi \in \Pi} \eta(\pi), $
where the policy value $\eta(\cdot)$ can be estimated via various \acrshort{OPE} methods discussed above. This estimated value is then incorporated into an optimization process to solve the $\argmax$ via off-the-self \textit{optimization algorithms} (such as the L-BFGS-B) \citep{kitagawa2018should, zhao2012estimating, liu2018augmented}. 

In particular, when we use \acrshort{IPW} as the policy value estimator, we can re-write the objective as 
\[
    \pi^* = 
    \argmax_{\pi \in \Pi} \mathbb{E} \Big[ \frac{R_i}{b(A_i | S_i)} \mathbb{I}(A_i \neq \pi(S_i) )\Big]. 
\]
When $R_i$ is non-negative, this goal corresponds to the objective function of a cost-sensitive classification problem 
with ${R_i}/{b(A_i|S_i)}$ as the weight, 
$A_i$ as the true label, 
and $\pi$ as the classifier to be learned. 
Then, any popular cost-sensitive classifiers, such as \acrfull{SVM} and \acrfull{CART}, can be applied to solve the policy learning problem. 
This is called outcome-weighted learning \citep{zhao2012estimating, liu2018augmented, song2015sparse}, providing flexibility in high-dimensional and complex scenarios. 
Furthermore, this framework can be extended to incorporate the \acrshort{DR} estimator, enhancing robustness against misspecifications of the propensity score model or the outcome model.

Decision lists and tree-based structures are interpretable approaches in policy learning that provide a clear framework for treatment decisions within dynamic treatment regimes. 
Decision lists operate as sequential if-then rules, where each rule specifies conditions based on patient characteristics to guide treatment selection in a straightforward, deterministic manner \citep{zhang2018interpretable, tschernutter2022interpretable}. 
It is advantageous for transparency. For increased interpretability, sparse decision lists and pruned trees reduce model complexity, maintaining essential decision criteria without sacrificing clarity.  

\textbf{Extensions to paradigms 2 and 3.}
Extensions of this approach have also been developed to address more complex settings, such as paradigms 2 and 3, which involve multi-stage decision-making scenarios \citep{liao2022batch, chen2023steel}. 
Essentially, we replace the \acrshort{OPE} methods in Paradigm 1 with their counterparts in Paradigms 2 and 3 introduced above. 







\subsubsection{Model-Based Approaches} 
\acrfull{MBRL} is a technique that leverages explicit models of the environment’s dynamics to guide policy learning. This approach, rather than relying solely on observed rewards, uses a parameterized model to predict state transitions and rewards, enabling it to generate synthetic experiences that help train policies without direct interaction with the environment. Techniques in \acrshort{MBRL} can include learning the dynamics of the environment to inform both planning and control, making it possible to learn policies even in complex, high-dimensional spaces \citep{deisenroth2011pilco}. These methods have shown effectiveness in offline reinforcement learning settings, as they mitigate the limitations of direct interaction by enabling supervised learning methods to fit the model and then use it for training or planning in a simulated environment \citep{sutton1991dyna, levine2016end}.

One key advantage of \acrshort{MBRL} is its potential for sample efficiency, as it reuses past experiences by generating additional trajectories, thus enhancing policy learning. Additionally, model-based methods are versatile; they can integrate uncertainty estimation techniques to counteract distributional shifts. 
By estimating epistemic uncertainty, \acrshort{MBRL} can prevent the exploitation of inaccuracies in the learned model. Recent studies, such as those using model-predictive control and policy rollouts, indicate promising results in high-dimensional tasks and show robust performance under various degrees of distributional shift, further affirming MBRL as a viable solution for offline policy learning \citep{nagabandi2018neural, chua2018deep}.


\subsubsection{Address Increased Uncertainty from Distribution Shift}\label{sec:pessimism}

Besides the selection bias caused by the distribution shift in offline data, another prominent issue is the inflated uncertainty. 
The increased uncertainty is resulted from the inherent limitations in the observational data used to inform policy decisions, which often fails to comprehensively represent the entire state and action space. 
Models trained on such non-representative data can yield overoptimistic predictions about the outcomes of actions, especially those that deviate substantially from the behavior policy used during data collection. As a result, suboptimal decisions would be made, as one policy may appear to be better just because its value estimate has a bigger variance. 
This problem is exacerbated in complex environments where the state and action spaces are vast and diverse, increasing the likelihood of encountering unrepresented scenarios. 

To mitigate the risks associated with overoptimistic predictions,  penalty or pessimism strategies are employed in offline policy learning. 
Penalty-based methods \citep{wu2019behavior, jaques2019way}  or constraint-based methods \citep{kumar2019stabilizing, fujimoto2019off, siegel2020keep} explicitly encourage or require the estimated optimal policy to stay close to the data distribution by introducing a penalty or constraint for taking actions that lead to high uncertainty. This penalty discourages the selection of such actions, steering the policy towards actions with more predictable outcomes based on the available data. 

In contrast, 
pessimism-based methods \citep{cief2022pessimistic, jeunen2021pessimistic, rashidinejad2021bridging, jeunen2021pessimistic, zhou2023optimizing, chen2023steel} use an implicit and data-driven way to stay conservative and close to data distribution. 
It is typically based on explicit uncertainty quantification for the value estimates and then selects the policy that optimizes the value lower bounds. 
This approach hence reduces the likelihood of the algorithm recommending policies that just happen to be optimal due to high uncertainty. 
Theoretically, the pessimism-based algorithms can find an optimal policy when the data cover the trajectories of an optimal policy, an assumption that is much weaker than the full-coverage requirement.

In summary, the necessity for value pessimism, policy penalty or policy constraint in offline policy learning arises from the need to counteract the inherent uncertainties associated with training models on limited observational data. 
By adopting these strategies, the reliability and safety of the policies derived from offline learning are enhanced, leading to more robust and effective decision-making in practice.

\section{Online \acrlong{CPL}}\label{sec:Online CPL}

This section explores strategies for addressing online decision-making problems with data structures outlined in Paradigms 4-6, where the treatment policy dynamically adjusts in real time based on data continuously collected from interactions with the environment, with the goal typically as to optimize cumulative rewards. 
A critical distinction between offline and online policy learning is the mode of data collection. While the performance of offline \acrshort{CPL} could be constrained by the quality and representativeness of the already collected dataset, online \acrshort{CPL} allows us to constantly acquire new data, facilitating ongoing enhancement of the learned policy alongside swift environmental adaptation. 
Online \acrshort{CPL} problems face challenges due to environmental and learning model uncertainties, particularly unobserved counterfactuals of untaken actions. This leads to the exploration-exploitation dilemma, where the difficulty is in balancing exploring new actions to gather information and exploiting known actions to maximize rewards. 


\subsection{Online Policy Optimization}
We categorize online policy optimization problems into three distinct groups based on their underlying causal structure assumptions. 
The first category encompasses problems that adhere to the data structure outlined in paradigm 4 and are widely studied within the framework of \textit{bandit} problems \citep{slivkins2019introduction,lattimore2020bandit}, characterized by sequentially updated policies and independent state information. 
To address the complexities introduced by potential long-term dependencies between states, the second category considers the data structure of paradigm 5 with a Markovian assumption on the state transition process, which is extensively researched as \textit{online \acrshort{RL}} \citep{sutton2018reinforcement}. 
The final category is a broader classification that captures all remaining online learning problems characterized by non-Markovian system dynamics in Paradigm 6. 
This includes, but is not limited to, \acrshort{POMDP} \citep{meng2021memory, spaan2012partially, zhu2017improving} and \acrshort{DTR} bandits \citep{hu2020dtr}. 
\acrshort{POMDP}s operate under the premise of an underlying \acrshort{MDP} model, albeit with the challenge that the state itself is not directly observable. 
Conversely, \acrshort{DTR} bandits leverage the entirety of historical information to iteratively learn an optimal treatment regime, typically within a short horizon due to computational complexity.

\subsubsection{Bandits (Paradigm 4)} \label{sec:bandit_optimization}
The bandit problems have been widely used in a variety of fields, including recommender systems \citep{zhou2017large}, clinical trials \citep{durand2018contextual}, and business economics scenarios \citep{shen2015portfolio, wei2023zero}. 
In essence, bandit algorithms are designed to select actions $A_t$ from a pre-defined, potentially time-varying action space $\mathcal{A}$ at each round $t$, and thereafter receive a reward $R_t$ from the environment based on the action taken. 
The inherent challenges lie in the unknown distributions of the counterfactual rewards $\{R_t(a)\}_{a\in \mathcal{A}}$, which necessitate decision-makers to interact with the environment in a sequential manner and learn reward distributions from received feedback. 
The objective of bandit learning is typically to maximize the cumulative rewards (i.e., $\sum_{t=0}^{T}R_t(A_t)$) or, equivalently, to minimize the cumulative regret (i.e., $\sum_{t=0}^{T}\max_{a\in \mathcal{A}}R_t(a) - R_t(A_t)$). 
The core of bandit algorithms lies in the process of selecting the arm based on the information collected so far, which involves addressing the exploration-exploitation tradeoff.
The most classical bandit problem is the \acrshort{MAB} \citep{slivkins2019introduction, bouneffouf2019survey}, whose action space is defined by a finite number of $K$ actions, or ``arms" (i.e., $\mathcal{A} = \{1,2,\dots, K\}$). At each round, the bandit agent selects an arm, receives a reward, and updates its knowledge of counterfactual outcomes to improve future selections.

Popular algorithms handling the exploration-exploitation tradeoff for \acrshort{MAB} problems can be grouped into three categories. 
The first category is \textbf{$\epsilon$-greedy} \citep{sutton1999reinforcement}, which takes the greedy action with the highest estimated mean with a probability of $\epsilon$ and selects a random action otherwise. 
A more adaptive variant is $\epsilon_{t}$-greedy, where the probability of taking a random action is defined as a decreasing function of $t$, achieving a lower regret bound than $\epsilon$-greedy \citep{cesa1998finite, auer2002finite}. 
While such an approach is simple, general, and intuitive, it lacks statistical efficiency in exploration because it does not quantify and utilize the uncertainty. 

To design a more efficient exploration process utilizing historical observations, 
the \textbf{\acrfull{UCB}}-type algorithm was proposed from a frequentist perspective, 
embodying the principle of \textit{optimism in the face of uncertainty} \citep{auer2002finite}. 
Essentially, at each round, \acrshort{UCB} estimates the upper confidence bound for $\mathbb{E}(R(a))$ of each arm 
$a$ by summing its estimated mean reward and a confidence radius reflecting uncertainty. The arm with the highest upper bound is then selected to balance choosing arms with higher estimated average rewards (exploitation) and those with greater uncertainty (exploration). While \acrshort{UCB} provides near-optimal theoretical guarantees \citep{slivkins2019introduction} and offers deterministic recommendations that facilitate tracking, it can be computationally inefficient for complex reward models and requires intricate derivations to obtain the confidence bounds. 

Alternatively, \textbf{\acrfull{TS}} takes a Bayesian approach to the exploration-exploitation trade-off, achieving theoretical performance comparable to \acrshort{UCB}-type algorithms \citep{russo2018tutorial,agrawal2013further}. \acrshort{TS} begins by assuming a prior distribution over each arm's mean reward. At each round, it updates its belief about the reward distribution by computing the posterior distribution of each arm’s mean reward based on observations collected so far. \acrshort{TS} then samples a mean reward from each posterior and selects the arm with the highest sampled reward. This sampling approach naturally balances exploration and exploitation, as arms with greater uncertainty would have wider posterior distributions, increasing their likelihood of selection. 
While the regret bound of \acrshort{TS} is similar to that of the \acrshort{UCB}, it is worth noting that \acrshort{TS} has been empirically observed to usually have a lower regret than \acrshort{UCB} in the long run \citep{chapelle2011empirical}. 
Furthermore, \acrshort{TS} is noted for its computational efficiency (especially with conjugate priors)  and its adaptability to complex models through bootstrapping techniques for approximating posteriors \citep{geman1984stochastic,chen2009bayesian,wan2023multiplier}. 
However, \acrshort{TS}'s efficacy heavily depends on the accuracy of the prior distribution \citep{lattimore2020bandit}. 
An inaccurate prior can lead to either excessive or insufficient exploration, and hence suboptimal regret. To further tackle this challenge, meta \acrshort{TS} \citep{kveton2021meta, wan2021metadata, wan2023towards} has recently been introduced, achieving better performance that is robust to the prior specifications. 

Recent research has examined numerous extensions of the \acrshort{MAB} framework to address complexities in a variety of real-world scenarios, 
such as personalization, large action spaces, and managing multiple decision-making tasks. 
Contextual Bandits \citep{chu2011contextual,agrawal2013thompson,kveton2020randomized,li2010contextual} use environmental contextual information, such as a user's gender, occupation, season, and so on, to tailor recommendations; 
structured bandits \citep{agrawal2017thompson,kveton2015cascading,chen2013combinatorial,wan2023towards} use diverse application-specific structures to specify the reward function and utilize the rich outcome information; 
and multi-task bandits \citep{kveton2021meta, wan2021metadata,hong2022hierarchical,basu2021no} share insights across similar tasks to optimize learning for newly introduced tasks. 

\textbf{Policy-based approaches. }
All of the algorithms previously discussed can be classified as value-based, i.e., focusing on estimating the reward function and then determining the optimal action based on the estimated rewards. 
In contrast, there are also policy-based approaches that directly learn preferences over actions. 
For instance, gradient bandit algorithms \citep{sutton2018reinforcement,mei2023stochastic} for \acrshort{MAB} problems maintain a numerical preference, $H_t(a)$, for each action $a$. 
Here, the probability of selecting action $a$ at time $t$ is determined by a softmax function: $\frac{exp(H_t(a))}{\sum_{b\in\mathcal{A}}exp(H_{t}(b))}$. Shifting the focus to contextual bandits, classification oracle-based algorithms \citep{slivkins2019introduction, langford2007epoch, dudik2011efficient, agarwal2014taming, krishnamurthy2016contextual} consider policies that map contexts to actions and determine the optimal policy over a pre-specified policy class. They propose to approach policy class optimization as a cost-sensitive multi-class classification problem, utilizing historical collections of $(\boldsymbol{s}_t, a_t, r_t)$ triplets. This formulation enables the use of a variety of cutting-edge machine learning methods, significantly improving computational efficiency and reducing running time \citep{slivkins2019introduction}.

\textbf{Bandits with causality to enhance learning efficiency. }
Recent advances in \acrshort{CSL} and \acrshort{CEL} offer a promised opportunity to increase the learning efficiency of Bandits. For example, \acrshort{CEL} methods like \acrshort{IPW} and \acrshort{DR} estimation have been adapted for value estimation to mitigate potential biases in bandit learning, resulting from reward model misspecification or covariate imbalances—particularly when training data are sparse or unrepresentative in specific areas of the context space. Two primary approaches have emerged for integrating \acrshort{IPW}/\acrshort{DR} into bandit algorithms: (i) generating unbiased pseudo-rewards from observed rewards and propensity scores, adhering to the principles of \acrshort{IPW}/\acrshort{DR} estimators in \acrshort{CEL}, which are subsequently used in the bandit update process in place of the observed rewards \citep{bietti2021contextual, kim2019doubly, kim2021doubly, kim2023double}; and (ii) employing importance-weighted regression, wherein each observation is weighted by the inverse of its propensity score \citep{dimakopoulou2019balanced, bietti2021contextual}. While these methods remain focused on maximizing rewards, recent research has also explored integrating classical bandit frameworks with meta-learners to optimize the incremental benefits of an action (e.g., $\tau_{a} = R(A = a) - R(A \neq a)$), aiming to enhance the return on investment \citep{sawant2018contextual, kanase2022application, zhao2022mitigating} or potentially simplify the model estimation \citep{carranza2023flexible}.

Another line of recent research leverages causal techniques to transfer knowledge from logged data to ``warm up" bandit agents. This approach initiates agents with an informative estimate of the environment, thereby reducing the number of rounds required for online exploration. \citet{li2021unifying} propose creating a pseudo-environment using logged data to synthesize action outcomes via matching and weighting and introduce a two-stage learning process under the \acrshort{UCB} framework. Specifically, in the first stage, the pseudo-environment would be used for the simulation of interactions with the bandit agent in order to prepare the agent for real-world engagement. In the second stage, the bandit agent uses the knowledge gained in the first stage to interact with the real world, significantly reducing the possibility of unnecessary exploration. Similarly, \citet{xu2023thompson} provide a complementary view with a \acrshort{TS}-inspired variant. Distinct from creating a pseudo-environment, \citet{zhang2017transfer} employ \acrshort{SCM}s to derive causal bounds for potential outcomes, facilitating the transfer of learnings between bandit agents. By leveraging the causal structure of the environment and the observed trajectories from completed bandit agents, they proposed to derive the 2-sided bounds of the potential rewards over the action space for the target bandit agent. These bounds are subsequently utilized to eliminate less effective options during the initialization phase and to refine the \acrshort{UCB} estimates throughout the learning process.

Furthermore, side causal information is particularly effective in improving learning efficiency in scenarios with multiple intervention variables. For example, \citet{lattimore2016causal} is the first to introduce such a class of causal bandits problems. Given a causal graph among variables that include either interventional/uninterventional variables and reward, agents are able to select more than one variable in the causal graph to intervene at each round. Utilizing the causal graph to transfer information among interventional variables and hence reduce the number of explorations needed, \citet{lattimore2016causal} and \citet{sen2017identifying} focus on the best arm identification problem, while \citet{lu2020regret} and \citet{nair2021budgeted} propose algorithms to minimize the cumulated regret. However, when considering a large number of interventional variables, \citet{lee2018structural} empirically showed that a brute-force way to apply standard bandit algorithms on all interventions can suffer huge regret. To further enhance the sampling efficiency, \citet{lee2018structural, lee2019structural} proposed narrowing the action space by determining the possibly-optimal minimal intervention set before applying standard bandit learning algorithms, while \citet{subramanian2021causal} suggested performing target interventions to allocate more samples to targeted subpopulations that are more informative about the most valuable interventions.

In addition to improving learning efficiency, causality also addresses broader challenges in bandit learning, such as assumption violations, discussions of which can be found in Chapter \ref{sec:assump_violated}.


\subsubsection{Online \acrlong{RL} (Paradigm 5 \& 6)} \label{sec:RL_optimization}
Unlike bandits, which assume actions have only immediate effects, \textbf{online \acrlong{RL}} \citep{sutton2018reinforcement} considers the influence of actions on future outcomes through transitions of the environment's state. This consideration is crucial in real-world applications such as autonomous vehicles \citep{kiran2021deep} and robotics \citep{singh2022reinforcement}, where the long-term effects of actions must be factored into decision-making. To account for these complex state transitions, \acrshort{MDP}s are typically employed as the standard framework for online \acrshort{RL}. Similar to offline \acrshort{RL} (Paradigm 2), online \acrshort{RL} algorithms are generally classified into value-based and policy-based approaches, depending on whether they estimate $V^{\pi}(\boldsymbol{s})/Q^{\pi}(a,\boldsymbol{s})$ to assist with policy optimization.

\textbf{Value-based approaches. }
Among value-based \acrshort{RL} algorithms, Monte Carlo sampling  \citep{singh1996reinforcement} is the most straightforward approach. It samples complete trajectories and then uses the average cumulative reward from relevant sub-trajectories as an estimator for the $Q$ and $V$ functions, which are then used directly for policy optimization. This method can be adapted to non-Markovian environments (Paradigm 6), offering flexibility. However, it requires waiting until the end of a trajectory to collect data, making it less suitable for online \acrshort{RL} problems with an infinite horizon. In contrast, \acrfull{TD}\citep{sutton1988learning} learning iteratively improves the estimate of $Q^{\pi}$ by leveraging the recursive form of the Bellman equation (\ref{eqn:bellman_Q}). For example, SARSA \citep{rummery1994line}, a type of \acrshort{TD} learning, updates the Q-function as
\begin{equation*}
    \hat{Q}^\pi(a, \boldsymbol{s}) \leftarrow \hat{Q}^\pi(a, \boldsymbol{s}) + \alpha \Big[r + \gamma \hat{Q}^\pi(\pi(\boldsymbol{s}'), \boldsymbol{s}')  - \hat{Q}^\pi(a, \boldsymbol{s}) \Big], 
\end{equation*} 
where $(\boldsymbol{s},a,r,\boldsymbol{s}')$ is a newly collected transition tuple, $\alpha$ is the learning rate, and $[r + \gamma \hat{Q}^\pi(\pi(\boldsymbol{s}'), \boldsymbol{s}')  - \hat{Q}^\pi(a, \boldsymbol{s})]$ is the temporal difference. Similarly, Q-learning  \citep{watkins1992q} employs a \acrshort{TD} approach but defines the \acrshort{TD} as $[r + \gamma max_a\hat{Q}^\pi(a, \boldsymbol{s}')  - \hat{Q}^\pi(a, \boldsymbol{s})]$. Another major value-based \acrshort{RL} method is A-learning \citep{baird1993advantage,gu2016continuous,schulman2015high,li2019hierarchical}, which, like its offline counterpart, focuses on learning the relative value of action policies (i.e., $A^{\pi}(a,\boldsymbol{s}) = Q^{\pi}(a,\boldsymbol{s}) - V^{\pi}(\boldsymbol{s})$) instead of the absolute value functions. 

\textbf{Policy-based approaches. }
Policy-based approaches, such as REINFORCE \citep{williams1992simple}, PPO  \citep{schulman2015trust}, and TRPO  \citep{schulman2017proximal}, directly optimize the policy $\pi^*$ by maximizing the objective function $J(\theta) = \mathbb{E}_{\tau\sim p(\cdot;\theta)}r(\tau)$, where $\tau$ denotes a complete trajectory of states and actions, $r(\tau) = \sum_{t} R_t(\boldsymbol{S}_t, A_t)$ is the cumulative reward along the trajectory, $\theta$ parameterizes the policy, and $p(\cdot;\theta)$ is the probability distribution over trajectories induced by policy $\pi_{\theta}$. Policy-based methods can be further categorized into gradient-based  \citep[e.g., ][]{williams1992simple,schulman2015trust,schulman2017proximal,levine2013guided,peters2008reinforcement,baxter2001infinite} and gradient-free  \citep[e.g.,][]{salimans2017evolution,koutnik2013evolving} approaches. Gradient-based approaches rely on the policy gradient theorem  \citep{williams1992simple}, such that
\begin{align}
    \nabla_{\theta}J_{\theta} &= \mathbb{E}_{\tau\sim p(\cdot;\theta)}r(\tau)\nabla_{\theta}log p(\tau;\theta),\nonumber\\
    &=\mathbb{E}_{\tau\sim p(\cdot;\theta)}\left[\sum_{t} R_t(\boldsymbol{S}_t, A_t)\right]\left[\sum_{t}\nabla_{\theta}log\pi_{\theta}(A_t|\boldsymbol{S}_t)\right],\label{policy-gradient theom}
\end{align}
to update $\theta$ via gradient ascent. 
With (\ref{policy-gradient theom}), $\nabla_{\theta}J_{\theta}$ is typically estimated using Monte Carlo rollouts  \citep{williams1992simple} or importance sampling with previously collected trajectories  \citep{levine2013guided}. Notably, as the derivation of $\nabla_{\theta}J(\theta)$ does not depend on the Markov property, most gradient-based approaches can be applied to \acrshort{POMDP}s (paradigm 6) without modification. On the other hand, gradient-free approaches focus on searching for the optimal policy within a predefined policy class. 

One limitation of the policy-based approaches is their sample inefficiency, as they require frequent generation of new trajectories from scratch to evaluate the current policy, which can lead to high variance. To mitigate this, the \textit{Actor-Critic} method combines elements of both value-based and policy-based approaches to improve efficiency \citep{sutton1999policy,mnih2016asynchronous,schulman2015high,gu2016q}. It maintains a policy function estimator (the actor) to select actions and a value function estimator (the critic) to evaluate and guide policy updates through gradient descent.

\textbf{Exploration. }
All of the aforementioned algorithms share strong connections with the offline \acrshort{RL} algorithms discussed earlier. The key distinction lies in the addition of an exploration component, which addresses the issue of distributional shift in offline \acrshort{RL} by strategically collecting new interaction data to continuously improve the policy. Similar to bandits, various exploration strategies are studied to enhance exploration efficiency in online \acrshort{RL}. For example, DQN \citep{mnih2015human} employs $\epsilon$-greedy for exploration. Inspired by \acrshort{TS}, bootstrapped DQN \citep{osband2016deep} maintains multiple Q-value estimators derived from bootstrapped sample datasets to approximate a distribution over Q-functions, allowing for exploration by randomly selecting a policy according to the distribution. \citet{bellemare2016unifying} employs the \acrshort{UCB} framework, adding a count-based exploration bonus to incentivize the agent to explore new or less frequently visited areas of the state space. Additionally, noise-based exploration, which involves adding noise to observation or parameter spaces, is widely used in deep \acrshort{RL}. For a comprehensive review of exploration strategies in deep \acrshort{RL}, see \citet{ladosz2022exploration}.

More recently, causal graph structures have been utilized to enable more efficient exploration in online \acrshort{RL}. 
For instance, \citet{seitzer2021causal} introduces a framework that employs situation-dependent causal influence, measured via conditional mutual information, to identify states where an agent can effectively influence its environment. Integrating this measure into \acrshort{RL} algorithms enhances exploration and off-policy learning, significantly improving data efficiency in robotic manipulation tasks. \citet{hu2022causality} proposes the Causality-Driven Hierarchical \acrshort{RL} framework, which leverages causal relationships among environment variables to discover and construct high-quality hierarchical structures for exploration, thereby avoiding inefficient randomness-driven exploration. More discussions on how causality facilitates online \acrshort{RL} from other perspectives can be found in Chapter \ref{sec:assump_violated}.

\subsection{Online Policy Evaluation (Paradigm 4 \& 5)} \label{sec:bandit_evaluation}

The evaluation of the performance of policies plays a vital role in many areas, including medicine and economics \citep[see e.g., ][]{chakraborty2013statistical,athey201921}. By evaluation, we aim to unbiasedly estimate the value of the optimal policy that the bandit policy is approaching and infer the corresponding estimate. Although there is an increasing trend in policy evaluation \citep[see e.g., ][]{li2011unbiased,dudik2011doubly,jiang2016doubly,swaminathan2017off,wang2017optimal,kallus2018policy,su2019doubly}, we note that all of these works focus on learning the value of a target policy offline using historical log data. Instead of a post-experiment investigation,  it has attracted more attention recently to evaluate the ongoing policy in real-time.

Despite the importance of policy evaluation in online learning, the current bandit literature suffers from three main challenges. 
First, the data, such as the actions and rewards sequentially collected from the online environment, are not independent and identically distributed (i.i.d.) since they depend on the previous history and the running policy. In contrast, the existing methods for the offline policy evaluation \citep[see e.g., ][]{li2011unbiased,dudik2011doubly} primarily assumed that the data are generated by the same behavior policy and i.i.d. across different individuals and time points. 
The second challenge lies in {estimating the mean outcome under the optimal policy online}. Although numerous methods have recently been proposed to evaluate the online sample mean for a fixed action \citep[see e.g., ][]{nie2018adaptively,neel2018mitigating,deshpande2018accurate,shin2019sample,shin2019bias,waisman2019online,hadad2019confidence,zhang2020inference}, none of these methods is directly applicable to online policy evaluation, as the sample mean only provides the impact of one particular arm, not the value of the optimal policy in bandits that considers the dynamics of the online environment. 
Third, given data generated by an online algorithm that maintains the exploration-and-exploitation trade-off sequentially, inferring the value of a policy online should consider such a trade-off and quantify the probability of exploration and exploitation. 

There are very few studies directly related to the topic of online policy evaluation. \cite{chambaz2017targeted} established the asymptotic normality for the conditional mean outcome under an optimal policy for sequential decision making.  Later, \cite{chen2020statistical} proposed an inverse probability weighted value estimator to infer the value of optimal policy using the $\epsilon$-Greedy method. 
Recently, to evaluate the value of a known policy based on the adaptive data, 
\citet{bibaut2021post} and \citet{zhan2021off} proposed to utilize the stabilized doubly robust estimator and the adaptive weighting doubly robust estimator, respectively. Both methods focused on obtaining a valid inference of the value estimator under a fixed policy by conveniently assuming a desired exploration rate to ensure sufficient sampling of different arms. 
Also see other recent advances that focus on statistical inference for adaptively collected data \citep{dimakopoulou2021online,zhang2021statistical,khamaru2021near,ramprasad2022online} in bandit or \acrshort{RL} setting. 
To infer the value of the optimal policy by investigating the exploration rate in online learning, \citet{ye2023doubly} explicitly characterized the trade-off between exploration and exploitation in online policy optimization by deriving the probability of exploration in bandit algorithms. Their work 
proposed the doubly robust interval estimation (DREAM) method to infer the mean outcome of the optimal online policy with double protection.




\section{Causal Assumptions Violated Scenarios}\label{sec:assump_violated}
In the sections above, much of the literature proceeds under the causal identification assumption outlined in Section \ref{sec:prelim_assump}. However, real-world scenarios sometimes violate these assumptions. 
For instance, the \acrshort{SUTVA} assumption assumes that a unit's outcome isn't influenced by the treatment of other units. 
Yet, for example, in the spread of COVID-19, if we simply assume each individual as one unit, the \acrshort{SUTVA} is violated as each patient's health can be affected by the immunization status of the entire community; 
the \acrshort{NUC} assumption is commonly violated in observational data due to those unmeasured confounders that researchers can't fully control without experimental design; 
the positivity assumption is violated when some individuals can not receive a specific treatment or control with certainty due to ethical issues or budget constraints. 

It is crucial to handle these cases for better decision-making, as accurately estimating causal effects in the face of assumption violations is usually the first step towards optimizing policies for higher rewards. Recently, literature has started to focus on addressing each assumption violation using various tools, most of which are originally developed in the area of causal inference. In this section, we'll provide a concise overview and outline open questions in the literature that need further exploration.

\subsection{Unmeasured Confounders}
\begin{table}[tbh]
\centering\renewcommand\cellalign{lc}
\begin{tabular}{l|ll}
\hline
Paradigm & \acrlong{IV} & Data Integration \\ 
\hline
Paradigm 1    &      \makecell{\cite{angrist1996identification}\\
\cite{wang2018bounded}\\ \cite{qiu2021optimal}\\
\cite{cui2021semiparametric}} & \cite{wu2022integrative}     \\
\hline
Paradigm 2/5    &     \makecell{\cite{li2021causal, fu2022offline}\\
\cite{xu2023instrumental}} & \makecell{\cite{gasse2021causal}\\
\cite{imbens2022long} }
\\
\hline
Paradigm 3         &        \makecell{\cite{chen2023estimating}\\
\cite{xu2023instrumental}} &  \cite{athey2020combining}        \\
\hline
Paradigm 4   &       \cite{kallus2018instrument} &\makecell{\cite{bareinboim2015bandits}\\
\cite{sen2017contextual}\\
\cite{xu2021deep}}
\\ 
\hline
\end{tabular}
\caption{A brief summary of papers when unmeasured confounders exist}\label{table:1}
\end{table}

Table \ref{table:1} provides a brief summary of some representative approaches in handling the violation of the \acrshort{NUC} assumption. 
When unmeasured confounder $U$ exists, traditional methods for \acrshort{CEL} and \acrshort{CPL} 
would results in biased estimates. To adjust for potential bias, the current literature can be broadly divided into two categories: (1) using proxies such as \acrfull{IV}, or (2) incorporating additional data sources, typically by integrating experimental data without unmeasured confounders with existing observational data (commonly referred to as data integration).

The use of \acrshort{IV} can be traced back to the 1920s \citep{wright1928tariff}, gaining widespread recognition with the introduction of the \acrfull{2SLS} method \citep{angrist1995two}. Typically, a variable $X$ is called an \acrshort{IV} when it satisfies the following three conditions: 
\begin{enumerate}
    \item[a.] \acrshort{IV} independence: $Z\perp U|S$.
    \item[b.] \acrshort{IV} relevance: $Z\not\perp A|S$.
    \item[c.] Exclusion restriction: $R(z,a)=R(a)$ for any $(z,a)\in \mathcal{Z}\otimes \mathcal{A}$.
\end{enumerate}

Notably, \acrshort{IV} has been employed without stringent modeling assumptions to effectively estimate \acrshort{ATE}s \citep{angrist1996identification}, where certain monotonicity assumption is usually imposed to guarantee the identifiability. In single-stage setting (Paradigm 1), recent advancements, such as \citet{wang2018bounded}, have focused on developing semi-parametric efficiency estimators for \acrshort{HTE} with \acrshort{IV}. Additionally, \citet{qiu2021optimal} and \citet{cui2021semiparametric} have contributed to the literature by deriving optimal policies that maximize rewards in the presence of \acrshort{IV}. 


Under \acrshort{MDP}s (Paradigms 2 or 5), the existence of unobserved state variables also greatly influences both estimation and decision making process. This problem has been explored in the literature of \acrshort{RL}, as evidenced by works such as \citet{li2021causal, fu2022offline, xu2023instrumental,gasse2021causal,imbens2022long}. In offline \acrshort{RL}, \citet{fu2022offline} and \citet{xu2023instrumental} introduced \acrshort{IV}-based methods to ensure consistent \acrshort{OPE} in scenarios involving confounded sequential decision-making. \citet{fu2022offline} focus on pessimistic \acrshort{RL} and offer finite-sample suboptimality guarantees, while \citet{xu2023instrumental} emphasize semi-parametric efficiency, statistical inference, and extensions to high-order \acrshort{MDP}s and \acrshort{POMDP}s.
In online settings, \citet{li2021causal} proposed an \acrshort{IV}-based Q-learning algorithm to learn optimal policies in an interactive decision making environment.
Aside from \acrshort{IV}s, another line of research aims to combine interventional data and observational data for better decision making. For instance, \citet{imbens2022long} combined short-term experimental data and long-term observational data with potential confounders to handle the identification and estimation of long-term treatment effects with asymptotic theory guarantees. \citet{gasse2021causal}, on the other hand, tailored on better policy learning by utilizing confounded information in offline data. 

In scenarios where the Markov assumption doesn't hold (Paradigm 3), addressing confounders requires tailored approaches depending on differences in data-generating structures. For instance, \citet{shi2022minimax} proposed an \acrshort{OPE} approach within the framework of \acrshort{POMDP}s, which is a natural extension of \acrshort{MDP}s to handle unmeasured confounders present in the latent state. In \acrshort{DTR}, \citet{chen2023estimating} also employed \acrshort{IV} for consistent treatment effect estimation and policy optimization. When only a short-term variable (i.e., a surrogate) is observed for long-term treatment effect estimation, and there are no clear state transitions defined as \acrshort{DTR}  or \acrshort{MDP}s, the data structure introduced by \citet{athey2020combining} is more suitable for identifying and estimating long-term effects using short-term experimental data.

In online bandits learning (Paradigm 4), some work has also been done with \acrshort{IV} to help detect possible unmeasured confounders and avoid biased policy learning \citep{sen2017contextual,kallus2018instrument,bareinboim2015bandits,xu2021deep}. 
In the context of causal inference, the bandits problem is also equivalent to first estimating $R(a)$ and adding proper exploration to obtain a suboptimal regret bound. When confounders exist, at least a portion of the relationship between $A$ and $R$ is not captured in the reward modeling process, leading to biased reward modeling and, consequently, biased policy learning outcomes. Given the flexible approaches in handling unmeasured confounders in \acrshort{CEL}, recent research has been developed to address this problem. Most of the existing literature focuses on introducing proxy variables, such as \acrshort{IV}s \citep{kallus2018instrument}, where authors investigate optimal policies to maximize intent-to-treat regret in the presence of potential non-compliance and unmeasured confounders, or combining observational data for confounding adjustment within the structural causal equation model \citep{bareinboim2015bandits}. Additionally, \citet{xu2021deep} proposed a two-stage regression scheme based on proxy variables to handle unmeasured confounders, especially when the data is high-dimensional and non-linear.


\subsection{Interference}
\begin{table}[tbh]
\centering
\renewcommand
\cellalign{lc}
\begin{tabular}{l|l}
\hline
Paradigm &  Methogologies \\ 
\hline
Paradigm 1    &    

\makecell{Experimental design: \makecell{\cite{viviano2020experimental, aronow2017estimating}\\
\cite{leung2022rate, viviano2024policy, leung2022causal}}\\ Observational studies: \makecell{\cite{forastiere2021identification, qu2021efficient}\\\cite{bargagli2020heterogeneous, su2019modelling}}}      \\
\hline
Paradigm 2/5       &         
Multi-agent \acrshort{RL}: \makecell{\cite{yang2018mean,shi2022multi,luo2024policy}\\
\cite{chen2021pessimism,yang2024spatially,jia2024multi}\\
\cite{yang2021believe, pan2022plan, jiang2021offline}
}
\\
\hline
Paradigm 3        &      \acrshort{DTR}  with interference: \cite{jiang2023dynamic}        \\
\hline
Paradigm 4      &    
Multi-agent bandits: \makecell{\cite{verstraeten2020multi, bargiacchi2018learning}\\
\cite{dubey2020kernel,jia2024multi}}
\\ 
\hline
\end{tabular}
\caption{A brief summary of papers when interference exists}\label{table:2}
\end{table}

Table \ref{table:2}  provides a brief summary of some representative approaches in handling the violation of the \acrshort{SUTVA} assumption. 
Interference, often known as the existence of \acrfull{SE}, is a commonly encountered problem in causal inference. 
Generally, it requires extending the \acrshort{SUTVA} assumption in that the potential outcomes of unit $i$ depend on not only $A_i$, but also the actions of other units. 
For example, the reward of unit $i$ under interference is denoted by $R_i(\boldsymbol{A})$, where $\boldsymbol{A} = \{A_1,\dots, A_n\}$. 
Following this definition, the treatment effect can be decomposed into two parts: \acrshort{DE} and \acrshort{SE} : 
\begin{equation*}
    \begin{aligned}
    &\text{DE}_i(a_i,a_i',\boldsymbol{a}_{-i})=R_i(a_i,\boldsymbol{a}_{-i})-R_i(a_i',\boldsymbol{a}_{-i}),\\
    &\text{SE}_i(a_i,\boldsymbol{a}_{-i},\boldsymbol{a}_{-i}')=R_i(a_i,\boldsymbol{a}_{-i})-R_i(a_i,\boldsymbol{a}_{-i}'),
    \end{aligned}
\end{equation*}
where $\boldsymbol{a}_{-i}$ denotes the action assignment vector for all units except unit $i$.

Due to this dependency, directly modeling the treatment effect without considering the actions of other units can lead to bias. 
However, in extreme situations where each unit's reward depends on every single unit's action, any reward modeling approach to generalize findings across units would fail, making causal effect identification impossible. 
Consequently, there is a growing trend in the literature toward identifying and estimating the \acrfull{DE} and \acrfull{SE} under various model structures, aiming to strike a balance between avoiding overly stringent assumptions on
interference structure and allowing learning from existing data. 

In existing literature, various interference structures have been considered, including but not limited to partial interference \citep{sobel2006randomized, qu2021efficient}, stratified interference \citep{hudgens2008toward}, neighborhood interference \citep{forastiere2021identification}, spatial interference \citep{leung2022rate}, and cluster network interference \citep{bargagli2020heterogeneous}. Despite various definitions, most types of interference share a similar two-step structure for simplifying the problem. First, units are typically categorized into groups through clustering or partitioning, assuming that interference occurs only within each group. Extensions of this assumption allow for interference between any units, where degree of interference decreases with distance but does not necessarily zero \citep{leung2022rate, leung2022causal}. 
Second, to further simplify causal identification and estimation within each cluster, where the strength of interference level may vary by domain assumption, interference is also commonly quantified by \textit{exposure mapping}. This concept, proposed by \citet{aronow2017estimating}, is similar to the ``effective treatments'' function in \citet{manski2013identification}. Generally, exposure mapping assumes that interference among units is passed through a lower-dimensional exposure mapping function, often known and assumed to follow a parametric form to easily quantify the \acrshort{SE}  between interfering units. This approach has been widely adopted in various papers under specific mapping forms, such as \citet{su2019modelling,qu2021efficient, viviano2024policy}.

Depending on the quantity of interest and the flexibility of managing treatment assignment, different methods have been developed to address various application needs. There is a growing trend of research focusing on estimating causal \acrshort{DE}s and \acrshort{SE}s in both (1) experimental design \citep{viviano2020experimental, aronow2017estimating, leung2022rate, viviano2024policy, leung2022causal} and (2) observational studies \citep{forastiere2021identification, qu2021efficient, bargagli2020heterogeneous, su2019modelling}. Specifically, among the approaches focusing on (1) experimental design, \citet{viviano2020experimental} and \citet{leung2022rate} concentrate on developing optimal designs to maximize specific goals, such as achieving near-optimal rates of convergence for global average causal effect estimation \citep{leung2022rate} or minimizing the variance of the estimators. In contrast, \citet{aronow2017estimating}, \citet{leung2022causal}, and \citet{viviano2024policy} focus more on estimation under randomized experiments, providing asymptotic guarantees for the proposed estimators for both average effects \citep{aronow2017estimating, leung2022causal} and heterogeneous effects \citep{viviano2024policy}. Among the approaches focusing on (2) observational studies, methods vary based on assumptions about the type of interference and reward modeling. For example, \citet{su2019modelling} considers the reward as a linear function of neighbors’ covariates and treatments, extending Q-learning and A-learning to scenarios with interference under certain structural assumptions. \citet{forastiere2021identification, qu2021efficient, bargagli2020heterogeneous} emphasize semi-parametric or non-parametric estimation and inference under looser modeling assumptions. Users can select the appropriate method based on their willingness to assume different interference structures and modeling approaches.

Moving to multi-stage settings, including \acrshort{MDP}s \citep{yang2018mean}, \acrshort{DTR} \citep{jiang2023dynamic} and beyond, interference often arises in a muti-agent system. In \acrfull{MARL}, the concept of neighborhood interference and exposure mapping is incorporated into the Q function. Here, the Q value of agent $j$ depends not only on $(s_j, a_j)$ but also on the actions of a neighborhood set $\boldsymbol{a}_{\mathcal{N}_j}$, which is the so-called mean-field approximation strategy \citep{yang2018mean}. Later on, this mean-field approximation method has been applied in various \acrshort{MARL} studies, including \citet{shi2022multi,luo2024policy,chen2021pessimism,yang2024spatially, jia2024multi}. Among these, \citet{shi2022multi} and \citet{luo2024policy} focus more on \acrshort{OPE} with observational data in the presence of both temporal and/or spatial dependences among agents, while
\citet{yang2024spatially} studies the efficiency of spatial randomization to account for interference in the experimental design/exploration side. 
Other related work in offline \acrshort{MARL} includes but not limited to \citet{yang2021believe, pan2022plan, jiang2021offline}. Specifically, \citet{pan2022plan} tackles challenges posed by an increasing number of agents on conservative offline \acrshort{RL}
algorithms, \citet{jiang2021offline} exploits value deviation and transition normalization in non-stationary transition dynamics, and \citet{yang2021believe} focuses on mitigating extrapolation error in offline evaluation.

The multi-agent system is also present in online bandits (paradigm 4) \citep{bargiacchi2018learning, verstraeten2020multi, dubey2020kernel}. In round $t$, each agent $i$ is able to interact with the environment and pull an arm. Interference exists since the actions of agents are mutually affected by each other due to the neighboring relationships among agents. Similar to single-stage setting, a common approach to handle this problem is to decompose the agents into fixed but potentially overlapping subgroups. This decomposition simplifies the reward of the joint action space to the summation of local reward functions, reducing the complexity of global exploration. Following this approach, \citet{bargiacchi2018learning} extended the \acrshort{UCB} algorithm from classical \acrshort{MAB} to the multi-agent scenario. Shortly after, \citet{verstraeten2020multi} extended the \acrshort{TS} algorithm to the multi-agent case under similar interference assumptions. \citet{dubey2020kernel} considered another reward modeling structure where interference is transmitted through network contexts, and proposed a kernelized \acrshort{UCB} algorithm to cooperatively maximize cumulative rewards. Rather than handling interference implicitly within multi-agent systems, recent pioneering works have explicitly addressed the interference issue in bandit settings. For example, \citet{jia2024multi} and \citet{agarwal2024multi} focus on \acrshort{MAB}, while \citet{xu2024linear} examines interference in contextual bandits.
Despite these advancements, this field remains relatively underexplored, highlighting a pressing need for more flexible and general approaches to address broader bandit settings using causal methodologies.

\subsection{Positivity}

The positivity (or overlap) assumption is a fundamental requirement ensuring that the data area covered by the target policy has been observed. This assumption posits that the conditional probability of any unit group taking each action is strictly greater than zero. However, in observational studies where actions are not controlled a priori, satisfying this assumption is often challenging due to several factors: (1) in continuous action spaces, it is naturally impossible for an observational study to exhaustively cover all actions; (2) when the feature space for state information $\boldsymbol{S}$ is high-dimensional, ensuring sufficient overlap becomes difficult; and (3) specific treatment options might lack observational data due to ethical concerns, high costs, or oversights during data collection. These issues are prevalent in offline studies (paradigms 1-3), whereas in online experiments, actions are typically selected by designers, thereby mitigating the non-overlapping problem.

Although the violation of the positivity assumption receives less attention compared to the previous two assumptions, existing literature still offers some solutions to address this issue. Given that the general strategies are quite similar across different data structures and that this problem is generally less studied in paradigms 2 and 3, we will focus on introducing approaches to handle the violation of the positivity assumption under paradigm 1.

In the three scenarios mentioned, continuous action space and high-dimensional covariate space are cases where existing papers may not specifically address non-overlapping issues. However, these approaches provide a general framework that naturally tackles this problem. For example, \citet{d2021overlap} discuss the trade-off between incorporating more covariates to mitigate unmeasured confounders and the difficulty of satisfying the positivity assumption. They argue that strict overlap is more restrictive than expected in many studies and derive explicit bounds on the average imbalance in covariate means.

In addressing continuous action space, most existing work tackles this problem through non-parametric smoothing \citep{kallus2018policy, chen2016personalized} or by combining it with (semi)-parametric modeling \citep{chernozhukov2019semi}. For instance, in the \acrshort{IPW} estimator, \citet{kallus2018policy} handle both policy evaluation and optimization by smoothly relaxing the indicator function using a kernel function, while \citet{chen2016personalized} replace the indicator function with a hinge loss. Continuous action space, as a well-established research area in \acrshort{CDM}, has extensive literature beyond the examples listed, and a more detailed review is beyond the scope of this paper.

The last stream of work addresses non-overlapping issues directly, without assuming specific dimensions for the action or covariate space. Due to the scarcity of data points in low-overlap areas, inferring causal relationships in these regions is challenging without additional smoothing or parametric modeling assumptions. Traditional approaches often use trimming \citep{rosenbaum1983central, crump2009dealing, petersen2012diagnosing}, which involves discarding units with estimated propensity scores outside a specific range $[0.1, 0.9]$, or matching \citep{visconti2018handling}. Recently, in \acrshort{OPE}, \citet{khan2023off} applied partial identification techniques from causal inference to derive \acrshort{OPE} results under non-parametric Lipschitz smoothness assumptions on the reward function. In \acrshort{OPO}, the pessimism technique becomes particularly important due to the increased uncertainty for candidate policies with poor coverage. For instance, \citet{jin2022policy} used pessimism with generalized empirical Bernstein’s inequality to study \acrshort{OPO} without the uniform overlap condition, and \citet{chen2023steel} proposed a general and adaptive framework that decomposes \acrshort{OPE} error into positive and singular parts, using corresponding uncertainty bounds to derive a pessimism-based \acrshort{OPO} algorithm.

\section{Real Data}\label{sec:real_data}

In this section, we present two representative examples, the \acrfull{MIMIC-III}\footnote{https://www.kaggle.com/datasets/asjad99/mimiciii} and MovieLens\footnote{https://grouplens.org/datasets/movielens/1m/} dataset, to illustrate the tasks and paradigms discussed in the previous sections. These examples provide a clear and concrete demonstration of how the entire \acrshort{CDM} process can be applied in contexts such as clinical trials, recommendation systems, and beyond. 

\subsection{\acrshort{MIMIC-III}}
The \acrshort{MIMIC-III} dataset \citep{johnson2016mimic} is a large, de-identified, and publicly available collection of medical records that contains comprehensive clinical data from patients admitted to the \acrfull{ICU} at a large tertiary care hospital. Of particular interest are the records of patients with sepsis, a life-threatening response to infection caused by harmful microorganisms, which significantly contributes to clinical research and practice. This disease accounts for over 200,000 annual deaths in the U.S. alone, showing the urgent need for effective and timely interventions. Given its treatable nature, sepsis demands prompt emergency care and robust decision-making frameworks to improve patient outcomes and reduce mortality.

However, the \acrfull{EHR} collected from patients with sepsis in ICUs present significant challenges to the application of existing decision-making methods. Specifically, records from hundreds of thousands of sepsis patients treated at the Beth Israel Deaconess Medical Center between 2001 and 2012 include numerous covariates such as demographics, vital signs, medical interventions, lab test results, and post-treatment outcomes. For researchers, it is essential to disentangle the complex and intricate causal relationships among these variables, understand the impacts of specific sepsis treatments, and select a sufficient and necessary set of variables to analyze the disease. This comprehensive causal understanding is critical to optimizing treatments and ultimately reducing the mortality associated with sepsis. 

To overcome these challenges, we utilize the three main tasks aforementioned and demonstrate the causal decision-making process using \acrshort{MIMIC-III} as an example. The first step \acrshort{CSL}, which aims to uncover the causal relationships between variables, allows us to pinpoint the right and informative set of state variables, the treatments, and mediators that may influence the mortality due to sepsis. In this context, we identify the IV-Input as an actionable treatment with the \acrfull{SOFA} score serves as an important mediator, which is known to describe organ dysfunction or failure. 
The second step \acrshort{CEL}, building upon the causal relationships identified in \acrshort{CSL}, quantifies the strength of causal links and thus measures the most effective treatments and informative mediators. For example, it estimates the average treatment effect to determine how intravenous fluid input (IV-Input) impacts mortality rates across the general patient population and the heterogeneous treatment effect for individual patients. 
The final step \acrshort{CPL} seeks to determine the optimal administration policy to minimize patient mortality rates. Each patient visit is treated as a stage, and depending on the \acrshort{MIMIC-III} data structure, various \acrshort{CPL} algorithms can be applied to the appropriate paradigm to learn the optimal policy from observational data.

Due to privacy concerns, we utilized a subset of the original data that is publicly available on Kaggle. For illustration purposes, we selected several representative features for the following analysis, as listed in Table \ref{table:MIMIC3}.
\begin{table}[ht]
\centering
\begin{tabular}{l|p{12cm}}
\hline
\textbf{Variable} & \textbf{Description} \\
\hline
Glucose & Glucose values of patients \\
PaO2\_FiO2 & The partial pressure of oxygen (PaO2) divided by the fraction of oxygen delivered (FIO2) \\
\acrshort{SOFA} & Sepsis-related Organ Failure Assessment score to describe organ dysfunction/failure \\
IV-Input & The volume of fluids that have been administered to the patient \\
Died\_within\_48H & The mortality status of the patient 48 hours after being administered \\
\hline
\end{tabular}
\caption{Description of variables used in \acrshort{MIMIC-III} data analysis}\label{table:MIMIC3}
\end{table}

In the next sections, we will start from \acrshort{CSL} to learn a significant causal diagram from the data, and then quantify the effect of treatment (IV-input) on the outcome (mortality status, denoted by Died\_within\_48H variable in the data) through \acrshort{CEL}. Finally, we use \acrshort{CPL} to find the best-individualized treatment rules under different settings.

\subsubsection{\acrshort{CSL} on \acrshort{MIMIC-III}}

For our analysis of the \acrshort{MIMIC-III} dataset, we employ the score-based method in \acrshort{CSL} to estimate underlying causal relationships among several key clinical features. The \acrshort{MIMIC-III} dataset comprises a comprehensive range of clinical data from a large cohort of \acrshort{ICU} patients. For this analysis, we selected a subset of variables including Glucose levels, PaO2/FiO2 ratio, \acrshort{SOFA} scores, IV-Input, and mortality within 48 hours. The details of selected features are pivotal in understanding patient outcomes in intensive care, and their descriptions are provided in Table \ref{table:MIMIC3}. Specifically, we utilize the NOTEARS algorithm \citep{zheng2018dags}, which is designed to learn a \acrshort{DAG} without cycle constraints from continuous data, given the complexity of the observed data. 

\begin{figure}
    \centering
    \includegraphics[width=0.5\linewidth]{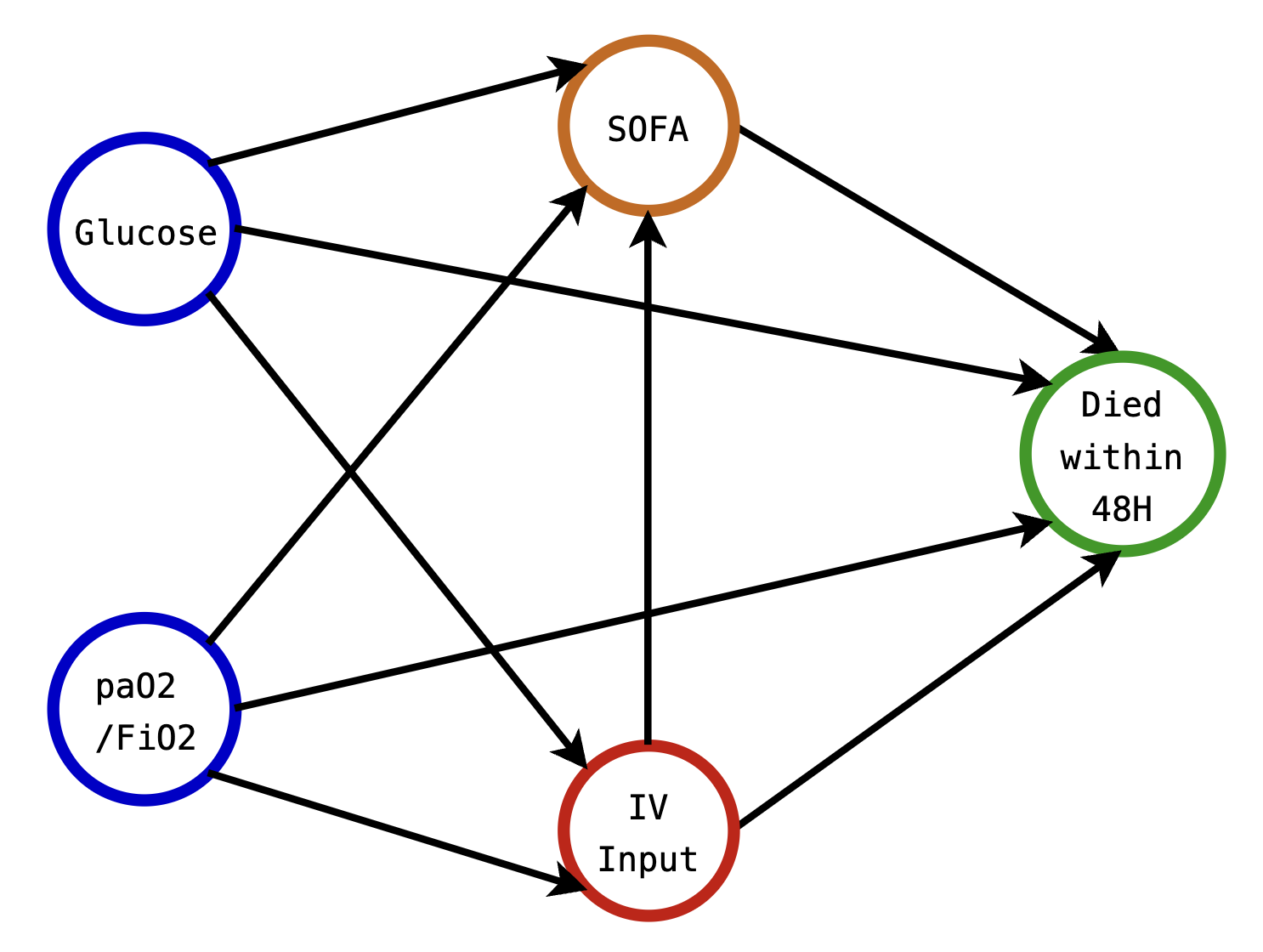}
    \caption{Estimated directed acyclic graph for the selected \acrshort{MIMIC-III} data analysis.}
    \label{fig:DAG_mimic}
\end{figure}

The resulting \acrshort{DAG} from the NOTEARS algorithm is presented in Figure \ref{fig:DAG_mimic}, which reveals a plausible causal structure among the variables. In particular, PaO2/FiO2, a measure of lung efficiency, is identified as an exogenous variable that influences other downstream variables but is not influenced by any of the variables selected in this analysis. Glucose levels and IV-Input appear causally prior to other variables, suggesting their role in early medical intervention. \acrshort{SOFA} score, a critical measure of organ failure, is influenced by both glucose levels and IV-Input, and it further influences the mortality outcome. 
The mortality within 48 hours variable is positioned as an end-point in the causal chain, influenced by the \acrshort{SOFA} score.

The learned causal graph highlights several clinically relevant pathways. Notably, the direct influence of IV-Input on \acrshort{SOFA} score and mortality shows the impact of fluid management in critical care settings. Additionally, the model suggests a pathway from metabolic control (Glucose) through organ function (\acrshort{SOFA} score) to mortality outcomes. These findings could inform targeted interventions aiming to optimize patient care and improve survival outcomes in the \acrshort{ICU}.

\subsubsection{\acrshort{CEL} on \acrshort{MIMIC-III}}
After establishing the causal diagram between the relevant variables, the next step is to quantify the causal effect of IV-Input on reducing the mortality rate of patients. For simplicity, we categorize the treatment, specifically IV-Input, into a binary action space: $A_i=1$ represents the ``High-IV-Input'' group with IV-Input volume greater than $1$, and $A_i=0$ corresponds to the ``Low-IV-Input'' group with IV-Input less than or equal to $1$. Our goal is to apply \acrshort{CEL} methods to estimate the ATE, quantifying the impact as $\mathbb{E}\{R(1) - R(0)\}$.

In the \acrshort{CSL} analysis, we examined the causal relationships between the variables listed in Table \ref{table:MIMIC3} and identified a mediator, the \acrshort{SOFA} score, which is influenced by the treatment (IV-Input) and subsequently impacts the mortality status of patients within the next 48 hours. Utilizing the direct estimator proposed by \citet{robins1992identifiability}, the IPW estimator from \citet{hong2010ratio}, and the robust estimator introduced by \citet{tchetgen2012semiparametric}, we evaluated the natural direct and indirect effects of the treatment regime based on observational data. The results are summarized in Table \ref{table:mediation_MIMIC3} below.

\begin{table}[ht]
\centering
\begin{tabular}{l|ccc}
\hline
\textbf{Methods}        & \textbf{DE}  & \textbf{IE}  & \textbf{TE}    \\ \hline
{Direct Estimator} & -0.2133 & 0.0030  & -0.2104 \\ \hline
{Inverse Probability Weighting}     & -0.2332 & 0       & -0.2332 \\ \hline
{Doubly Robust}  & -0.2276 & -0.0164 & -0.2440 \\ \hline
\end{tabular}
\caption{Comparison of DE, IE, and TE across different estimation methods.}
\label{table:mediation_MIMIC3}
\end{table}
Specifically, compared to the lower IV-Input group, constantly administering a high volume of IV-Input shows a negative impact on survival rates, with an estimated $20\%-25\%$ increase in mortality. This effect is largely driven by the direct influence of the treatment on the final outcome. While this result may seem counterintuitive at first, it could be partly attributed to the general overuse of IV-Input in medical care. 

Therefore, developing a personalized treatment regime to administer the optimal volume of IV-Input tailored to individual patient characteristics is crucial to avoid overuse and meet specific treatment needs. This challenge motivates our exploration of offline policy learning in the next section.


\subsubsection{Offline \acrshort{CPL} on \acrshort{MIMIC-III}}
In this section, we demonstrate the results when applying classic offline \acrshort{CPL} methods to this dataset. 
The simplest modeling usually starts from paradigm 1, where we aggregated the observations for each patient and used the averaged observations as the dataset for the analysis.

As an example, we use the Q-learning algorithm to evaluate two simple treatment rules based on the observed data. 
We specify the following linear model as our Q-function: 
$$
\begin{aligned}
Q(s, a, \beta) &= \beta_{00} + \beta_{01} \cdot \text{Glucose} + \beta_{02} \cdot \text{PaO2\_FiO2} \\ \qquad\qquad\qquad
&+ I(a_1 = 1) \cdot \left(\beta_{10} + \beta_{11} \cdot \text{Glucose} + \beta_{12} \cdot \text{PaO2\_FiO2}\right)
\end{aligned}
$$
We evaluate two target policies. The first is a fixed treatment regime that does not apply treatment, for which Q-learning has an estimated value of $.99$. Another is a fixed treatment regime that applies treatment all the time, with an estimated value of $.76$. Therefore, the result implies that a high dose of IV-Input naively always is worse and increases the mortality rate, aligned with the \acrshort{CEL} results. 

We take one step further to find an optimal policy maximizing the expected value. We use the Q-learning algorithm again to do policy optimization. 
Using the regression model we specified above, the estimated optimal policy is to recommend $A = 0 \ (\text{IV-Input} = 0) \text{ if } -0.0003 \cdot \text{Glucose} + 0.0012 \cdot \text{PaO2\_FiO2} < 0.5633$ and $A = 1 \ (\text{IV-Input} = 1)$ otherwise. 
When applying the estimated optimal treatment regime to individuals in the observed data, IV-Input would be administered to 6 out of the 57 patients.

Based on the domain knowledge, it is usually more plausible to believe the outcome of a patient depends only on his treatment and condition in the past stage. Therefore, we can apply a 3-stage Q-learning in Paradigm 3 to learn the policy. The Q-function is specified as a linear one that considers all the previous stages' states and actions. The learned optimal policy is as follows. 
\begin{itemize}
    \item \textbf{Stage 1:} recommend \( A = 0 \) (IV-Input = 0) if 
        $ 0.0001 \cdot \text{Glucose}_1 + 0.0012 \cdot \text{PaO2\_FiO2}_1 > 0.0551$ and \( A = 1 (\text{IV-Input} = 1)\)  otherwise.     
    \item \textbf{Stage 2:} recommend \( A = 0 \) (IV-Input = 0) if 
        $ 0.0002 \cdot \text{Glucose}_2 - 0.00001 \cdot \text{PaO2\_FiO2}_2 + 0.0070 \cdot \text{\acrshort{SOFA}}_1 < 0.0721 $ and \( A = 1 \) (IV-Input = 1)  otherwise.        
    \item \textbf{Stage 3:} recommend \( A = 0 \) (IV-Input = 0) if 
        $ -0.0005 \cdot \text{Glucose}_2 + 0.0008 \cdot \text{PaO2\_FiO2}_2 - 0.0114 \cdot \text{\acrshort{SOFA}}_2 < 0.2068$ and recommend \( A = 1 \) (IV-Input = 1)  otherwise.    
\end{itemize}

Applying the estimated optimal regime to individuals in the observed data, we can have personalized treatment plans. 
For example, 23 patients will receive IV-Input in the first two stages and no inputs in the last one; while 10 other patients will receive IV-Input only in the first phase.



\subsection{MovieLens}

Recommender systems play a crucial role in personalizing user experiences across various industries. A common example is movie recommendation, where understanding user preferences across different movie genres is essential. However, this task is challenging due to the inherent difficulty of estimating counterfactuals, i.e. how users would have responded if presented with different options. To illustrate how recommender systems can be optimized through causal decision-making, we use movie recommendations as an example, starting with the well-known Movielens dataset.

The MovieLens 1M dataset, derived from an online movie recommendation experiment, is a widely used benchmark for generating data in online bandit simulation studies. User information in this dataset is categorized by age, gender, and occupation. For simplicity, we focus on the top five movie genres in the dataset (Comedy $a=0$, Drama $a=1$, Action $a=2$, Thriller $a=3$, Sci-Fi $a=4$), and analyze users from the top five occupations. The realized reward $R$ is a numerical variable ranging from $\{1,2,3,4,5\}$, with $1$ indicating the lowest level of satisfaction and $5$ representing the highest. As the causal structure of this problem has been well defined, with movie genre as the action and users' rating as the reward, our objectives are two-fold, mainly focusing on CEL and CPL.

First, we begin the process with \acrshort{CEL}, where scientists analyze the logged data to identify general patterns of user preferences. Specifically, the \acrshort{ATE} of one movie genre relative to another is calculated to reveal overarching trends across the user population, and \acrshort{HTE}s are estimated to capture variations in preferences across different user segments, providing a more granular understanding of how different groups respond to various types of content. These insights lay the groundwork for \acrshort{CPL}.

Second, given the dynamic nature of user preferences and frequent interactions between users and the system, movie recommendation is typically approached as an online \acrshort{CPL} problem. The primary challenge is balancing the exploitation of existing knowledge about user preferences with the need to explore new data to improve counterfactual estimations. We will detail in this section that, offline counterfactual estimates obtained through \acrshort{CEL} offer valuable guidance for managing the exploration-exploitation trade-off in the early stages of online policy learning by informing data collection strategies. By further employing diverse online \acrshort{CPL} methodologies, the recommender system can dynamically adapt and refine its recommendations, leading to a more personalized and optimal user experience.

To simulate a real-world recommender system, we randomly sampled 1\% of the dataset to serve as the logged data currently available for offline analysis, while using the entire dataset to estimate the true reward distribution, which will be used to simulate the observed rewards during online interactions.


\subsubsection{\acrshort{CEL} on MovieLens}
In \acrshort{CEL}, we aim to estimate the potential outcomes (i.e. the expected ratings) of users across different movie genres. 
Using the T-learner approach, we fit a separate regression model for each genre (arm) to estimate the expected rating for each individual user. Table \ref{tab:CEL_movielens} below provides a summary of the expected ratings for two subgroups, Female and Male, across these different movie genres.
\begin{table}[h!]
\centering
\begin{tabular}{c|c|c}
\hline
\textbf{Genre} & {Expected Rating (Female)} & {Expected Rating (Male)} \\ \hline
        Comedy  & 3.580 & 3.445 \\
        Drama   & 3.403 & 3.424 \\
        Action  & 3.282 & 3.073 \\
        Thriller & 3.512 & 3.236 \\
        Sci-Fi  & 3.082 & 2.958 \\ \hline
\end{tabular}
\caption{Expected ratings of movie genres for different gender group}\label{tab:CEL_movielens}
\end{table}

In table \ref{tab:CEL_movielens}, we can see that except for Drama, females tend to provide higher expected ratings compared to males across the various movie genres. Depending on the researcher's objectives, \acrshort{CEL} approaches, including both ATE and HTE estimators provided in Section \ref{sec:CEL_p1}, offer multiple avenues for understanding individual preferences in movie genres.

\subsubsection{\acrshort{CPL} on MovieLens}
\begin{figure}
    \centering
    \includegraphics[width=.8\linewidth]{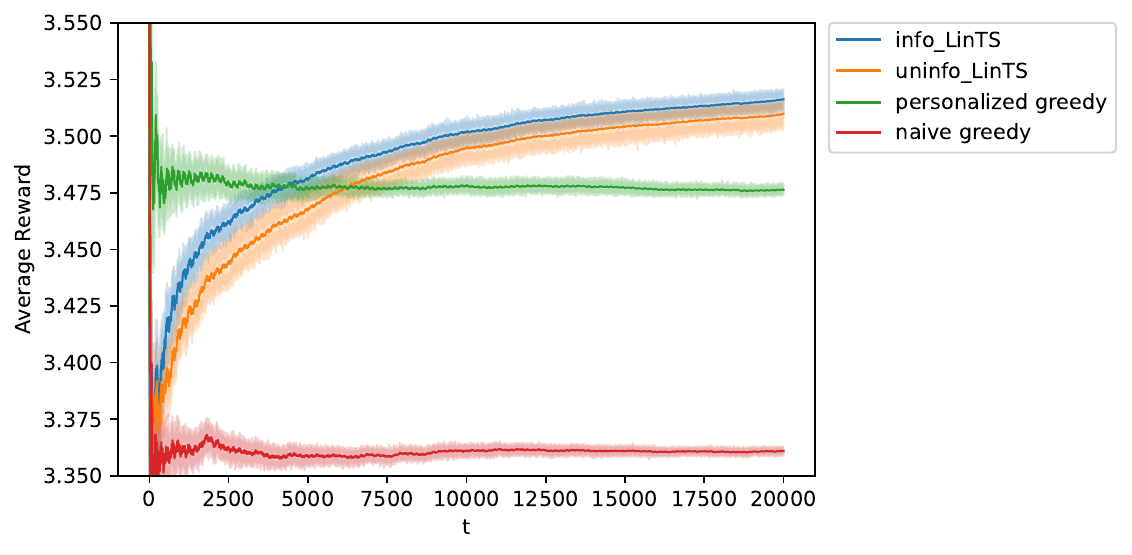}
    \caption{Simulation results for movie recommendation. Shaded areas indicate the 95\% confidence interval of the averages over 50 replicates.}
    \label{Fig:MovieLens_result}
\end{figure}

Movie recommendation has been extensively studied as an online bandit problem (paradigm 5), with its continuous feasibility of data collection. In this section, we simulate a real-world movie recommendation system using the full MovieLens dataset to demonstrate the necessity of online learning in decision-making and to further illustrate how insights from \acrshort{CEL} can be applied to \acrshort{CPL}. As an example, we implement the Linear Thompson Sampling (LinTS) algorithm to learn the optimal policy online. Specifically, we assume that for each arm, $R_t(a)\sim\mathcal{N}(s_a^T\gamma, \sigma^2)$, where $s_a$ is a vector contains feature information for the movie genre $a$, $\gamma$ is a vector of parameters, and $\sigma^2$ is the variance of the random noise. Using the 1\% logged data and the estimates from the \acrshort{CEL} step, we first estimated $\sigma$ and $\gamma$. These estimates were then used to construct an informative LinTS, with $\mathcal{N}(\hat{\boldsymbol{\gamma}}, 0.05\boldsymbol{I})$ as the prior distribution for $\gamma$, where $\boldsymbol{I}$ is an identity matrix and 0.05 is the prior variance, selected to reflect a reasonable confidence in the estimated $\hat{\boldsymbol{\gamma}}$ from the logged data while acknowledging the remaining uncertainty that requires further exploration. 

To highlight the advantages of incorporating information from the \acrshort{CEL} step, we also implement an uninformative LinTS using $\mathcal{N}(\boldsymbol{0},1000\boldsymbol{I})$ as the prior for $\gamma$. Additionally, in \acrshort{CEL}, we observed that both females and males, on average, prefer thriller movies with the highest expected ratings. Based on this insight, a simple greedy algorithm recommending thrillers to all users is considered as a baseline. We also implement a personalized greedy algorithm that generates tailored recommendations derived from more granular, individual-level estimates produced by \acrshort{CEL}. The simulation results are presented in Figure \ref{Fig:MovieLens_result}.

Overall, as expected, the naive greedy algorithm, which ignores personalized preferences, performs the worst. While the personalized greedy algorithm outperforms the LinTS algorithms in the early stages due to less random exploration, both LinTS algorithms continue to gather new information from the environment and eventually surpass the performance of the personalized greedy algorithm. Furthermore, when comparing the uninformative LinTS to the informative LinTS, it is evident that the latter performs better—especially in the early stages—thanks to the prior knowledge acquired from the logged data and the \acrshort{CEL} step.

\section{Conclusion and Discussion}\label{sec:conclusion}

Causality seeks to explain \textit{how actions lead to effects}, while decision-making focuses on \textit{how to take actions that yield the greatest effects}. In this paper, we present a comprehensive framework for decision-making through a causal lens. We decompose \acrshort{CDM} into three key tasks (\acrshort{CSL}, \acrshort{CEL}, \acrshort{CPL}) and six paradigms (distinguished by differences in data structure and offline/online learning settings), with each accompanied by a detailed review of state-of-the-art methods. We take an affirmative step in highlighting the widespread use of causality in decision-making by integrating all three tasks into a unified framework (see Section \ref{Sec:CSL}-\ref{sec:Online CPL}), with an extra emphasis on the assumption violated scenarios (see Section \ref{sec:assump_violated}). 
To provide a hands-on tutorial, we developed a \href{https://causaldm.github.io/Causal-Decision-Making}{GitHub notebook} with a Python package that summarizes popular methods for each task (\acrshort{CSL}, \acrshort{CEL}, \acrshort{CPL}), which are widely used in real-world applications. Combined with the real-data applications discussed in Section \ref{sec:real_data}, we believe that this paper offers a comprehensive tutorial for practitioners interested in the intersection of causality and decision-making.

Several intriguing extensions to classical decision making methods have been, and continue to be, actively explored. These include scenarios where we extend beyond the objective of reward maximization. For instance, recent studies have increasingly investigated how \acrshort{CSL} and \acrshort{CEL} techniques can enhance decision-making by incorporating additional objectives such as \textbf{fairness} and \textbf{explainability}. 
Motivated by the raising awareness of potential discrimination issues, which is essential in building a trustworthy recommendation system, \acrshort{SCM} is widely utilized to help understand the \textbf{fairness} issue. 
\citet{zhang2018fairness} decomposes the effect of a natural variation of a feature, and adopts the \acrshort{SCM} to infer and distinguish different types of natural discriminations; \citet{huang2022achieving} evaluates the counterfactual effect of sensitive attributes on the reward and limits the action space to arms satisfying the counterfactual fairness constraints; and \citet{balakrishnan2022scales} defines a \acrfull{PCE} to quantify the causal effect of protected attribute on reward through a specific path and formulates the fairness-aware recommendation problem as a constrained \acrshort{MDP} problem. Causal knowledge is also useful in enhancing the \textbf{explanability} of decision making. \citet{madumal2020explainable} introduced an action influence model that captures the causal relationships between variables using structural equations. By continuously learning the \acrshort{SCM}, they provide insights into the behavior of \acrshort{RL} agents by generating explanations for ``why A" and ``why not A" questions through counterfactual reasoning based on the learned \acrshort{SCM}. Instead of focusing on explaining a single action choice, \citet{tsirtsis2021counterfactual} aims to explain an observed sequence of multiple, interdependent actions. In scenarios involving multiple agents or more complex environments, through counterfactual reasoning using \acrshort{SCM}, \citet{triantafyllou2022actual} investigated multi-agent \acrshort{RL} to disentangle the contributions of individual agents, while \citet{mesnard2020counterfactual} differentiated the effect of an action from that of external factors on future rewards. These interconnected topics not only highlight their synergy within causal decision-making, but also pave the way for exciting future research directions.

\newpage

\appendix

\printglossary[type=\acronymtype]

\printglossary
\newpage
\clearpage

\bibliographystyle{jds}
\bibliography{0_ref1}

\end{document}